%% file: main.tex
\documentclass[11pt,letterpaper]{Sengoku}
\pdfmapfile{+noto.map}
\usepackage{courier}
\usepackage{booktabs,array,multirow,longtable,pifont}
\usepackage{placeins,float,flafter,url}
\renewcommand{\floatpagefraction}{.8}
\definecolor{insightpurple}{HTML}{654078}
\newcommand{\inlinepoint}[2]{\textcolor{insightpurple}{\ding{#1}~\textit{#2}}}
\newcolumntype{Y}{>{\raggedright\arraybackslash}X}
\newcommand{\ArenaTableFont}{\fontsize{9}{11}\selectfont}
\newcommand{\ArenaDenseTableFont}{\fontsize{8.5}{10.5}\selectfont}
\newcommand{\arena}{Embodied Agent Arena}
\renewcommand{\gradienttitle}[4]{%
  \noindent\begin{minipage}[c]{16mm}\centering\SengokuRenderTitleLogo\end{minipage}%
  \hspace{2.5mm}%
  \begin{minipage}[c]{\dimexpr\linewidth-37mm\relax}%
    \centering\fontsize{18.5}{23.5}\selectfont
    \textcolor{SengokuNavy}{#1}\\
    \textcolor{SengokuTitleTransitionA}{#2}\par
  \end{minipage}%
  \hspace{2.5mm}\begin{minipage}[c]{16mm}\mbox{}\end{minipage}\par
  {\centering\fontsize{15.5}{20}\selectfont
   \textcolor{SengokuTitleTransitionB}{#3}\space
   \textcolor{SengokuTitleRose}{#4}\par}}
\title{\gradienttitle{Are Frontier VLM Agents}{Ready to Be Robot Generalists?}{An Empirical Study}{with the Embodied Agent Arena}}
\shorttitle{Embodied Agent Arena}
\runningauthor{Haojian Huang et al.}
\affiliationlayout{inline}
\affiliation{hkustgz}{HKUST (GZ)}
\affiliation{cuhk}{The Chinese University of Hong Kong}
\affiliation{knowin}{Knowin AI}
\author[hkustgz,knowin]{Haojian Huang}
\author[knowin]{Pukun Zhao}
\author[cuhk,knowin]{Zexi Li}
\author[hkustgz,knowin]{Yehang Zhang}
\author[knowin]{Yangkai Wei}
\author[knowin]{Wenqian Li}
\author[knowin]{Han Yang}
\author[knowin]{Kaiwen Zhou}
\author[hkustgz,knowin]{Ying-Cong Chen}
\author[knowin]{Yinchuan Li}
\ExplSyntaxOn
\cs_set_protected:Npn \sengoku_author_item:nn #1#2
  {
    \hbox_set:Nn \l_sengoku_author_box
      {
        \strut
        \sengoku_author_badges:n {#2}
        {\bfseries #1}
        \tl_if_blank:nF {#2}
          {\nobreak\textsuperscript{\sengoku_affiliation_numbers:n {#2}%
            \str_case:nn {#1}
              {{Ying-Cong~Chen}{,\ensuremath{\dagger}}
               {Yinchuan~Li}{,\ensuremath{\dagger}}}}}
      }
    \mode_leave_vertical:
    \dim_compare:nNnTF {\box_wd:N \l_sengoku_author_box} > {\linewidth}
      {\hbox_unpack_drop:N \l_sengoku_author_box}
      {\box_use_drop:N \l_sengoku_author_box}
  }
\ExplSyntaxOff
\input{project-link}
\newcommand{\ArenaLogoGraphic}[1][16mm]{%
  \includegraphics[height=#1]{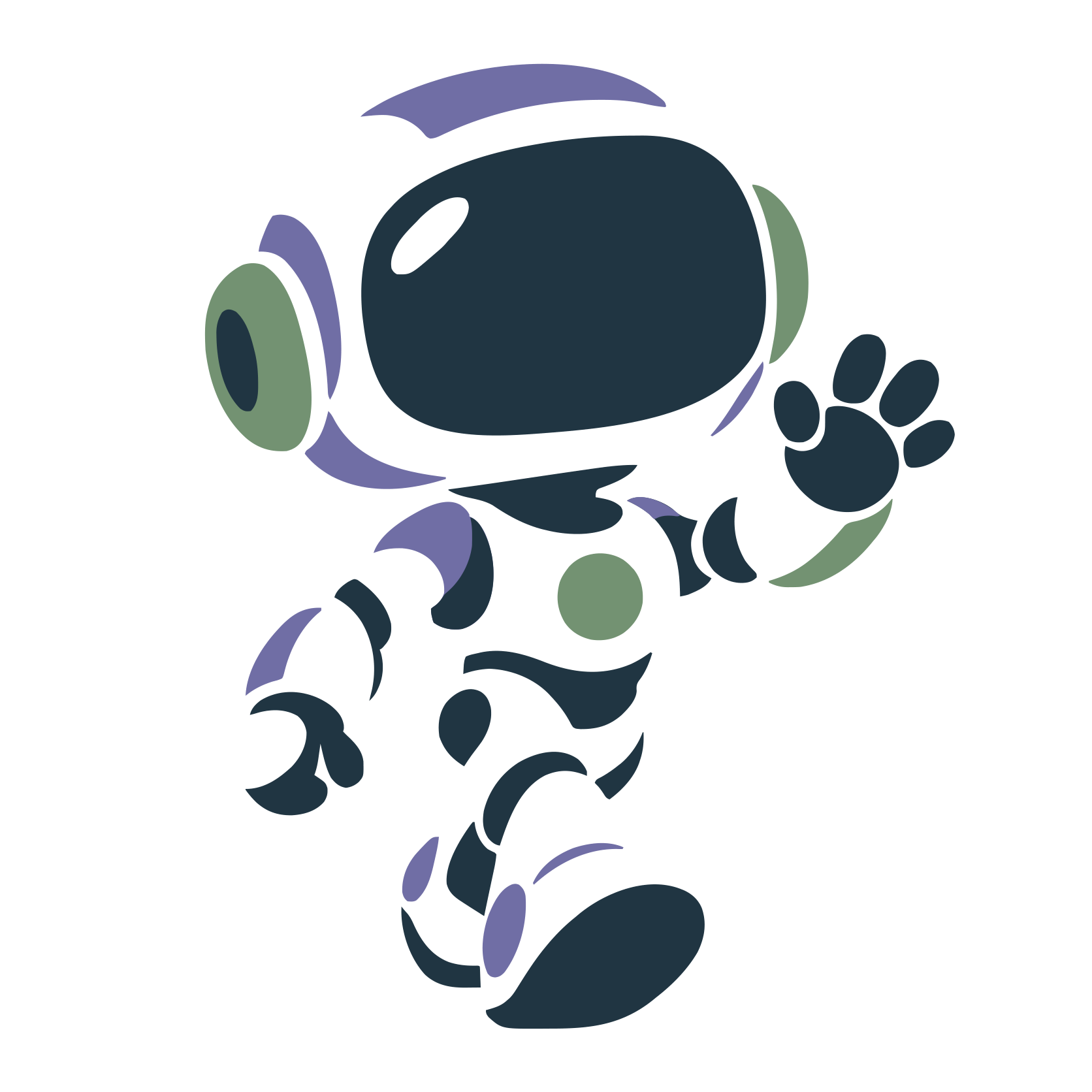}}
\renewcommand{\SengokuRenderTopLogoBlock}{}
\titlelogo[height=16mm]{figures/arena-logo-transparent.png}
\renewcommand{\SengokuRenderTitleLogo}{\ArenaLogoGraphic}
\newcommand{\ArenaProjectLink}{%
  \href{\ArenaProjectURL}{%
    \raisebox{-.28\height}{\ArenaLogoGraphic[5mm]}\hspace{1.5mm}%
    \textcolor{SengokuNavy}{Project Page}}}
\newcommand{\ArenaGitHubLink}{%
  \href{\ArenaGitHubURL}{%
    \raisebox{-.25\height}{\includegraphics[height=4mm]{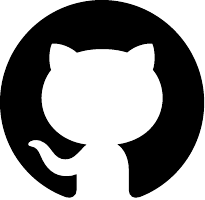}}\hspace{1.5mm}%
    \textcolor{SengokuNavy}{GitHub}}}
\renewcommand{\SengokuRenderMetadata}{%
  \par\addvspace{2mm}%
  {\centering\normalfont\SengokuSerif\fontsize{10.5}{12.5}\selectfont
   \ArenaProjectLink\hspace{7mm}\ArenaGitHubLink\par}}
\hypersetup{pdftitle={Are Frontier VLM Agents Ready to Be Robot Generalists? An Empirical Study with the Embodied Agent Arena},pdfauthor={Haojian Huang, Pukun Zhao, Zexi Li, Yehang Zhang, Yangkai Wei, Wenqian Li, Han Yang, Kaiwen Zhou, Ying-Cong Chen, Yinchuan Li},pdfsubject={Robotic capabilities of frontier vision-language agents},pdfkeywords={Vision-Language Agents; Robot Generalists; Embodied Evaluation}}
\input{sections/figures_verified}
\begin{document}
\input{sections/abstract}
\maketitle
\begingroup
\renewcommand{\thefootnote}{\ensuremath{\dagger}}
\footnotetext[0]{Corresponding authors.}
\endgroup
\VerifiedOverview
\input{sections/introduction}
\input{sections/related}
\input{sections/arena}
\input{sections/experiments}

\input{sections/discussion}
\begingroup
\let\ArenaOriginalBibliography\thebibliography
\renewcommand{\thebibliography}[1]{\ArenaOriginalBibliography{#1}\addtolength{\itemsep}{5.5pt}\finalhyphendemerits=100000000\exhyphenpenalty=10000}
\providecommand{\doi}[1]{\href{https://doi.org/#1}{\mbox{doi: #1}}}

\input{main.bbl}
\bibliographystyle{iclr2027_conference}
\endgroup
\par
\appendix
\renewcommand{\floatpagefraction}{.95}
\setcounter{topnumber}{2}
\setcounter{bottomnumber}{1}
\setcounter{totalnumber}{2}
\setcounter{table}{0}
\renewcommand{\thetable}{A\arabic{table}}
\renewcommand{\theHtable}{appendix.\arabic{table}}
\setcounter{figure}{0}
\renewcommand{\thefigure}{A\arabic{figure}}
\renewcommand{\theHfigure}{appendix.\arabic{figure}}
\setlength{\LTcapwidth}{\linewidth}
\makeatletter
\setlength{\@fptop}{0pt}
\setlength{\@fpsep}{12pt}
\setlength{\@fpbot}{0pt plus 1fil}
\makeatother
\input{sections/appendix}

\end{document}

%% file: project-link.tex
\newcommand{\ArenaProjectURL}{https://embodied-agent-arena.github.io/embodied-agent-arena/}
\newcommand{\ArenaGitHubURL}{https://github.com/embodied-agent-arena/embodied-agent-arena}

%% file: sections/figures_verified.tex
\newcommand{\VerifiedOverview}{%
\begin{figure}[t]
\centering
\includegraphics[width=\linewidth]{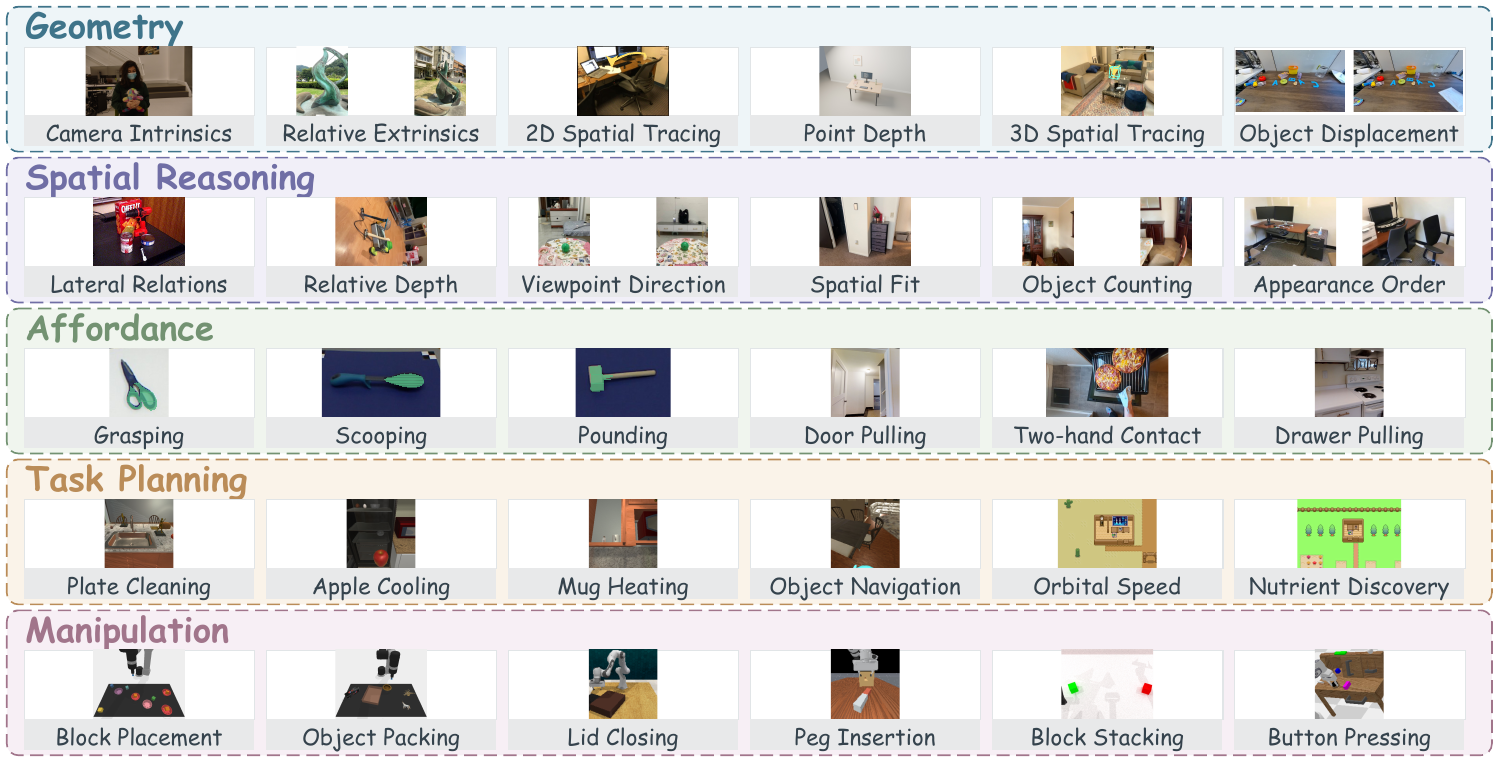}
\caption{Overview of Embodied Agent Arena. Each row illustrates six tasks in one capability domain, spanning camera and scene estimation, spatial relations, functional grounding, goal-directed planning, and robot manipulation. Fixed-input probes test estimates and judgments; interactive episodes test goal completion. Spatial-tracing overlays show reference paths and a 3D target region.}
\label{fig:overview}
\end{figure}%
}

%% file: sections/abstract.tex
\begin{abstract}
A robot generalist must interpret a scene, act on it, and satisfy every condition of a task. How far do frontier vision-language models (VLMs) meet these demands? We introduce Embodied Agent Arena, evaluating ten VLM agents across Geometry, Spatial Reasoning, Affordance, Task Planning, and Manipulation. Its 1,000 cases combine 32 established sources with GeoProbe, our 168-case benchmark for camera, motion, depth, and scale estimation. A minimal harness leaves task decisions to the model through source-native interfaces. Closed-weight models lead all five domains, whereas open-weight VLMs show useful spatial reasoning but substantially lower planning and manipulation success. Beneath these averages, similar geometric errors conceal different scale and shape distortions, and valid region estimates need not identify usable action points. During execution, agents can achieve early subgoals yet lose object control, bind an action to the wrong receiving region, or stop before the required physical state holds. Additional computation does not uniformly resolve these failures: code repair can restore execution without completing the task, while self-review can reverse correct judgments. Failure costs also differ: the strongest planners concentrate failed-token expenditure in prolonged body-control tasks, whereas other models fail broadly across procedures. Matched traces show how successful recovery uses feedback to restore a missing prerequisite and advance the remaining task.
\end{abstract}

%% file: sections/introduction.tex
\section{Introduction}
A robot generalist must turn scene understanding into actions that satisfy a complete goal. In an object-placement task, locating the target and grasping it are intermediate steps; success also requires placing it at the specified destination. Vision-language models (VLMs) increasingly address spatial relations~\citep{threedsr,vsi} and action-relevant grounding~\citep{reasonaff,robospatial}. Learned visuomotor policies connect observations to robot actions~\citep{openvla,pi0}, while language-model controllers select skills or generate executable programs~\citep{saycan,codepolicies}. Perception support and visual action rehearsal further extend these systems~\citep{capx,planar,li2026roboharness,zhang2026worldaction}. As these systems take on more diverse tasks, the central question becomes: \emph{Are frontier VLM agents ready to be robot generalists?}

Answering this question requires identifying which task requirements remain unsatisfied. For example, a trajectory can closely follow a reference yet miss its endpoint, while approaching a seat and issuing a sit action need not establish a seated state. Average accuracy can obscure these failures, whereas final failure alone hides achieved subgoals. Evaluation must therefore distinguish prediction accuracy, achieved subgoals, and final task success.

To distinguish these requirements systematically, we introduce \arena{}, evaluating ten VLM agents across Geometry, Spatial Reasoning, Affordance, Task Planning, and Manipulation (Figure~\ref{fig:overview}). The arena combines established benchmarks with GeoProbe, whose controlled variations in camera motion, object motion, and scale references diagnose geometric estimation. A minimal harness retains source interfaces and leaves task decisions to the model. This framework separates metric precision, functional grounding, and native task success, while execution traces identify achieved subgoals and unmet requirements.

The comparison shows that closed-weight models lead all five domains, whereas open-weight VLMs complete far fewer Planning and Manipulation tasks. Yet the aggregate gaps conceal distinct errors. Comparable depth errors can arise from a common scale bias or from residual distortion; a region can overlap the correct functional support while its proposed action point misses that support. Execution makes these distinctions consequential: a valid placement can address the outer fixture instead of its basin, and a precise gripper motion can leave the object ungrasped. Longer sequences then require maintaining or restoring the physical state on which the next subgoal depends.

These findings also explain why additional computation has uneven value. Code repair may restore an admissible program without establishing the required object state, while self-review can fix coordinate bounds and introduce new relational errors. The costs are similarly structured: for the strongest planners, failed-token expenditure is concentrated in prolonged body-control attempts, whereas other models fail across a broader range of procedures. By connecting these distributions to matched action--feedback traces, the arena identifies whether recovery must change a target binding, repair a prerequisite state, or revise an unsupported claim of task completion.

\Needspace{3\baselineskip}
Our contributions are:\nopagebreak[4]
\begingroup
\setlength{\leftmargini}{1em}
\setlength{\labelsep}{0.35em}
\begin{itemize}
\item \textbf{An evaluation centered on complete task requirements.} We evaluate prediction accuracy, functional grounding, subgoal achievement, and final task success across five domains.
\item \textbf{A unified arena with a new geometric benchmark, GeoProbe.} The 1,000-case arena combines 32 established sources with 168 GeoProbe cases, retaining native operations and success criteria in a shared model-directed evaluation framework.
\item \textbf{A diagnosis of task failure after successful intermediate steps.} Ten-model comparisons reveal complementary successes and shared failures; traces identify invalid endpoints, unmet goal conditions, and attempts that achieve no further subgoals.
\end{itemize}
\endgroup

%% file: sections/related.tex
\section{Related Work}
\label{sec:related}
\input{sections/table_benchmark_comparison}
Embodied evaluation has developed along two complementary directions. Control suites measure execution and task transfer~\citep{rlbench,calvin,libero}. Varying environments and physical deployment broaden this evaluation to robustness and multiple dimensions of manipulation competence~\citep{robotwin,robodojo}. Visual diagnostics isolate spatial inference~\citep{threedsr,vsi} and interaction grounding~\citep{robospatial,threedoi}. Scene and task question answering~\citep{majumdar2024openeqa,sermanet2023robovqa} extends this diagnostic view to embodied decisions~\citep{eai,bear}. Scene-linked assessment connects grounding, relational inference, and action to distinguish performance at individual stages from errors propagated between stages~\citep{ruan2026anvil}. As agents undertake longer interactions, final success leaves important aspects of execution unresolved: adherence to constraints~\citep{park2026howlevel}, use of relevant history~\citep{luo2026beyondsuccess}, and recognition of progress and recovery~\citep{liu2026robovalue}. These developments motivate measuring decisions within an episode alongside complete outcomes.

The transition from judgment to action depends on where control is represented. Learned policies map observations and instructions directly to actions~\citep{brohan2023rt1,zitkovich2023rt2}, with broader task coverage motivating generalist policies~\citep{octo,openvla,pi0}. Language-mediated control instead composes supplied skills~\citep{saycan,huang2023innermonologue} or executable programs~\citep{codepolicies,singh2023progprompt}; hybrid coordination selects between modular skills and learned policies from the current state~\citep{zhou2026roboaware}. Explicit programs make dependencies~\citep{capx,du2025dependeval}, geometric feedback~\citep{voxposer}, and reusable interfaces~\citep{li2026racap} available for model-directed composition. Evaluation consequently shifts from the availability of individual actions to the controller's ability to coordinate a complete procedure.

Composition also requires evidence that remains useful as the scene changes. Visual grounding can fail under physical stress or misleading background cues~\citep{li2025scagiqa,wu2026robostressbench,huang2026noplace}, motivating methods that repair context and reason over evolving spatial evidence~\citep{huang2026findfixreason,zhao2026evoempirbench,zhao2026star}. Demonstrations supply task context~\citep{vima,huang2026roboticl}; retained histories preserve state~\citep{worldlines}; verification checks whether execution meets its conditions~\citep{guo2024doremi,affordanceharness}. This support changes what the overall system can accomplish~\citep{wang2026cracking}, making the supplied interfaces and feedback part of the evaluation design.

Broader benchmarks bring perception and goal-directed interaction into one evaluation~\citep{embodiedbench,embodiedevel}, extending visual probes toward navigation~\citep{embodiedarena,rasheed2026frontiervision} and diverse manipulation tasks~\citep{vlabench}. Table~\ref{tab:benchmark-comparison} compares their coverage, diagnostics, and scale. \arena{} spans metric geometry, spatial inference, functional grounding, planning, and manipulation, with GeoProbe isolating camera, motion, and scale requirements. Its prediction metrics and execution records complement final task success by identifying estimation errors and unsatisfied goal conditions.

%% file: sections/table_benchmark_comparison.tex
\begin{table}[t]
\centering
\caption{\textbf{Evaluation coverage, diagnostic analysis, and sample scale.} Fixed probes use supplied observations; region probes score image grounding; manipulation requires arm/gripper control; memory spans sessions. Calib./pose tests camera calibration or relative pose; Forms denotes task-type/condition breakdowns. Scale counts QA items or task episodes; NR: aggregate count not reported (Appendix~\ref{app:benchmark-resources}).}
\label{tab:benchmark-comparison}
\begingroup
\fontsize{9}{11}\selectfont
\setlength{\tabcolsep}{3pt}
\renewcommand{\arraystretch}{1.2}
\newcommand{\benchyes}{\textcolor[HTML]{527653}{\ding{51}}}
\newcommand{\benchno}{\textcolor[HTML]{69737A}{\ding{55}}}
\begin{tabularx}{\linewidth}{@{}l>{\centering\arraybackslash}p{.17\linewidth}*{4}{>{\centering\arraybackslash}X}>{\centering\arraybackslash}p{41pt}>{\centering\arraybackslash}p{49pt}>{\centering\arraybackslash}p{34pt}@{}}
\toprule
& \textcolor[HTML]{B47B42}{\textbf{Sample scale}} & \multicolumn{5}{c}{\textcolor[HTML]{3F748A}{\textbf{Evaluation coverage}}} & \multicolumn{2}{c}{\textcolor[HTML]{706EA5}{\textbf{Diagnostic analysis}}}\\
\cmidrule(lr){2-2}\cmidrule(lr){3-7}\cmidrule(l){8-9}
Benchmark & Instances & Fixed & Region & Nav. & Manip. & Memory & Calib./pose & Forms\\
\midrule
OpenEQA~\citeyearpar{majumdar2024openeqa} & 1,636 & \benchyes & \benchno & \benchyes & \benchno & \benchno & \benchno & \benchyes\\
VLABench~\citeyearpar{vlabench} & NR & \benchyes & \benchno & \benchno & \benchyes & \benchno & \benchno & \benchyes\\
BEAR~\citeyearpar{bear} & 4,469 & \benchyes & \benchyes & \benchno & \benchyes & \benchno & \benchno & \benchyes\\
EmbodiedBench~\citeyearpar{embodiedbench} & 1,128 & \benchno & \benchno & \benchyes & \benchyes & \benchno & \benchno & \benchyes\\
EmbodiedEval~\citeyearpar{embodiedevel} & 328 & \benchno & \benchno & \benchyes & \benchno & \benchno & \benchno & \benchyes\\
Embodied Arena~\citeyearpar{embodiedarena} & NR & \benchyes & \benchyes & \benchyes & \benchno & \benchno & \benchno & \benchyes\\
CaP-X~\citeyearpar{capx} & 700/tier & \benchno & \benchno & \benchyes & \benchyes & \benchno & \benchno & \benchyes\\
WorldLines~\citeyearpar{worldlines} & 331 & \benchyes & \benchno & \benchno & \benchno & \benchyes & \benchno & \benchyes\\
Hard/Easy Vision~\citeyearpar{rasheed2026frontiervision} & NR & \benchyes & \benchyes & \benchyes & \benchno & \benchno & \benchno & \benchyes\\
\midrule
\rowcolor[HTML]{EEF5F8}
\textbf{Embodied Agent Arena} & \textbf{1,000} & \benchyes & \benchyes & \benchyes & \benchyes & \benchno & \benchyes & \benchyes\\
\bottomrule
\end{tabularx}
\par\smallskip
\raggedright\footnotesize\noindent
\benchyes\ included; \benchno\ not evaluated. BEAR manipulation: separate BEAR-Agent online validation.\par
\endgroup
\end{table}

%% file: sections/arena.tex
\begin{figure}[t]
\centering
\includegraphics[width=\linewidth]{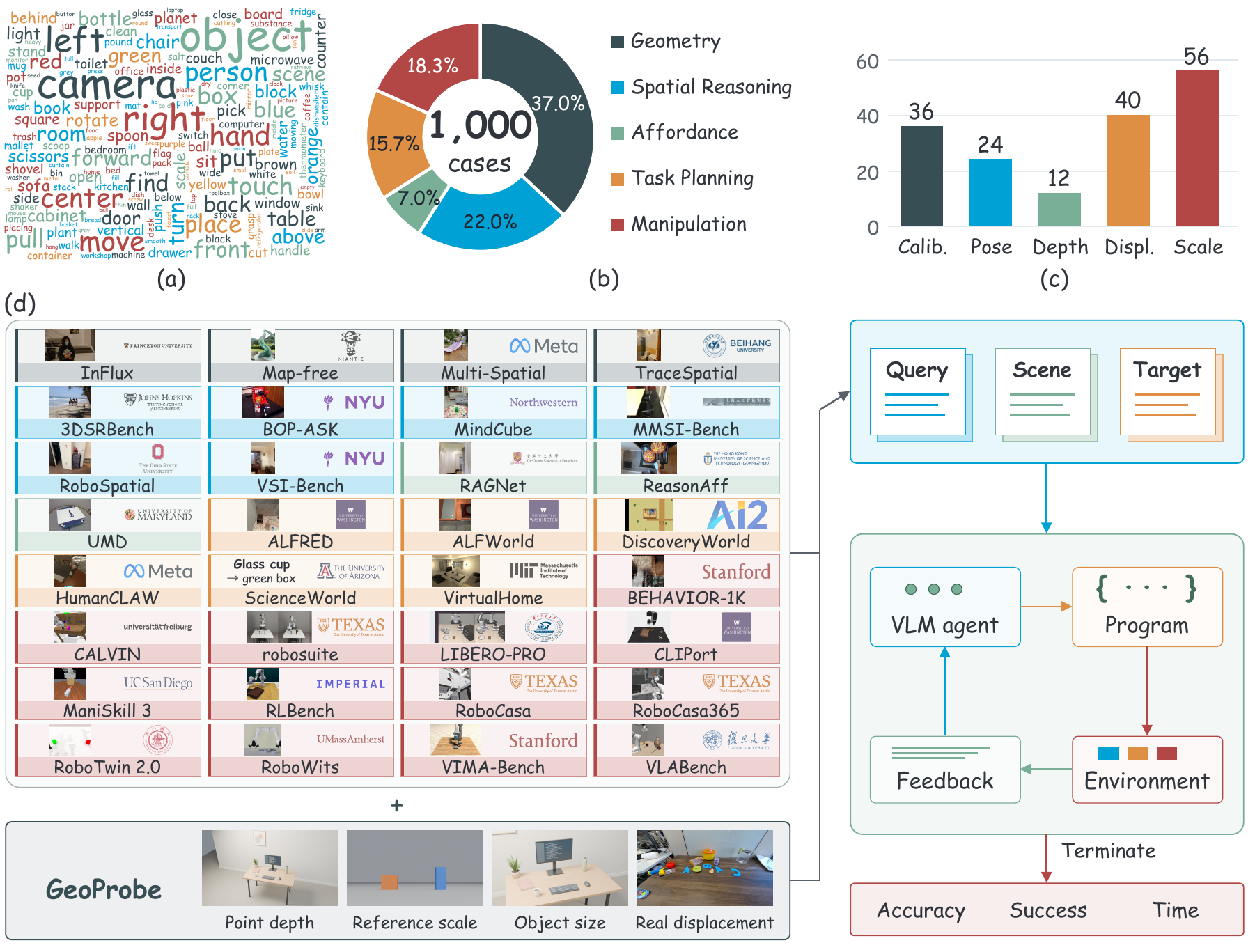}
\caption{Construction of the 1,000-case arena. (a) Representative task vocabulary across sources. (b) Five-domain composition, including 112 spatial-tracing cases in Geometry. (c) GeoProbe case counts for calibration, pose, depth, displacement, and scale. (d) Source queries, scenes, and targets form cases for a shared agent--environment loop; state persists between rounds, and native criteria determine termination and scoring. Colors identify domains; logos denote source institutions.}
\label{fig:construction-statistics}
\end{figure}

\section{The Embodied Agent Arena}
\label{sec:arena}
\subsection{Benchmark Construction and Capability Domains}
\label{sec:construction}

\noindent\textbf{Prediction, grounding, and interactive task success.}
The arena organizes 1,000 cases from 32 established sources and GeoProbe into five capability domains (Figures~\ref{fig:overview}--\ref{fig:construction-statistics}; Appendix~\ref{app:inventory}). Each case specifies observations, a query or goal, available operations, and scoring. Geometry, Spatial Reasoning, and Affordance isolate what the agent must estimate or ground; Task Planning and Manipulation test whether decisions produce the required environment state. GeoProbe adds 168 cases that vary camera motion, object motion, and scale references in controlled Blender scenes, complemented by real images. Separating these factors makes it possible to distinguish a failure to infer the scene from a failure to complete an interaction.

\noindent\textbf{Geometry: metric precision and feasibility.}
Agents estimate intrinsics, pose, depth, dimensions, or displacement from images. InFlux and Map-free test calibration and relative pose~\citep{influx,mapfree}; GeoProbe varies camera motion and scale references. TraceSpatial-Bench adds paths on 56 scenes~\citep{robotracer}. Trajectory error quantifies path similarity; initiation, arrival, and collision checks test spatial feasibility.

\noindent\textbf{Spatial Reasoning: relations and reference frames.}
Agents infer relations, viewpoints, configurations, and temporal order from images or video~\citep{threedsr,mindcube,vsi}. Comparing object distances, imagining another observer's view, and inferring camera motion require different reference-frame transformations. Questions specify the relevant frame and expect a category, quantity, or point set.

\noindent\textbf{Affordance: contact and functional extent.}
Agents ground an intended action through a contact point or region. UMD tests tool functions such as cutting and scooping~\citep{umd}; instruction- and action-conditioned 3DOI queries test interaction contacts~\citep{threedoi}. Point validity checks whether a predicted contact point lies inside the reference region. Region IoU measures intersection over union with the reference mask, using native boxes and, separately, direct polygons.

\noindent\textbf{Task Planning: satisfying all goal conditions.}
Agents compose source-provided operations into household, navigation, or scientific procedures~\citep{alfred,scienceworld}, using visual, textual, or symbolic feedback. Compound goals require several conditions to hold, such as heating an object and placing it at the destination. For terminal state $s_T$ and required predicates $g_1,\ldots,g_K$, complete success is
\begin{equation}
\mathrm{Success}(s_T,g)=\mathbf{1}\!\left[\bigwedge_{k=1}^{K}g_k(s_T)\right].
\label{eq:goal-completion}
\end{equation}
Where available, condition-level records identify unsatisfied predicates in Equation~\eqref{eq:goal-completion}; aggregate intermediate scores summarize partial task achievement.

\noindent\textbf{Manipulation: contact conditions and sequential goals.}
Agents execute robot motions or skills for placement, insertion, articulation, and coordinated handling~\citep{rlbench,robotwin}. Language-conditioned variants change the target relation or object~\citep{vlabench}; sequential tasks require completing ordered subgoals from the states left by preceding actions. Native task success records completion, while condition-level scores expose achieved relations and control events.

\subsection{Unified Agent Harness}
\label{sec:harness}
\noindent\textbf{Task decisions remain with the model.}
The harness provides persistent interaction through each source's observations and operations (Figure~\ref{fig:construction-statistics}d). Models interpret observations, select operations, and construct action sequences from feedback. We use \emph{native capability} to mean these model-directed decisions through the supplied source interfaces, including their helpers. The baseline adds no perception or control modules beyond those interfaces. Geometry, Spatial Reasoning, Affordance, and Planning use a Python runtime; Manipulation follows its native agent loop. Persistent variables let the agent build on earlier computations, while images produced during a round enter the next model request. Appendix~\ref{app:interfaces} specifies the interfaces; the supplementary review study adds specialist measurements (Section~\ref{sec:further})~\citep{a4agent,affordanceharness}.

\noindent\textbf{Termination is distinct from success.}
Fixed-input submissions commit predictions. Interactive episodes end on agent termination, an environment terminal condition, budget exhaustion, or an unrecovered API or execution error. The evaluator then checks the submitted output or environment state independently of the agent's declaration. Declaring completion therefore cannot substitute for satisfying the goal. The execution budgets are specified in Appendix~\ref{app:settings}.

\input{sections/table_profile}
\input{sections/figure3_tradeoffs}

\subsection{Evaluation Protocol and Metrics}
\label{sec:metrics}
\noindent\textbf{Task success and prediction accuracy.}
Spatial Reasoning and Planning report Pass@1; categorical spatial accuracy is also reported separately. Manipulation pairs native success with consecutive goals and task-specific condition or control-evidence scores. Geometry and numerical spatial queries use MAE, RMSE, AbsRel/MAPE, displacement error, or pose error as appropriate to the target. Tracing pairs trajectory RMSE with offline 3D constraint satisfaction.

\noindent\textbf{Functional grounding.}
The main table reports box--mask IoU for native rectangles and polygon--mask IoU for direct polygons on the same cases. Both compare rasterized predictions with the reference mask. Contact validity, box--box IoU, and strict box success provide complementary diagnostics in the appendix.

\noindent\textbf{Variation and resource use.}
Repeated metrics report the equally weighted mean and sample standard deviation (SD) of run-level scores. Continuous scores average valid predictions within each run; binary success rates retain all assigned cases, with error-terminated episodes scored as unsuccessful. Tokens, action calls, model rounds, and elapsed time describe different resources consumed before termination. Section~\ref{sec:setup} specifies repetitions, and Appendix~\ref{app:scoring} defines aggregation and accounting.

%% file: sections/table_profile.tex
\begin{table}[t]
\centering
\setlength{\abovecaptionskip}{0pt}
\caption{Performance across five domains. Box--mask and Polygon IoU use separate outputs on the same 70 cases. Planning covers 157 cases; Manipulation covers 182. Repeated metrics show mean $\pm$ run SD (five runs; Fable: three); the last three columns use one run. Best means are bold; Manipulation execution settings are detailed in Appendix~\ref{app:w4-comparison}.}
\label{tab:profile}
\fontsize{8.3}{10}\selectfont
\setlength{\tabcolsep}{2pt}
\begin{tabular*}{\linewidth}{@{\extracolsep{\fill}}l*{8}{c}@{}}
\toprule
 & \multicolumn{2}{c}{Geometry} & \multicolumn{2}{c}{Spatial Reasoning} & \multicolumn{2}{c}{Affordance} & Task Planning & Manipulation\\
\cmidrule(lr){2-3}\cmidrule(lr){4-5}\cmidrule(lr){6-7}\cmidrule(lr){8-8}\cmidrule(l){9-9}
Agent & Rot.$\downarrow$ & Trans.$\downarrow$ & Pass@1$\uparrow$ & AbsRel$\downarrow$ & Box--mask$\uparrow$ & Polygon$\uparrow$ & Pass@1$\uparrow$ & Task SR$\uparrow$\\
 & ($^\circ$) & (cm) & (\%) & & IoU & IoU & (\%) & (\%)\\
\midrule
Astra & $\mathbf{17.2}\!\pm\!6.7$ & $106.9\!\pm\!36.0$ & $69.4\!\pm\!1.0$ & $0.214\!\pm\!0.016$ & $0.286\!\pm\!0.010$ & 0.379 & \textbf{80.3} & 41.8\\
Sol & $27.0\!\pm\!1.5$ & $151.9\!\pm\!6.2$ & $60.7\!\pm\!2.5$ & $0.273\!\pm\!0.016$ & $0.264\!\pm\!0.010$ & 0.325 & 44.6 & 19.8\\
Fable & $24.2\!\pm\!8.3$ & $116.6\!\pm\!42.6$ & $62.0\!\pm\!4.6$ & $0.185\!\pm\!0.011$ & $0.284\!\pm\!0.013$ & 0.328 & 63.7 & 24.7\\
\mbox{Qwen-Max} & $20.5\!\pm\!3.5$ & $138.1\!\pm\!10.2$ & $68.9\!\pm\!1.1$ & $0.206\!\pm\!0.033$ & $0.272\!\pm\!0.017$ & 0.307 & 36.9 & 9.3\\
Gemini~3.8 & $23.7\!\pm\!5.6$ & $138.0\!\pm\!41.7$ & $65.4\!\pm\!1.3$ & $\mathbf{0.161}\!\pm\!0.028$ & $0.251\!\pm\!0.081$ & 0.291 & 48.4 & 18.7\\
\mbox{Qwen-397B} & $38.4\!\pm\!1.1$ & $249.8\!\pm\!10.2$ & $50.6\!\pm\!1.8$ & $0.251\!\pm\!0.027$ & $0.116\!\pm\!0.014$ & 0.092 & 19.1 & 9.9\\
\mbox{Qwen-27B} & $38.4\!\pm\!3.6$ & $213.7\!\pm\!12.6$ & $48.0\!\pm\!0.9$ & $0.288\!\pm\!0.023$ & $0.013\!\pm\!0.006$ & 0.051 & 19.1 & 11.0\\
Opus~5.5 & $23.4\!\pm\!3.7$ & $135.8\!\pm\!12.7$ & $67.5\!\pm\!5.1$ & $0.184\!\pm\!0.040$ & $0.311\!\pm\!0.020$ & 0.397 & 48.4 & \textbf{47.8}\\
\mbox{DeepSeek-1017} & $30.2\!\pm\!3.9$ & $182.0\!\pm\!29.0$ & $51.6\!\pm\!1.5$ & $0.309\!\pm\!0.035$ & $0.217\!\pm\!0.011$ & 0.249 & 31.2 & 20.9\\
Gemini~4~Argon & $18.5\!\pm\!1.8$ & $\mathbf{105.6}\!\pm\!9.6$ & $\mathbf{71.6}\!\pm\!0.7$ & $0.169\!\pm\!0.023$ & $\mathbf{0.356}\!\pm\!0.009$ & \textbf{0.441} & 70.1 & 44.0\\
\bottomrule
\end{tabular*}
\end{table}

%% file: sections/figure3_tradeoffs.tex
\begin{figure}[!t]
\centering
\includegraphics[width=\linewidth]{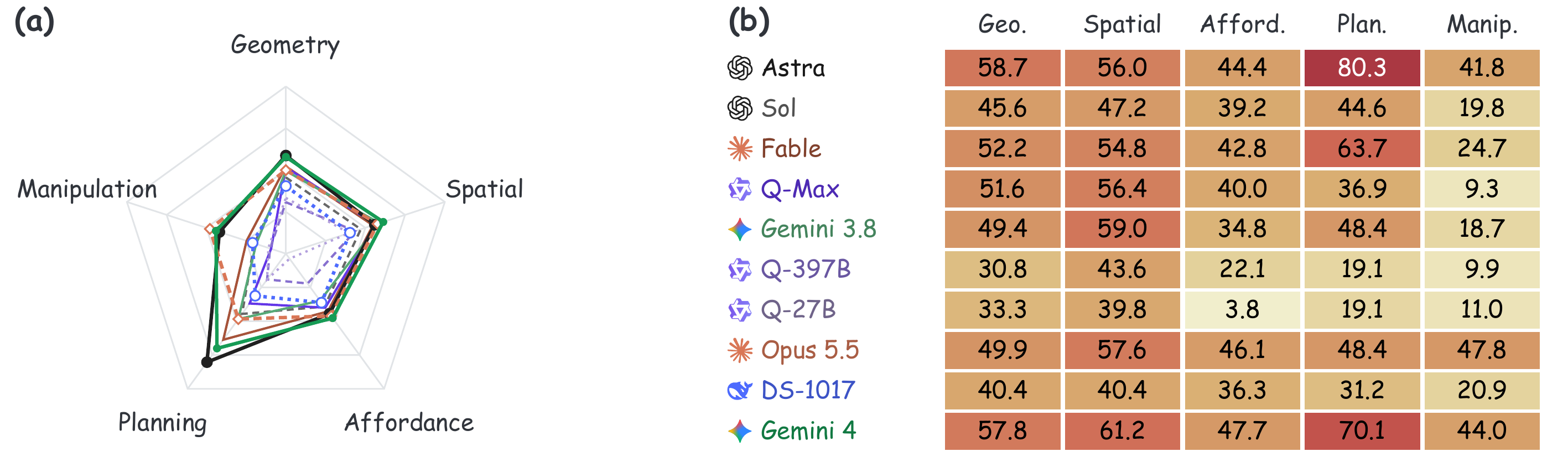}
\caption{Complementary capability profiles across ten agents. (a) Thicker lines highlight Astra, Opus~5.5, \mbox{DeepSeek-1017}, and Gemini~4~Argon. Row labels in (b) identify model colors in both panels. (b) Domain indices on fixed 0--100 scales (Appendix~\ref{app:cost-profile}), each summarizing its own metrics. Affordance combines box--mask IoU and contact validity; Figure~\ref{fig:astra-affordance} examines polygons. Manipulation uses the recorded schedules and budgets in Appendix~\ref{app:w4-comparison}.}
\label{fig:capability-cost}
\end{figure}

%% file: sections/experiments.tex
\section{Experiments}
\label{sec:experiments}
\input{sections/figure4_astra}
\input{sections/figure5_spatial}
\input{sections/figure6_affordance}
\input{sections/figure7_planning}
\input{sections/figure8_manipulation}
We compare capability profiles, diagnose unmet task requirements, and examine how feedback changes predictions and execution.
\subsection{Experimental Setup}
\label{sec:setup}
\textbf{Models.}
We evaluate Astra, Sol, Fable, Opus~5.5, \mbox{Qwen-Max}, Gemini~3.8 Flash, Gemini~4~Argon, \mbox{Qwen-397B}, \mbox{Qwen-27B}, and DeepSeek-V4.1-Flash-1017 (\mbox{DeepSeek-1017}). \mbox{Qwen-397B}, \mbox{Qwen-27B}, and \mbox{DeepSeek-1017} have released VLM weights; the other evaluated visual models are closed-weight services. Appendix~\ref{app:settings} gives identifiers and release scope. \mbox{DeepSeek-1017} uses the \texttt{deepseek-flash} endpoint updated on 17 October 2026.

\textbf{Repetitions and execution conditions.}
Non-tracing Geometry, Spatial Reasoning, and point-and-box Affordance use five runs per case for nine models and three for Fable. We report the mean and sample SD across run-level scores. Tracing, direct polygons, Planning, and Manipulation each use one episode per assignment. Polygon IoU covers the same 70 Affordance cases for all ten agents (Appendix~\ref{app:polygon}). Planning and Manipulation retain all 157 and 182 binary assignments, respectively. Episodes terminated by API or execution errors count as unsuccessful, including two \mbox{Qwen-Max} HumanCLAW provider errors (Appendix~\ref{app:latest-results}). All ten agents are evaluated on the same assigned tasks through source-native interfaces, with source-provided observations and native success criteria. Manipulation retains each evaluation batch's recorded observation schedule and execution ceilings; Appendix~\ref{app:w4-comparison} distinguishes these settings. Appendices~\ref{app:opus-statistics}, \ref{app:deepseek}, and~\ref{app:argon} specify model-specific repetitions and configurations.

\Needspace{5\baselineskip}
\subsection{Main Results}
\label{sec:cross-model}
\textbf{Does perceptual strength predict execution success?}
Closed-weight models lead all five domains, with distinct profiles (Table~\ref{tab:profile}).
\inlinepoint{172}{Broad strengths.} Astra and Gemini~4~Argon occupy the top two positions in Spatial Reasoning and Planning, and the top three in Manipulation (Figure~\ref{fig:capability-cost}).
\inlinepoint{173}{Selective strengths.} \mbox{Qwen-Max} nearly matches Astra in Spatial Pass@1, 68.9\% versus 69.4\%, but completes 36.9\% versus 80.3\% of Planning tasks. Conversely, Opus~5.5 leads Manipulation at 47.8\% while trailing Astra and Fable in Planning. Fixed-input questions assess a supplied observation; interactive tasks additionally require choosing operations, tracking their effects, and meeting a terminal condition. These demands distinguish perceptual competence from procedural and control competence.
\inlinepoint{174}{Open-weight execution gaps.} \mbox{Qwen-397B}, \mbox{Qwen-27B}, and \mbox{DeepSeek-1017} achieve 48.0--51.6\% Spatial Pass@1, but only 19.1--31.2\% Planning and 9.9--20.9\% Manipulation success. Variation among closed-weight models likewise shows that release category does not specify a model's execution strengths.

\textbf{Where does model diversity help?}
Planning overlap separates broadly shared success from model-specific and shared failures (Figure~\ref{fig:astra-planning}).
\inlinepoint{172}{Broadly shared success is concentrated.} Only eight of 157 tasks are solved by all ten models; seven belong to symbolic VirtualHome. Conversely, 39 tasks have only one or two successful models, including twenty ALFRED or ALFWorld tasks. The contrast places much of the disagreement in visually grounded household procedures, rather than distributing it evenly across sources.
\inlinepoint{173}{Complementarity has diminishing returns.} Astra solves 126 tasks; adding Gemini~4 covers 137, and all ten cover 142. Thus one additional model supplies eleven of the sixteen extra successes, while the other eight add five. Yet the fifteen common failures are heterogeneous: eight sitting tasks, four ALFRED/ALFWorld procedures, two ScienceWorld experiments, and one television task. Their traces include both prolonged score plateaus and terminal failures after partial progress (Appendix~\ref{app:trace-diagnostics}). These shared failures require different corrections.

\textbf{Which rankings survive a different task mixture?}
\inlinepoint{172}{A lead spanning sources.} Removing ScienceWorld and VirtualHome lowers every Planning score, yet Astra still leads on the remaining 63 tasks at 66.7\%, followed by Gemini~4 at 60.3\% and Fable at 33.3\%. On ALFRED and ALFWorld alone, Astra completes 29/40 and Gemini~4 30/40, versus 4--8/40 for each open-weight VLM. This advantage extends to visually grounded household procedures.
\inlinepoint{173}{A lead tied to source weight.} Opus~5.5 leads Gemini~4 in Manipulation by 47.8\% to 44.0\% over assignments, but equal weighting of thirteen sources reverses their order to 44.5\% versus 47.4\% (Appendix~\ref{app:execution-analysis}). RoboTwin supplies 50/182 assignments and favors Opus, 23 versus fourteen successes, whereas Gemini~4 is stronger on CALVIN and RLBench. The same outcomes yield different leaders because the models specialize in different task collections. Aggregate superiority should therefore be read together with the distribution of task requirements.

\textbf{Why do unsuccessful attempts dominate computation?}
Separating task frequency, source budgets, and recorded behavior changes the interpretation (Table~\ref{tab:failure-cost-decomposition}).
\inlinepoint{172}{A costly residual in long interactions.} HumanCLAW supplies only 13/157 tasks but 81.7\% of Astra's failed tokens; its eight sitting tasks alone supply 63.7\%. These episodes allow 101 rounds and 2M tokens, versus eight rounds and 320k tokens for several other Planning sources. Excluding HumanCLAW lowers the failed-token share from 55.5\% to 19.4\% for Astra and from 64.0\% to 33.5\% for Gemini~4. Their failed-token expenditure is concentrated in a small, generously budgeted set of unresolved physical interactions.
\inlinepoint{173}{Failures spread across ordinary procedures.} These shares remain 55.3--67.8\% for the other closed-weight models and 73.6--85.8\% for the three open-weight VLMs. The deficit persists in text-only ScienceWorld, beyond visual perception. On its matched animal-transfer task, DeepSeek spends three of eight decision rounds inspecting state and stops with 89 environment steps unused; Opus identifies the blue-jay egg as an animal and completes the focus-and-transfer procedure. Both execute without a code exception. Repeated inspection consumes scarce decision rounds without resolving which object satisfies the goal.

\input{sections/astra_analysis}
\input{sections/further_investigations}

%% file: sections/figure4_astra.tex
\begin{figure}[t]
\centering
\includegraphics[width=\linewidth]{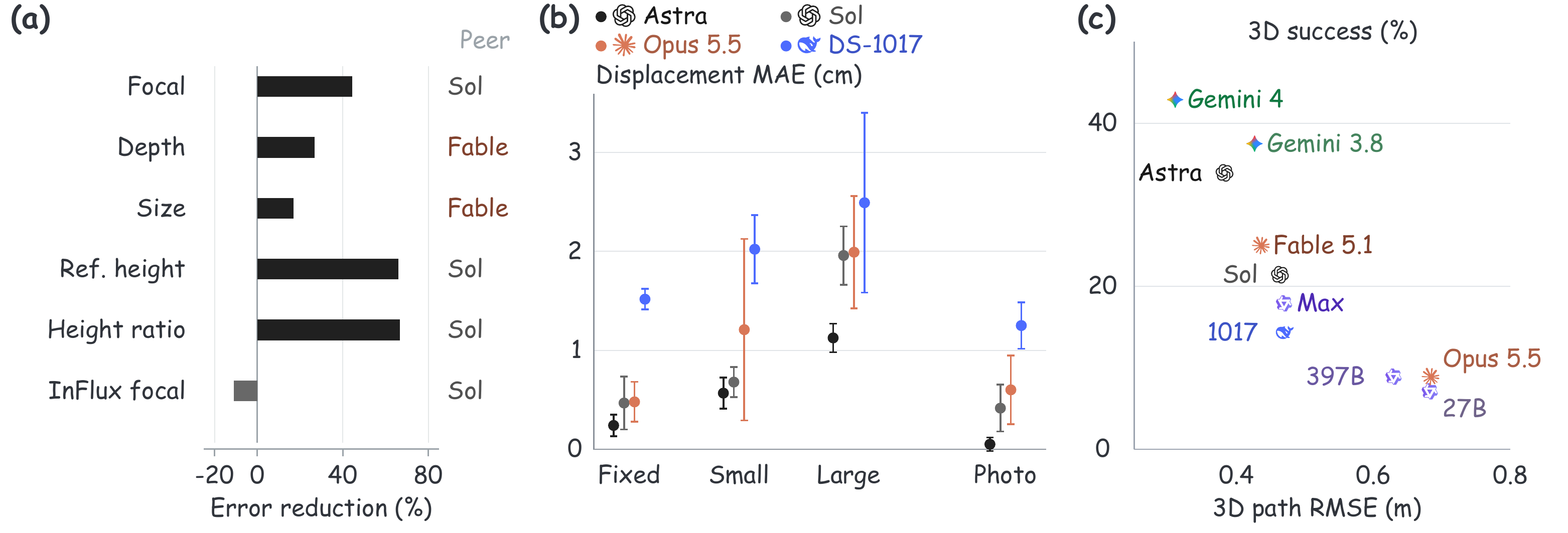}
\caption{Estimation error and trajectory constraint satisfaction. (a) Astra's error reduction relative to the lowest-error original peer; the first five targets are GeoProbe. (b) Displacement mean absolute error (MAE) with fixed, small-shift, and large-shift Blender cameras; Photo uses real image pairs. Points and whiskers show the mean $\pm$ sample SD of five run means. (c) Single-run 3D tracing error versus joint success for all ten agents. Logos mark measured coordinates; adjacent labels distinguish model versions. Details: Appendix~\ref{app:geometry-focus}.}
\label{fig:astra-geometry}
\end{figure}

%% file: sections/figure5_spatial.tex
\begin{figure}[t]
\centering
\includegraphics[width=\linewidth]{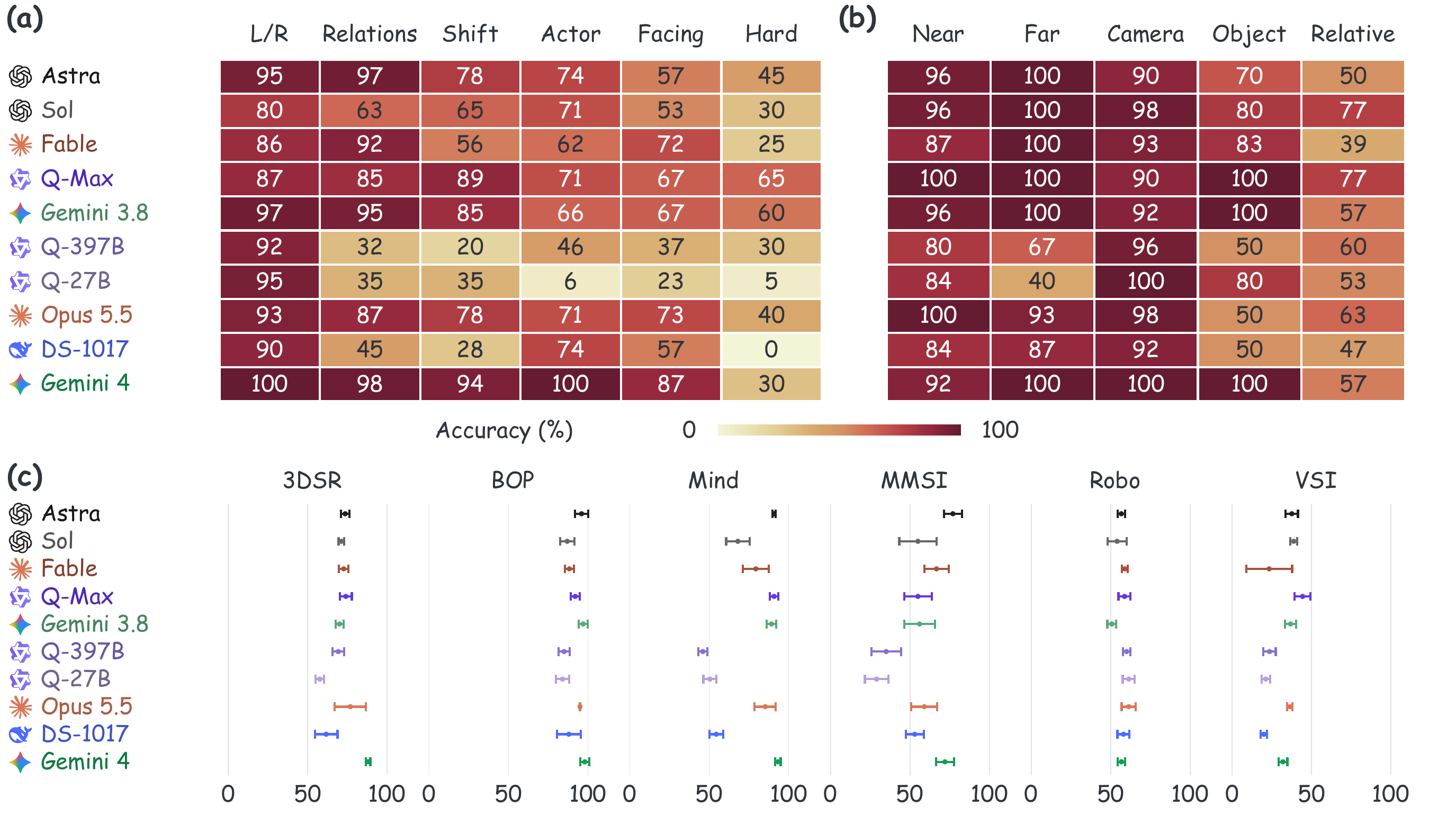}
\caption{Spatial strengths change with the required relation. (a) L/R: image left/right; Relations: multiview object relations; Shift: camera motion; Actor: person-centered directions; Facing: object orientation; Hard: three-object directions from an imagined viewpoint in video. (b) Near/Far: nearer/farther from the camera; Camera: camera-distance comparison; Object: distance to a reference object; Relative: nearest-object selection in video. Cells show mean accuracy (\%) for all ten models on the same groups. (c) Native source Pass@1 (\%): dots and whiskers show means and sample SD over five runs (three for Fable). 3DSR, Mind, Robo, and VSI denote 3DSRBench, MindCube, RoboSpatial-Home, and VSI-Bench. Appendix~\ref{app:spatial-focus} gives denominators. Gemini~3.8 and Gemini~4 denote Gemini~3.8 Flash and Gemini~4~Argon.}
\label{fig:astra-spatial}
\end{figure}

%% file: sections/figure6_affordance.tex
\begin{figure}[t]
\centering
\includegraphics[width=\linewidth]{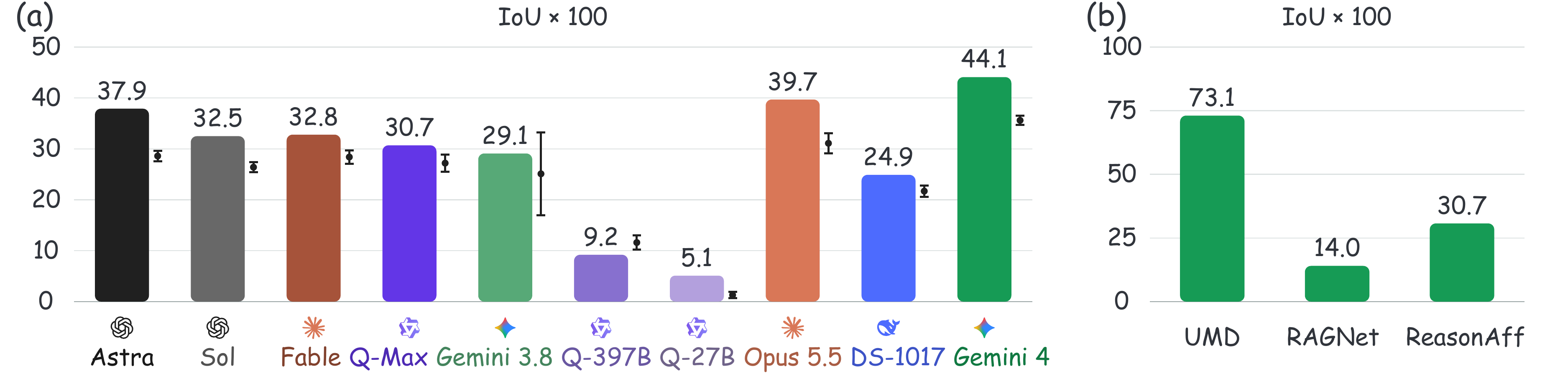}
\caption{Polygon outputs attain higher region overlap in nine of ten model comparisons. (a) Direct-polygon IoU (bars) and box--mask IoU (dots, with run SD) on the same 70 cases for all ten models. Boxes use five runs, or three for Fable; polygons use one separate run. Invalid or absent polygon outputs score zero. (b) Gemini~4~Argon, the aggregate leader, scores much higher on UMD tools than on RAGNet and ReasonAff interaction scenes. Appendix~\ref{app:polygon} and Figure~\ref{fig:affordance-detail} detail source and contact results.}
\label{fig:astra-affordance}
\end{figure}

%% file: sections/figure7_planning.tex
\begin{figure}[t]
\centering
\includegraphics[width=\linewidth]{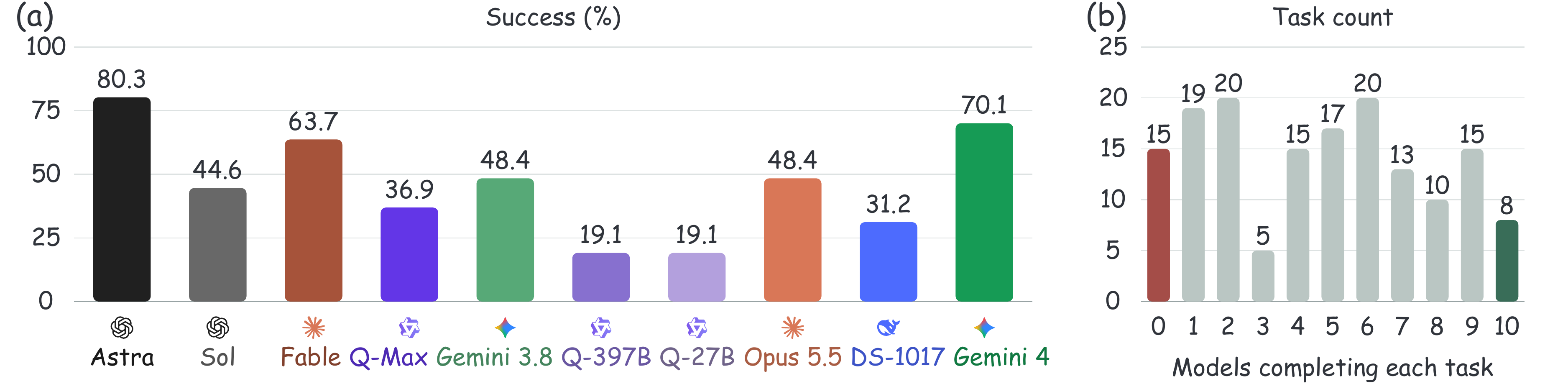}
\caption{Complementary successes coexist with shared failures. (a) Planning success across all 157 assigned tasks, with one episode per model and task. (b) The same tasks grouped by how many of the ten models complete them: 15 defeat every model, while only eight are completed by all ten. The union of completed tasks gives retrospective coverage of 142, versus Astra's 126. Figure~\ref{fig:planning-detail} and Appendix~\ref{app:planning-focus} provide source and goal-family diagnostics.}
\label{fig:astra-planning}
\end{figure}

%% file: sections/figure8_manipulation.tex
\begin{figure}[t]
\centering
\setlength{\abovecaptionskip}{5pt}
\setlength{\belowcaptionskip}{0pt}
\includegraphics[width=\linewidth]{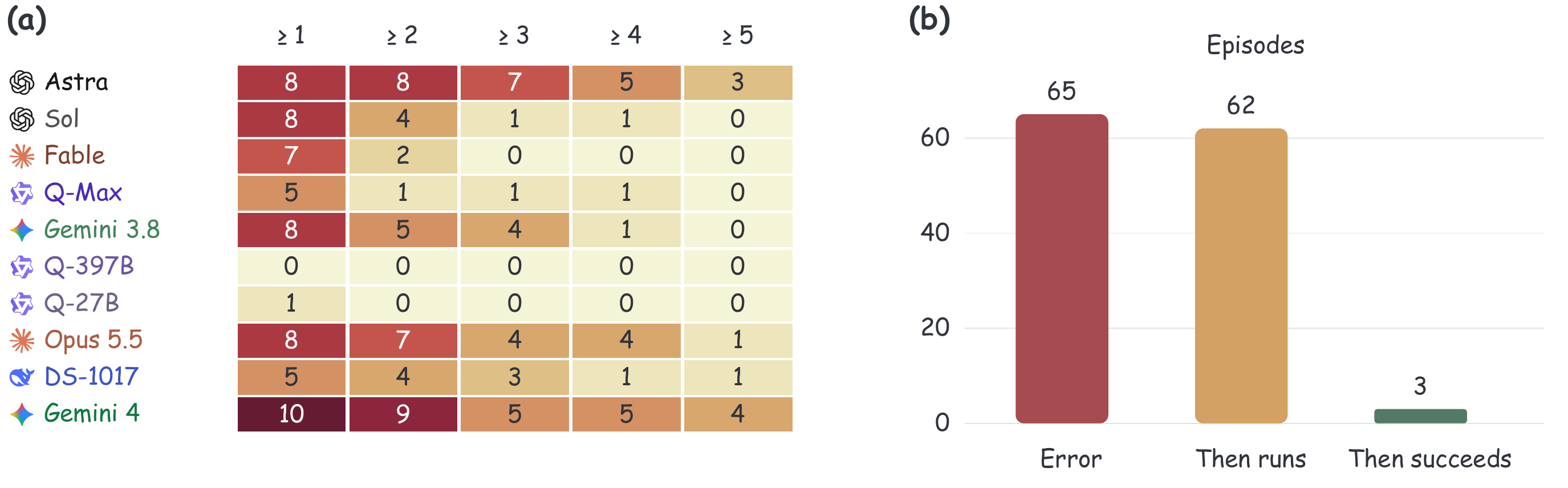}
\caption{Most CALVIN episodes remain incomplete after code execution resumes. All 100 CALVIN episodes: ten sequences per model. (a) Counts out of ten episodes per model reaching at least the indicated number of consecutive goals; the fifth column is complete-sequence success. Unverified progress receives no credit. (b) Of 65 episodes with an execution error, 62 later execute code without an error, but only three subsequently pass the native task verifier. These are nested episode counts; API transport failures, timeouts, and budget guards are excluded from execution errors. Appendix~\ref{app:calvin-recovery} gives model-level counts, missing-prefix handling, and the recorded execution settings.}
\label{fig:astra-manipulation}
\end{figure}

%% file: sections/astra_analysis.tex
\subsection{Diagnosing Unmet Task Requirements}
\label{sec:astra}
We distinguish errors in interpreting the observation, satisfying a precise condition, and completing dependent actions through diagnostic measurements and traces (Figures~\ref{fig:astra-geometry}--\ref{fig:astra-manipulation}).

\textbf{Geometry: which error remains after an apparently correct estimate?}
\inlinepoint{172}{Scale bias and relative-shape error require different diagnoses.} On GeoProbe depth, Gemini~4 and \mbox{Qwen-397B} have similar raw AbsRel, 0.299 and 0.312, but oracle scale alignment leaves 0.065 and 0.203. Gemini~4 has a median predicted/true depth ratio below one in all sixty case--run observations; much of its metric error is a common-scale component. The larger Qwen residual persists after per-case scalar rescaling. Astra combines 96.3\% ordinal accuracy with 0.104 raw AbsRel, falling to 0.038 after alignment. Relative ordering, metric calibration, and residual shape thus separate models that a single error average would group together.
\inlinepoint{173}{Camera sensitivity differs from a motion-detection floor.} With matched object interventions, Astra and Gemini~4 miss none of fifty moving observations under a fixed camera, but miss 26 and 37 under a large view change; neither falsely labels the stationary controls as moving. The additional errors therefore concern real object motion lost under camera change. In contrast, \mbox{Qwen-27B} misses all fifty moving observations in each camera condition. Its unchanged error is a failure floor, not robustness to the perturbation (Appendix~\ref{app:geometry-focus}).
\inlinepoint{174}{Projected endpoints do not settle physical constraints.} Astra hits both image endpoint regions on 35/56 TraceSpatial paths, yet 21 of these fail a 3D boundary condition. Gemini~4 improves to 45 image-valid paths, with sixteen such failures, but has ten collision-only failures versus Astra's one. The remaining failure profile differs across models: a path with valid endpoints must also respect intervening geometry. Joint checks distinguish projection, metric placement, and collision failures on the same trajectories (Appendix~\ref{app:trace-diagnostics}).

\textbf{Spatial Reasoning: is the failure in the relation or in returning an answer?}
\inlinepoint{172}{Lateral errors persist across repetitions.} All ten models reach 80--100\% on BOP-ASK left/right and 90--100\% on camera-depth ordering, but actor-relative sides separate them (Figure~\ref{fig:astra-spatial}). \mbox{Qwen-27B} answers six of seven actor-left/right cases incorrectly in all five runs, despite 19/20 correct judgments on actor-front cases. Gemini~4 answers all 35 actor-left/right observations correctly. The repeated lateral errors span both left and right ground truths; they are neither a constant-side response nor a uniformly weak judgment of actor-centered relations.
\inlinepoint{173}{Video scores also include submission failures.} Gemini~4 scores 30\% on four hard video-direction cases, but its twenty runs contain six correct answers, four incorrect answers, and ten token-guard stops without an answer. On one question, three runs answer correctly and two never submit. Consequently, the low aggregate mixes relational mistakes with failure to deliver a decision under the execution budget. Separating these outcomes preserves the all-assignment score while identifying whether to improve spatial judgment or the delivery of a decision.

\textbf{Affordance: which part of the action target is grounded?}
\inlinepoint{172}{A valid point can leave the functional extent unresolved.} Across 350 case--run predictions, \mbox{Qwen-397B} passes the point criterion alone 96 times, but both point and box criteria only eighteen times (Table~\ref{tab:affordance-joint}). The matched pulling example shows the boundary error: all four displayed models select the handle but their contours cover too little of the annotated region, yielding 0.096--0.140 IoU (Figure~\ref{fig:latest-affordance}).
\inlinepoint{173}{An accepted region need not supply a valid action point.} Gemini~4 exhibits the reverse pattern sixteen times: its box passes, while its point lies outside the functional mask, although inside its own box. In a scissors-grasp case, rectangle IoU is 0.784 but the chosen point misses the grasp mask; this repeats in four of five runs. Region enclosure and selection of a usable point are distinct requirements, even for one correctly localized object.
\inlinepoint{174}{Wrong-object selection precedes either geometric correction.} In the matched RAGNet two-hand case, Opus~5.5 selects the referenced target and reaches 0.942 IoU; Astra, Fable, and DeepSeek outline other objects and score zero. A sharper boundary around the wrong object cannot repair the instruction--target binding. These cases separate selecting the entity, delineating its functional support, and choosing a point on that support.

\textbf{Task Planning: does an executed action establish the required state?}
\inlinepoint{172}{The right operation can target the wrong receiving region.} On the matched ALFRED soap task, DeepSeek cleans the soap and places it in \texttt{Bathtub}, yet finishes with only 1/3 goal conditions. Gemini~3.8 initially reaches the same incomplete state, then picks up the soap and changes the destination to \texttt{BathtubBasin}, reaching 3/3; \mbox{Qwen-Max} also changes the receiving region after inspecting the parent receptacle and re-washing the soap (Table~\ref{tab:trace-state-changes}). The failure is more specific than forgetting a procedural step: a valid placement call refers to the outer fixture rather than its required basin. Feedback helps when it changes that binding, rather than merely repeating the intended operation.
\inlinepoint{173}{Attempted interaction can be mistaken for the terminal state.} HumanCLAW sitting has no recorded success, but its process evidence separates target acquisition from interaction. Fable finds no target in eight episodes; Astra and Gemini~4 record both proximity within 20\,cm and a sit action in five and seven, respectively (Table~\ref{tab:humanclaw-stages}). On the same couch case, Gemini~4 and DeepSeek issue \texttt{sit}, then declare themselves seated and stop after 44 and 70 rounds; both fail the private native verifier. Here a visible approach and an issued command do not establish the requested body--object relation. The trace establishes a stopping misjudgment; the private verifier does not identify which pose or contact condition remained unmet. \mbox{Qwen-397B} and \mbox{Qwen-27B} likewise show proximity plus a sit action in one and two of eight episodes without success.

\textbf{Manipulation: what must persist between successful actions?}
\inlinepoint{172}{Starting a sequence is much easier than finishing it.} Fable, Sol, Gemini~3.8, Astra, and Opus~5.5 reach CALVIN's first goal in seven or eight of ten episodes, but reach the third in zero, one, four, seven, and four, respectively (Figure~\ref{fig:astra-manipulation}a). Across all models, only nine of eighteen episodes reaching four goals finish the fifth. Aggregate first-goal success therefore hides both early and late attrition.
\inlinepoint{173}{Physical progress can be lost and rebuilt.} The matched CALVIN sequence~0526 reveals one source of additional work. Opus and DeepSeek both finish under the same 40-round allowance, but use eighteen and 37 rounds. DeepSeek remains on block retrieval during rounds~17--27: measured block height returns to about 2\,cm after a carrying attempt, and stacking begins only at round~28. Opus proceeds from retrieval at round~11 to stacking at round~13. The later subgoal inherits the actual object state left by preceding actions; an attempted carry can require renewed retrieval before the sequence advances.
\inlinepoint{174}{Tool position and object state can disagree.} On matched PlaceSphere, DeepSeek brings the gripper within 2.8\,mm of its requested hover target, yet four later checks still return \texttt{is\_grasped=false}; the sphere shifts away from its initial location. Opus instead confirms a grasp, releases the sphere, and obtains the required on-bin and static conditions. Coordinate convergence alone therefore fails to certify transport. Contact-rich insertion imposes further constraints: Opus completes PickCube, PlaceSphere, and StackCube but fails PegInsertionSide and PlugCharger, whose charger criterion requires both 5\,mm positional and 0.2\,rad angular accuracy. These traces locate the missing evidence in grasp, object motion, and final contact conditions (Appendix~\ref{app:trace-diagnostics}).

%% file: sections/further_investigations.tex
\subsection{When Feedback Changes the Outcome}
\label{sec:further}
Feedback can change an answer, make code executable, or help achieve a required state. We distinguish these outcomes before assessing the cost of further interaction.

\textbf{Does reconsideration repair the error that determines acceptance?}
The paired review records distinguish changes in semantics, coordinate validity, and task acceptance (Table~\ref{tab:review}).
\inlinepoint{172}{Corrections compete with regressions.} On eighteen Spatial cases, \mbox{Qwen-27B}'s self-review fixes one wrong answer but changes two correct answers to wrong; crop review fixes two but breaks four. Both reviews change the correct ``back'' to ``left'' when judging which side of a girl faces a car. The net accuracy decrease therefore conceals genuine corrections alongside new reference-frame errors. Reinspection does not consistently preserve already-correct judgments.
\inlinepoint{173}{Valid coordinates are not sufficient grounding.} Six of nine initial Affordance boxes extend beyond image bounds. Self-review repairs all six bounds violations, raising mean IoU from 0.034 to 0.152, yet no box reaches the 0.5 criterion. On a RAGNet scene, the revised point reaches the functional mask, while an entirely out-of-image rectangle becomes an in-image box with IoU~0.447. This is a concrete repair of representation validity, followed by a remaining error in functional extent.
\inlinepoint{174}{Mean improvement and accepted corrections can diverge.} Tool-assisted review produces a lower mean IoU of 0.145 but one accepted box; Astra also gains one accepted box. Thus a broadly improved set of still-insufficient predictions can yield fewer accepted boxes than a lower-mean set. The tool condition adds measurements and a larger response allowance (Appendix~\ref{app:multi-round-review}); its aggregate gain describes that combined procedure. Paired self-/crop-review examples and their exact changes appear in Appendix~\ref{app:trace-diagnostics}.

\textbf{Which feedback repairs the actual execution bottleneck?}
\inlinepoint{172}{Interface compliance consumes interaction opportunities.} \mbox{Qwen-Max} records execution errors in 103/200 CALVIN rounds, including 95 syntax errors; Fable has syntax errors in 34/36 erroneous rounds (Table~\ref{tab:calvin-error-types}). Syntax-invalid responses cannot carry out their intended control program. Treating the resulting failures as physical reasoning errors would overlook opportunities lost before environment interaction.
\inlinepoint{173}{Valid execution can still leave control unresolved.} Among thirty CALVIN episodes with an observed native prefix and no execution error, 24 fail: eleven achieve no subgoal, and thirteen achieve only part of the sequence. Likewise, 62/65 error-containing episodes later execute code successfully, but only three finish (Figure~\ref{fig:astra-manipulation}b). Code repair addresses an interface condition; it does not certify grasp, object placement, or subsequent subgoals. The PlaceSphere checks locate this distinction directly in the object's measured state.
\inlinepoint{174}{Productive recovery restores a prerequisite.} On matched Archaeology Dating, Opus~5.5's first two rounds fail before any environment action. It then acquires tools, excavates three sites, reads artifact ages, and places the flag at the oldest site, completing in round seven; Gemini~4 finishes in eight (Figure~\ref{fig:execution-progress}). DeepSeek reaches score 0.5, but its final three rounds raise code errors without new environment steps. The successful trace connects each correction to evidence needed by the next decision, while the stalled trace receives no further environmental feedback.

\textbf{What does extra computation actually buy?}
\inlinepoint{172}{Repeated inputs can dominate token cost.} In seven HumanCLAW traces, 98.23--99.46\% of reported tokens are prompt tokens, including the inputs processed for successive decisions. Gemini~4's unresolved plant search records 1.95M tokens across 73 executed rounds, with 48 turns and 24 translations; its action descriptions repeatedly report no plant found. The token total principally reflects repeated input processing during unsuccessful search. In other traces the model stops shortly after a sitting attempt despite an unmet native condition. Search allocation and stopping assessment therefore require different diagnoses.
\inlinepoint{173}{The binding budget determines the next opportunity.} DeepSeek's ScienceWorld animal search exhausts decision rounds with 89 environment steps unused. Conversely, Gemini~4's BEHAVIOR-1K toolbox attempt is blocked by the 640k-token ceiling after 37.5 minutes of a 180-minute allowance. More wall time would admit neither next decision under the recorded protocol. Even successful episodes separate costs: on matched chair navigation, Gemini~4 uses seventeen rounds and 158k tokens versus DeepSeek's 65 and 938k, yet takes 437 versus 406 seconds. Progress per decision, token expenditure, and elapsed time must therefore be assessed separately against the resource that actually ends the attempt.

%% file: sections/discussion.tex
\section{Conclusion}
Closed-weight models lead this five-domain evaluation, while open-weight VLMs show useful spatial reasoning but substantially lower Planning and Manipulation success. The deeper distinction lies in the errors behind the scores: scale bias and residual geometric distortion, region enclosure and valid action-point selection, and executable operations and achieved object states require different corrections. During longer sequences, lost physical progress may have to be recovered before the next subgoal becomes feasible.

Feedback is useful when it resolves the particular requirement blocking progress. The matched traces show a receiving-region correction completing an otherwise valid procedure and a confirmed grasp enabling transport; they also show repeated valid code and unsupported stopping judgments leaving goals unmet. Token expenditure alone obscures this distinction. For Astra and Gemini~4, most failed tokens are concentrated in HumanCLAW, whereas unsuccessful attempts remain costly across other sources for the remaining models. Assessing frontier embodied agents therefore requires both complete-task outcomes and evidence of the state changes through which those outcomes are reached.

%% file: sections/appendix.tex
\Needspace{12\baselineskip}
\section*{Appendix: Supporting Evidence and Protocols}
\noindent\textbf{Reading guide.}
Appendices~\ref{app:inventory}--\ref{app:protocol} define task construction, interfaces, and scoring. Appendix~\ref{app:task-results} expands Table~\ref{tab:profile}, including source-composition sensitivity and dispersion for Opus~5.5 and \mbox{DeepSeek-1017}. Appendix~\ref{app:astra-subtasks} follows the five domains of Section~\ref{sec:astra}, from full task comparisons to Manipulation's auxiliary metric definitions. Appendix~\ref{app:setting-comparisons} specifies recorded execution settings and the fixed-input multi-round review study, and Appendix~\ref{app:case-studies} illustrates recorded actions and terminal outcomes, with additional examples across all five capabilities in Appendix~\ref{app:latest-cases}. Appendix~\ref{app:execution-analysis} analyzes shared failures, resources, and recovery using the complete Planning records and linked execution traces. Appendix~\ref{app:benchmark-resources} gives the evaluation scales and public \mbox{resources behind Table~\ref{tab:benchmark-comparison}.}
\input{sections/appendix_inventory}
\input{sections/appendix_protocol}
\input{sections/appendix_comparisons}

\input{sections/appendix_astra_subtasks}

\Needspace*{5\baselineskip}
\input{sections/appendix_setting_comparisons}
\FloatBarrier
\Needspace*{5\baselineskip}
\input{sections/appendix_case_studies}

\Needspace*{5\baselineskip}
\input{sections/appendix_benchmark_resources}

\input{sections/appendix_latest_cases}

\input{sections/appendix_argon}
\input{sections/appendix_execution_analysis}

%% file: sections/appendix_inventory.tex
\section{Benchmark Construction and Source Inventory}
\label{app:inventory}
Embodied Agent Arena contains 1,000 cases from 32 established sources and our GeoProbe benchmark (Table~\ref{tab:inventory}). A case is one query or episode; multiple queries can share a scene. Benchmark-specific allocation uses seed 42 and targets diagnostic coverage of the five capabilities in Section~\ref{sec:construction}.

\input{sections/table_inventory}

\noindent\textbf{Query vocabulary.}
Figure~\ref{fig:construction-statistics}a counts each term once per query, balances weights across source groups and domains, and includes English translations of GeoProbe queries. GeoProbe's ten protocols form one source group.

\noindent\textbf{Geometry (370 cases).}
InFlux-Real~\citep{influx} contributes 30 single-frame intrinsic-estimation cases, of which four have invalid reference annotations and are excluded from continuous-error calculations. Map-free~\citep{mapfree} contributes 30 relative-pose image pairs with known camera intrinsics. MultiSPA-derived uses Map-free images and poses with the camera-motion question formulation associated with Multi-SpatialMLLM~\citep{multispa}: 12 angle, ten distance, and eight displacement-vector questions. This local adaptation retains the Map-free image and pose references. The pose transformation and camera-center displacement use distinct coordinate conventions and reference targets.

GeoProbe comprises ten diagnostic protocols: 168 queries over 130 unique RGB images, predominantly rendered in Blender, with four real-photo displacement cases. Shared scenes separate calibration, pose, depth, displacement, and scale requirements (Table~\ref{tab:geoprobe-inventory}).

TraceSpatial-Bench~\citep{robotracer}, introduced with RoboTracer, adds paired 2D and 3D queries on 56 scenes. These 112 Geometry cases predict offline object trajectories, scored for reference agreement and geometric constraints.

\begin{table}[H]
\centering
\caption{GeoProbe's ten diagnostic protocols, with shared images and scenes.}\label{tab:geoprobe-inventory}
\ArenaTableFont
\begin{tabularx}{\linewidth}{@{}YrYr@{}}
\toprule
Diagnostic & Cases & Diagnostic & Cases\\
\midrule
Object displacement, fixed camera & 12 & Full camera intrinsics & 24\\
Object displacement, small view change & 12 & Focal length, centered principal point & 12\\
Object displacement, large view change & 12 & Relative camera pose & 24\\
Real tabletop displacement & 4 & Sparse depth & 12\\
Object dimensions & 12 & Reference-based scale & 44\\
\bottomrule
\end{tabularx}
\end{table}

\noindent\textbf{Spatial Reasoning (220 cases).}
3DSRBench~\citep{threedsr} contributes 50 relation questions; 28 flip-tagged entries were displayed unflipped. Scores use corrected references for five affected answers under the unflipped presentation. MindCube~\citep{mindcube} contributes 50 multi-view cases: 39 among-view, six around-view, and five rotation. VSI-Bench~\citep{vsi} contributes 50 video questions: nine absolute-distance, nine object-size, fourteen relative-direction, six relative-distance, five counting, four room-size, two appearance-order, and one route-planning. Eight frames form its initial observation. RoboSpatial-Home~\citep{robospatial} contributes ten cases each for compatibility, configuration, and context. BOP-ASK~\citep{bopask} contributes twelve relative-position and eight relative-depth questions. MMSI-Bench~\citep{mmsi} contributes 20 cross-image questions. Together these comprise 183 categorical, 27 numerical, and ten point-set questions, scored as defined in Appendix~\ref{app:scoring}.

\noindent\textbf{Affordance (70 cases).}
UMD~\citep{umd} contributes 30 views spanning five tool classes and six functions: grasp, cut, scoop, contain, pound, and support. Multiple views can depict the same object. The 20 ReasonAff-style cases adapt instruction-conditioned queries motivated by ReasonAff~\citep{reasonaff} to 3DOI~\citep{threedoi} interaction annotations. Another 20 RAGNet-3DOI cases use explicit action-conditioned queries~\citep{ragnet}. The two subsets use disjoint images. These images and annotations support native point and region evaluation.

\noindent\textbf{Task Planning (157 cases).}
ALFRED~\citep{alfred} and ALFWorld~\citep{alfworld} contribute 20 visual cases each, covering placement, heating, cooling, cleaning, containers, multiple objects, and illumination. ScienceWorld~\citep{scienceworld} contributes 73 text-based cases across eighteen topics. VirtualHome~\citep{virtualhome} contributes 21 symbolic programs in seven scenes. The composition analysis removes these 94 cases together. DiscoveryWorld~\citep{discoveryworld} contributes ten tasks with structured observations. HumanCLAW contributes thirteen Habitat navigation and sitting episodes: five sofa, three toilet, two chair, two plant, and one television target. Observations are visual, textual, or symbolic.

\begin{table}[!htbp]
\centering
\caption{Manipulation tasks across 13 sources, with identical assignments for all agents.}
\label{tab:w4inventory}
\ArenaTableFont
\begin{tabularx}{\linewidth}{@{}lrY@{}}
\toprule
Benchmark & Cases & Task coverage\\
\midrule
CALVIN~\citep{calvin} & 10 & Ten sequences, each with five ordered subgoals\\
CLIPort~\citep{cliport} & 18 & Eighteen task and color-generalization configurations\\
RLBench~\citep{rlbench} & 10 & Ten task configurations\\
RoboCasa~\citep{robocasa} & 10 & Seven atomic and three composite tasks\\
RoboCasa365 & 10 & Three atomic and seven composite tasks\\
RoboWits & 10 & Ten official task configurations\\
RoboTwin 2.0~\citep{robotwin} & 50 & Fifty distinct manipulation tasks\\
VIMA-Bench~\citep{vima} & 10 & Six placement-generalization and four new-task cases\\
VLABench~\citep{vlabench} & 25 & 8 ID, 3 category, 4 commonsense, 7 semantic, 3 texture\\
ManiSkill~\citep{maniskill} & 10 & Nine binary tabletop tasks; one quadruped reward task\\
LIBERO-PRO~\citep{capx} & 10 & Ten object, spatial, goal, and multi-object configurations\\
robosuite & 5 & Two lift seeds; stacking, restacking, and wiping\\
BEHAVIOR-1K~\citep{behavior1k} & 5 & Five multi-object household activities\\
\bottomrule
\end{tabularx}
\end{table}

\noindent\textbf{Manipulation (183 cases).}
Table~\ref{tab:w4inventory} lists 182 binary tasks and one continuous-reward task. LIBERO-PRO, robosuite, and BEHAVIOR-1K use benchmark-specific code interfaces~\citep{capx}. RoboCasa365 uses variations designated for pretraining in the source inventory. Appendix~\ref{app:interfaces} specifies the native assistance available to agents.

%% file: sections/table_inventory.tex
\begin{table}[ht]
\centering
\caption{The unified 1,000-case arena, drawn from 32 established sources and our GeoProbe benchmark. Detailed source allocations follow below.}
\label{tab:inventory}
\small
\begin{tabularx}{\linewidth}{@{}lrrY@{}}
\toprule
Domain & Cases & Share (\%) & Task coverage\\
\midrule
Geometry & 370 & 37.0 & Calibration, pose, scale, tracing\\
Spatial Reasoning & 220 & 22.0 & Relations, views, video\\
Affordance & 70 & 7.0 & Functional parts, contact\\
Task Planning & 157 & 15.7 & Household, navigation, science\\
Manipulation & 183 & 18.3 & Robot operations, sequences\\
\midrule
Total & 1,000 & 100.0 & 660 fixed-input; 340 interactive\\
\bottomrule
\end{tabularx}
\end{table}

%% file: sections/appendix_protocol.tex
\section{Evaluation Protocol and Reproducibility}
\label{app:protocol}
\subsection{Observations and Model-Directed Interaction}
\label{app:interfaces}
Geometry, Spatial Reasoning, and Affordance inspect fixed observations; Planning and Manipulation interact with an environment. All domains except Manipulation use a persistent Python runtime; Manipulation uses its native agent loop. Both enforce episode-level budgets and termination. Variables and environment state persist between code rounds, and images produced by a cell reach the next model request after that cell ends.

\begin{table}[!htbp]
\centering
\caption{Model-visible observations, actions, and feedback. Private scoring annotations are unavailable to the agent.}
\ArenaTableFont
\begin{tabularx}{\linewidth}{@{}lYY@{}}
\toprule
Domain & Observation support & Actions and feedback\\
\midrule
Geometry & Image or image pair; intrinsics for Map-free & One final numerical submission\\
Spatial Reasoning & Images, crops, and video-frame queries & Evidence inspection and answer submission\\
Affordance & Image and interaction instruction & Point and box submission; polygons in the added experiment\\
Task Planning & Images, text, or symbolic state & Navigation, interaction, programs, public progress\\
Manipulation & Protocol-specific scene/state observations and native grounding helpers & Motion or skill calls; execution and task checks\\
\bottomrule
\end{tabularx}
\end{table}

ALFRED and visual ALFWorld expose observation, search, navigation, interaction, and memory primitives. CLIPort and VIMA expose instance inspection and pick-and-place actions. RoboTwin provides actor localization, pose inspection, planning probes, and execution. LIBERO-PRO, robosuite, and BEHAVIOR-1K expose benchmark-specific observation and action interfaces; RoboCasa includes fixture and contact localization. The model makes decisions using these native helpers; the baseline adds no specialist perception or verification pipeline. Geometry requests direct visual estimates and excludes pose-solver implementation.

A runtime exception can be followed by another code round if budget remains; prior state changes are not automatically undone. Model requests, code rounds, primitive calls, and environment steps count different units. Stored visualization frames are also distinct from images actually sent to the model. Qualitative figures sample the recorded camera streams.

\subsection{Models and Execution Budgets}
\label{app:settings}
\begin{table}[!htbp]
\centering
\caption{Models and configured identifiers for all ten agents in the five domains.}
\ArenaTableFont
\begin{tabular*}{\linewidth}{@{\extracolsep{\fill}}ll@{}}
\toprule
Model label & Identifier\\
\midrule
GPT-6 Astra & \texttt{gpt-6-astra}\\
GPT-6 Sol & \texttt{gpt-6-sol}\\
Claude Fable 5.1 & \texttt{claude-fable-5-1}\\
Claude Opus~5.5 & \texttt{claude-opus-5-5}\\
DeepSeek-V4.1-Flash-1017 & \texttt{deepseek-flash}\\
Gemini~3.8 Flash & \texttt{gemini-3.8-flash}\\
Gemini~4~Argon & \texttt{gemini-4-argon}\\
Qwen 3.8 Max & \texttt{qwen3.8-max}\\
Qwen 3.5 397B-A17B & \texttt{qwen3.5-397b-a17b}\\
Qwen 3.5 27B & \texttt{qwen3.5-27b}\\
\bottomrule
\end{tabular*}
\end{table}

Gemini~3.8 Flash and Gemini~4~Argon are labeled Gemini~3.8 and Gemini~4 in compact figures and tables. Their recorded API identifiers are \texttt{gemini-3.8-flash} and \texttt{gemini-4-argon}, respectively. Appendix~\ref{app:argon} details Gemini~4~Argon's coverage and execution settings.

Open-weight VLMs comprise \href{https://huggingface.co/Qwen/Qwen3.5-397B-A17B}{Qwen-397B}, \href{https://huggingface.co/Qwen/Qwen3.5-27B}{Qwen-27B}, and \href{https://huggingface.co/deepseek-ai/DeepSeek-V4.1-Flash-1017}{DeepSeek-1017}. This grouping concerns released visual-model weights, not the use of an API for inference. Qwen-Max is evaluated through its visual API: the \href{https://huggingface.co/Qwen/Qwen3.8-2.4T-A95B}{Qwen3.8 open release} provides text-model weights, while vision input is an additional capability of the official service. The evaluated visual service is therefore grouped with the closed-weight models.

\mbox{DeepSeek-1017} (\mbox{DS-1017} in compact figures and tables) denotes the 17 October 2026 release \href{https://api-docs.deepseek.com/updates/}{DeepSeek-V4.1-Flash-1017}, served by \texttt{deepseek-flash}; this alias is not an immutable model snapshot. Its evaluation was recorded on 18--19 October 2026. \mbox{DeepSeek-1017} Spatial runs use no YOLOE, and 3DSRBench inputs use the corrected image flips and matching native labels. The original fixed-input and Planning configurations use temperature zero. Opus~5.5 uses the recorded identifier above; its Manipulation reports retain their own source-specific execution settings. Baseline request limits are 1,800 output tokens and 360 seconds; fixed-input profiles also allow up to 16,384 tokens and 900 seconds. On the configured CLI interface, reasoning effort is maximum for Sol and high for Astra and Fable; API requests omit this parameter. Astra's first repetition on the original 90-case Geometry core uses the high-effort CLI configuration. Manipulation's request settings and recorded execution limits are specified in Appendix~\ref{app:w4-comparison}. 

\begin{table}[!htbp]
\centering
\caption{Baseline episode budgets. Tokens accumulate within each case; rounds are code interactions, not repetitions. Dashes denote unspecified fields. Appendix~\ref{app:latest-results} lists fixed-input settings.}\label{tab:baseline-budgets}
\ArenaTableFont
\begin{tabular*}{\linewidth}{@{\extracolsep{\fill}}lrrrrrr@{}}
\toprule
Entry & Code rounds & Tokens & Env. steps & Primitives & \shortstack{Episode\\seconds} & \shortstack{Task\\seconds}\\
\midrule
Geometry & 12 & 120k & -- & -- & -- & 960\\
Spatial: non-VSI & 6 & 96k & -- & -- & 900 & 960\\
Spatial: VSI & 8 & 240k & -- & -- & 1,200 & 1,500\\
ALFRED, ALFWorld & 8 & 320k & 120 & 540 & 1,200 & 1,500\\
DiscoveryWorld & 8 & 320k & 100 & 400 & 1,200 & 1,500\\
HumanCLAW & 101 & 2,000k & 100 & 400 & 10,800 & 11,000\\
ScienceWorld & 8 & 320k & 100 & 500 & 1,200 & 1,500\\
VirtualHome & 8 & 320k & 16 & 120 & 1,200 & 1,500\\
Affordance & 12 & 120k & -- & -- & -- & 960\\
\bottomrule
\end{tabular*}
\end{table}

Execution and native-call time limits are clipped by remaining budgets. Appendix~\ref{app:w4-comparison} gives Manipulation's recorded execution settings. 

\subsection{Scoring and Repeated Evaluation}
\label{app:scoring}

\noindent\textbf{Submission and termination.}
A final submission commits a fixed-input prediction. Interactive runs end when the agent stops, the environment reaches a terminal condition, a source-specific round, token, action, or time limit is reached, or an API or execution error terminates the episode. The evaluator scores the submitted output or recorded state independently of the agent's declaration. A complete episode includes all its internal reasoning, actions, and recovery; these steps are not separate attempts.

\noindent\textbf{Task scores and aggregation.}
Categorical questions use answer accuracy and interactive tasks use native goal success. Invalid or absent predictions and error-terminated episodes count as unsuccessful within the fixed assignment denominator. Termination reasons distinguish provider errors from code-execution errors and unmet task conditions. Invalid reference annotations are excluded before scoring. Continuous errors average valid predictions with compatible metrics and units within each run. Repeated metrics then weight the run-level scores equally, with valid coverage reported separately. Numerical Spatial Reasoning uses exact-match success alongside continuous errors, with the criterion
\begin{equation*}
|\hat y-y|<10^{-6}.
\end{equation*}
This differs from VSI-Bench's official mean relative accuracy. Point-set questions retain \mbox{their benchmark-specific set criterion.}

\noindent\textbf{Pass@1 and repeated scores.}
For $N$ assigned cases and $R$ repetitions, let $s_{ir}\in\{0,1\}$ denote success on case $i$ in run $r$. The average single-episode success rate is
\begin{equation}
\mathrm{Pass@1}=\frac{1}{R}\sum_{r=1}^{R}\left(\frac{1}{N}\sum_{i=1}^{N}s_{ir}\right).
\end{equation}
This estimates success in one complete episode under the specified condition. Fresh agent state is used for formal repetitions. For any per-run score $M_r$, we compute
\begin{equation}
\bar M=\frac{1}{R}\sum_{r=1}^{R}M_r,\qquad
v_M=\frac{1}{R-1}\sum_{r=1}^{R}(M_r-\bar M)^2,\qquad
\mathrm{SD}_M=\sqrt{v_M}.
\label{eq:repeat-statistics}
\end{equation}
Tables using $\bar M\pm\mathrm{SD}_M$ report variation between run-level scores. Variance is supplied separately in squared metric units. Figure~\ref{fig:astra-geometry} averages valid-case errors within each run and then weights the recorded runs equally. Its coverage sensitivity instead fixes the cases valid in every shared repetition for each model pair; those common-cohort estimates are reported separately. Figure~\ref{fig:astra-spatial}a--b uses per-run categorical accuracies; panel c uses full-source native Pass@1. The original pairwise gaps average accuracy differences over shared run indices, with the comparator fixed per subtask. Run indices align reporting repetitions; seeds and tool configurations can differ. Run-to-run SD requires $R>1$.

\noindent\textbf{Scalar estimates and depth.}
\label{app:geometry}
For $J$ valid targets, the residual, mean absolute error, and root mean squared error are
\begin{equation*}
\begin{aligned}
e_j &= \hat y_j-y_j,\\
\mathrm{MAE} &= \frac{1}{J}\sum_{j=1}^{J}|e_j|,\qquad
\mathrm{RMSE}=\sqrt{\frac{1}{J}\sum_{j=1}^{J}e_j^2}.
\end{aligned}
\end{equation*}
On nonzero targets, absolute relative error and its percentage form are
\begin{equation*}
\mathrm{AbsRel}=\frac{1}{J}\sum_{j=1}^{J}\frac{|e_j|}{|y_j|},\qquad
\mathrm{MAPE}=100\,\mathrm{AbsRel}.
\end{equation*} Size and focal-length estimates use absolute and relative errors. Depth additionally reports RMSE in meters. Depth errors are computed over positive finite predictions within each image, then averaged across images, with valid-point coverage reported. Both the fraction within 10\% relative error and depth threshold accuracy use all queried points as their denominator; invalid predictions contribute zero. The latter is
\begin{equation*}
\delta_1=\frac{1}{J_{\rm all}}\sum_{j=1}^{J_{\rm all}}
\mathbf{1}\!\left[\max\!\left(\frac{\hat y_j}{y_j},\frac{y_j}{\hat y_j}\right)<1.25\right].
\end{equation*} Ordinal accuracy is the fraction of eligible point pairs with the correct depth ordering, with invalid predictions counted as incorrect. Scale alignment removes the global depth factor.

\noindent\textbf{Displacement and camera pose.}
For nonzero scalar or vector $y$, relative motion error is
\begin{equation*}
\frac{\|\hat y-y\|_2}{\|y\|_2}.
\end{equation*} The MultiSPA-derived tolerance is 20\%; a zero target requires zero absolute error. GeoProbe object-displacement diagnostics separately test movement status, direction, absolute distance error, and direction agreement with distance error at most 1\,cm. Stationary cases are evaluated without division by zero. Camera translation and rotation errors are
\begin{equation*}
\begin{aligned}
e_t &= \|\hat{\mathbf t}-\mathbf t\|_2,\\
e_R &= \arccos\!\left(\frac{\mathrm{tr}(R^{\mathsf T}\hat R)-1}{2}\right).
\end{aligned}
\end{equation*}
Translation is reported in centimeters and rotation in degrees, clipping the arccosine argument to $[-1,1]$. GeoProbe reports component errors and joint success within $2^\circ$ and 2\,cm.

\noindent\textbf{Camera calibration.}
Focal MAPE averages the absolute relative errors of $f_x$ and $f_y$ per case. Principal-point error, reported in pixels and also normalized by the image diagonal $D$, is
\begin{equation*}
e_c=\|\hat{\mathbf c}-\mathbf c\|_2.
\end{equation*}
Joint calibration requires both focal components to satisfy 10\% relative error and $e_c\leq0.02D$. Main-table pose errors, complementary continuous metrics, and Figure~\ref{fig:astra-geometry} use equal-run means and run SD, as specified above.

\noindent\textbf{Spatial tracing.}
For each case, predicted and reference paths are independently resampled to $L=100$ equally spaced arc-length positions. Trajectory error is
\begin{equation*}
\mathrm{RMSE}_{\rm trace}=\sqrt{\frac{1}{L}\sum_{\ell=1}^{L}\|\hat{\mathbf p}_\ell-\mathbf p_\ell\|_2^2}.
\end{equation*} We average this per-case value over valid scenes. The 2D metric uses normalized image coordinates; the 3D metric uses meters. The two metrics retain their respective units. The supplied variance is the population variance of the 56 per-scene RMSE values for a single run.

Offline 3D success requires valid initiation, target arrival, and collision checks to pass together. The predicted start must lie within 20\,cm of the object point cloud or the reference start. At least one of the final three predicted points must lie within 20\,cm of the target 3D box; for two-point paths, only the endpoint is checked. Translating the object point cloud to each checked predicted position must yield at most 20\% overlap with the environment. Collision is checked at the sampled positions. Missing or invalid trajectories count as failures. 2D tracing reports continuous error. \mbox{Qwen-27B} trajectory outputs undergo format normalization that preserves all predicted coordinate values.

\noindent\textbf{Native affordance grounding.}
For a directly predicted bounding box $B$ and ground-truth mask $G$, the box interior is rasterized. Box--mask overlap and contact validity for the submitted point $p$ are
\begin{equation*}
\mathrm{IoU}_{B,G}=\frac{|B\cap G|}{|B\cup G|},\qquad
\mathrm{ContactValid}=\mathbf{1}[p\in G].
\end{equation*}
Contact validity uses a reference-box fallback only where a mask is unavailable. A separate diagnostic compares the prediction with the reference box $B_G$:
\begin{equation*}
\mathrm{IoU}_{\rm box}=\frac{|B\cap B_G|}{|B\cup B_G|}.
\end{equation*} Strict box success requires $\mathrm{IoU}_{\rm box}\geq0.5$; contact validity is scored independently. The source field named \texttt{giou} contains ordinary box--mask IoU, not generalized IoU. The separate polygon protocol rasterizes the directly submitted polygons into a region $P$ and scores
\begin{equation*}
\mathrm{IoU}_{P,G}=\frac{|P\cap G|}{|P\cup G|}.
\end{equation*} It adds no external segmenter or iterative segmentation refinement; Appendix~\ref{app:polygon} gives \mbox{output validation and coverage.}

\noindent\textbf{Resource accounting.}
Inference tokens, action calls, model rounds, and elapsed time measure different aspects of execution. The CALVIN recovery analysis counts episodes; detailed traces report actions and elapsed time. Token fractions use reported input and output usage, excluding missing values and unknown transport or retry usage. Episode duration includes worker setup, retries, and teardown for the Qwen HumanCLAW records; other durations use the reported episode total or last loop time. Request, trial, and native-call time limits are specified separately in Appendix~\ref{app:w4-comparison}.

\subsection{Repetition Protocol and Result Accounting}
\label{app:latest-results}
\noindent\textbf{Repetitions.}
Non-tracing Geometry, Spatial Reasoning, and point-and-box Affordance use five repetitions per case for nine models and three for Fable. Tracing, Planning, Manipulation, and direct-polygon prediction use one run per assignment for all ten models. Run-to-run SD is reported only for repeated measurements. Manipulation's observation schedules appear in Appendices~\ref{app:w4-comparison} and~\ref{app:deepseek}.

\noindent\textbf{Additional execution configurations.}
Fixed-input evaluation also uses per-case limits of up to 48 code rounds and 480k tokens for Geometry and Affordance, 24 rounds and 384k tokens for non-VSI Spatial Reasoning, and 32 rounds and 960k tokens for VSI. Task time limits reach 7,200 seconds, or 10,800 seconds for VSI. Budgets vary by source and execution configuration.

\noindent\textbf{Evaluation records.}
Each case--run slot contributes one designated evaluation record, selected independently of task success. Historical retries are not additional repetitions. Table~\ref{tab:profile} uses the fixed ten-model evaluation population; supplementary case recordings remain separate where they represent different episodes. Source-specific observation schedules, execution ceilings, and missing-record accounting are specified in Appendix~\ref{app:w4-comparison}.

\noindent\textbf{Coverage and scoring denominators.}
Table~\ref{tab:coverage} reports submission coverage for all ten agents; Table~\ref{tab:opus-dispersion} gives Opus~5.5's metric-specific valid counts. HumanCLAW success uses the native navigation or sitting criterion. \mbox{Qwen-Max} and \mbox{Qwen-397B} each complete two of thirteen assignments; \mbox{Qwen-27B} completes none. Two \mbox{Qwen-Max} sitting assignments end in provider errors and count as unsuccessful episodes. Their partial process observations remain available for diagnosis. Opus~5.5 Manipulation counts two RLBench exceptions and one RoboTwin exception as unsuccessful, yielding 87/182 (47.8\%). All models retain 182 binary Manipulation assignments, including five BEHAVIOR-1K tasks; AnymalC-Spin is evaluated separately by continuous return. InFlux excludes four invalid references, and continuous metrics average finite valid predictions.
\input{sections/table_coverage}

\noindent\textbf{Manipulation diagnostics.}
CALVIN reports average consecutive goals completed. Other auxiliary metrics average task-normalized native criteria or control evidence over fixed applicable subsets. They differ in their scoring items and evidence time windows; Appendix~\ref{app:w4-metrics} defines the aggregation, coverage, and missing-record treatment.

\noindent\textbf{Main-table metric scope.}
Geometry rotation and translation average valid predictions from the 30 Map-free and 24 GeoProbe pose cases within each run, in degrees and centimeters; the main table reports the mean and sample SD across runs. Spatial Pass@1 covers all 220 cases; numerical AbsRel covers 27, and categorical accuracy separately covers 183. Original Affordance box--mask IoU and contact validity cover all three sources, with 70 cases. Polygon IoU evaluates the same allocation under the direct-polygon protocol. Planning uses terminal success across 157 cases for every model. Manipulation uses task-weighted native success across 182 binary assignments; Tables~\ref{tab:allmanip} and~\ref{tab:opus-sources} report each source separately.

\noindent\textbf{Dispersion and valid subsets.}
Continuous metrics first average valid predictions within each run, then report the equal-weight mean and sample SD across runs (Equation~\ref{eq:repeat-statistics}). This SD measures variation between repeated evaluations, not variation among individual cases. Common-valid-cohort checks appear in Appendices~\ref{app:geometry-focus} and~\ref{app:spatial-focus}.

%% file: sections/table_coverage.tex
\begin{table}[!htbp]\centering\ArenaTableFont
\caption{Submission coverage for all ten agents. Fixed-input columns count valid submissions over assigned slots, including repetitions and invalid-reference slots; invalid references are excluded from scoring. Planning counts outcome records over assignments; provider-error terminations count as unsuccessful. Geometry includes 112 single-run tracing cases.}\label{tab:coverage}
\begin{tabular*}{\linewidth}{@{\extracolsep{\fill}}lrrrr@{}}\toprule
Model & Geometry & Spatial & Affordance & Planning\\\midrule
Astra & 1382/1402 & 1060/1100 & 350/350 & 157/157\\
Sol & 1382/1402 & 1100/1100 & 350/350 & 157/157\\
Fable & 819/886 & 586/660 & 208/210 & 157/157\\
Q-Max & 1382/1402 & 1097/1100 & 350/350 & 157/157\\
Gemini~3.8 & 1314/1402 & 1077/1100 & 335/350 & 157/157\\
Q-397B & 1382/1402 & 1033/1100 & 349/350 & 157/157\\
Q-27B & 1376/1402 & 982/1100 & 349/350 & 157/157\\
Opus~5.5 & 1374/1402 & 1075/1100 & 350/350 & 157/157\\
DS-1017 & 1356/1402 & 954/1100 & 348/350 & 157/157\\
Gemini~4 & 1376/1402 & 1003/1100 & 350/350 & 157/157\\
\bottomrule\end{tabular*}\end{table}

%% file: sections/appendix_comparisons.tex
\section{Complete Model Comparisons}
\label{app:task-results}
\label{app:counts}
This section supports Section~\ref{sec:cross-model} at three resolutions: complementary native metrics, differences among source benchmarks, and the metric profiles shown in Figure~\ref{fig:capability-cost}. Task-level explanations follow in Appendix~\ref{app:astra-subtasks}.

\subsection{Metrics and Source-Level Results}
Table~\ref{tab:profile-full} gives complementary native metrics. Table~\ref{tab:source-comparisons} resolves all ten models by source, using the denominators and repetitions in Appendix~\ref{app:latest-results}. These breakdowns distinguish broad domain performance from strengths concentrated in particular task collections.
\noindent\textbf{Reading the metric families.}
GeoProbe focal MAPE combines full-intrinsics and focal-only queries; displacement MAE combines its four camera/environment conditions. Scale MAPE combines object dimensions and numerical reference-scale queries, while qualitative scale judgments retain Pass@1. TraceSpatial reports 2D and 3D errors separately because their coordinates and units differ. Each metric retains the target's physical interpretation.

\input{sections/table_profile_full}
\input{sections/table_source_comparisons}

\FloatBarrier
\subsection{Sensitivity to Source Composition}
\label{app:sensitivity}
\input{sections/table_composition}
Removing ScienceWorld and VirtualHome changes Astra's Planning success from 80.3\% to 66.7\%, and \mbox{Qwen-Max}'s from 36.9\% to 15.9\%. The same table isolates categorical Spatial Reasoning without VSI-Bench and strict affordance-box success without UMD. These subset comparisons measure sensitivity to the task mixture.

\noindent\textbf{Equal source weighting.}
To measure sensitivity to source allocation across all ten agents, we average complete source scores with equal weight:
\begin{equation*}
M_{\mathrm{source}}=\frac{1}{B}\sum_{b=1}^{B}M_b.
\end{equation*}
Every assigned case is retained within each source. Gemini~4~Argon leads equal-source Spatial accuracy at 73.3\%, ahead of Astra at 71.9\%. Astra retains the strongest equal-source Planning score (75.2\%), followed by Gemini~4~Argon (65.3\%). For contact validity, Opus~5.5 and Gemini~4~Argon score 55.3\% and 54.2\%, respectively. Removing the two largest Planning sources leaves Gemini~4~Argon at 60.3\%, versus Astra's 66.7\%; removing UMD leaves Gemini~4~Argon at 28.0\% strict box success. These models therefore retain substantial strengths outside the largest collections, while task composition changes the size of the observed gaps. These analyses use the complete domain allocations; the 46-case review diagnostic in Section~\ref{sec:further} uses a separate selected subset.

\input{sections/appendix_opus_statistics}
\input{sections/appendix_deepseek}
\input{sections/appendix_polygon}
\input{sections/appendix_cost_profile}

%% file: sections/table_profile_full.tex
{\ArenaDenseTableFont
\setlength{\tabcolsep}{1.6pt}
\renewcommand{\arraystretch}{1.08}
\begin{longtable}{@{}lr*{10}{r}@{}}
\caption{Complementary native metrics across four domains; \mbox{DS-1017} abbreviates DeepSeek-V4.1-Flash-1017. Repeated metrics show the mean above $\pm$ run SD (five runs; three for Fable). Each run averages valid continuous predictions or scores all assigned binary cases; runs receive equal weight. Polygon IoU uses a separate single run for all ten models. Planning uses all 157 cases for every model. $N$ counts eligible cases. GP denotes GeoProbe; Manipulation source results appear in Tables~\ref{tab:allmanip} and~\ref{tab:opus-sources}.}\label{tab:profile-full}\\
\toprule
Metric & $N$ & Astra & Sol & Fable & Q-Max & Gemini~3.8 & Q-397B & Q-27B & Opus~5.5 & \mbox{DS-1017} & Gemini~4\\\midrule\endfirsthead
\multicolumn{12}{l}{\textit{Complementary metrics, continued.}}\\\toprule
Metric & $N$ & Astra & Sol & Fable & Q-Max & Gemini~3.8 & Q-397B & Q-27B & Opus~5.5 & \mbox{DS-1017} & Gemini~4\\\midrule\endhead
\bottomrule\endfoot
\multicolumn{12}{@{}l}{\textit{Geometry (370 cases)}}\\*
InFlux focal MAPE (\%)$\downarrow$ & 26 & \shortstack{53.02\\$\pm2.15$} & \shortstack{47.79\\$\pm2.00$} & \shortstack{52.65\\$\pm2.22$} & \shortstack{55.50\\$\pm1.04$} & \shortstack{52.40\\$\pm3.40$} & \shortstack{67.82\\$\pm1.82$} & \shortstack{65.86\\$\pm2.08$} & \shortstack{60.27\\$\pm1.49$} & \shortstack{62.92\\$\pm1.57$} & \shortstack{34.84\\$\pm1.63$}\\
InFlux principal error (px)$\downarrow$ & 26 & \shortstack{3.82\\$\pm0.00$} & \shortstack{3.82\\$\pm0.00$} & \shortstack{3.82\\$\pm0.00$} & \shortstack{3.82\\$\pm0.00$} & \shortstack{3.88\\$\pm0.09$} & \shortstack{3.82\\$\pm0.00$} & \shortstack{3.82\\$\pm0.00$} & \shortstack{3.82\\$\pm0.00$} & \shortstack{3.55\\$\pm0.46$} & \shortstack{3.82\\$\pm0.00$}\\
Angle relative L2$\downarrow$ & 12 & \shortstack{0.760\\$\pm0.130$} & \shortstack{0.915\\$\pm0.442$} & \shortstack{1.318\\$\pm0.046$} & \shortstack{1.410\\$\pm0.281$} & \shortstack{0.961\\$\pm0.083$} & \shortstack{1.298\\$\pm0.172$} & \shortstack{2.037\\$\pm0.331$} & \shortstack{1.166\\$\pm0.113$} & \shortstack{1.497\\$\pm0.352$} & \shortstack{0.735\\$\pm0.046$}\\
Distance relative L2$\downarrow$ & 10 & \shortstack{0.294\\$\pm0.024$} & \shortstack{0.418\\$\pm0.130$} & \shortstack{0.785\\$\pm0.115$} & \shortstack{0.560\\$\pm0.090$} & \shortstack{0.419\\$\pm0.077$} & \shortstack{0.411\\$\pm0.059$} & \shortstack{0.331\\$\pm0.095$} & \shortstack{0.455\\$\pm0.069$} & \shortstack{0.461\\$\pm0.196$} & \shortstack{0.313\\$\pm0.017$}\\
Vector relative L2$\downarrow$ & 8 & \shortstack{0.425\\$\pm0.092$} & \shortstack{0.591\\$\pm0.155$} & \shortstack{0.625\\$\pm0.094$} & \shortstack{0.454\\$\pm0.122$} & \shortstack{0.643\\$\pm0.071$} & \shortstack{1.018\\$\pm0.080$} & \shortstack{1.031\\$\pm0.060$} & \shortstack{0.516\\$\pm0.042$} & \shortstack{0.743\\$\pm0.254$} & \shortstack{0.529\\$\pm0.120$}\\
Map-free rotation ($^\circ$)$\downarrow$ & 30 & \shortstack{30.08\\$\pm12.15$} & \shortstack{46.60\\$\pm2.25$} & \shortstack{50.81\\$\pm8.99$} & \shortstack{35.11\\$\pm6.25$} & \shortstack{45.19\\$\pm7.64$} & \shortstack{63.03\\$\pm1.61$} & \shortstack{64.94\\$\pm6.89$} & \shortstack{40.95\\$\pm8.10$} & \shortstack{52.68\\$\pm6.56$} & \shortstack{31.75\\$\pm3.52$}\\
Map-free translation (cm)$\downarrow$ & 30 & \shortstack{188.98\\$\pm65.21$} & \shortstack{265.85\\$\pm10.41$} & \shortstack{248.74\\$\pm50.12$} & \shortstack{243.06\\$\pm18.41$} & \shortstack{272.95\\$\pm59.81$} & \shortstack{430.82\\$\pm18.13$} & \shortstack{374.96\\$\pm22.44$} & \shortstack{243.13\\$\pm35.82$} & \shortstack{324.19\\$\pm46.48$} & \shortstack{185.06\\$\pm14.78$}\\
Map-free direction ($^\circ$)$\downarrow$ & 30 & \shortstack{26.43\\$\pm12.73$} & \shortstack{48.91\\$\pm2.14$} & \shortstack{54.20\\$\pm7.11$} & \shortstack{38.92\\$\pm5.54$} & \shortstack{50.72\\$\pm14.17$} & \shortstack{95.15\\$\pm6.59$} & \shortstack{78.78\\$\pm9.15$} & \shortstack{44.29\\$\pm9.06$} & \shortstack{64.63\\$\pm6.45$} & \shortstack{27.47\\$\pm2.95$}\\
GP focal MAPE (\%)$\downarrow$ & 36 & \shortstack{6.92\\$\pm4.90$} & \shortstack{9.57\\$\pm1.25$} & \shortstack{31.77\\$\pm1.05$} & \shortstack{14.67\\$\pm2.72$} & \shortstack{26.59\\$\pm4.73$} & \shortstack{33.39\\$\pm2.99$} & \shortstack{38.56\\$\pm2.95$} & \shortstack{16.48\\$\pm6.13$} & \shortstack{33.99\\$\pm5.33$} & \shortstack{28.39\\$\pm0.91$}\\
GP principal error (px)$\downarrow$ & 24 & \shortstack{51.95\\$\pm8.15$} & \shortstack{110.62\\$\pm18.22$} & \shortstack{62.81\\$\pm8.54$} & \shortstack{78.41\\$\pm13.77$} & \shortstack{66.83\\$\pm9.70$} & \shortstack{58.23\\$\pm1.02$} & \shortstack{59.05\\$\pm1.01$} & \shortstack{60.46\\$\pm4.13$} & \shortstack{75.40\\$\pm9.96$} & \shortstack{56.05\\$\pm1.89$}\\
Depth MAE (m)$\downarrow$ & 12 & \shortstack{0.307\\$\pm0.089$} & \shortstack{0.633\\$\pm0.068$} & \shortstack{0.416\\$\pm0.088$} & \shortstack{0.490\\$\pm0.042$} & \shortstack{0.587\\$\pm0.067$} & \shortstack{0.934\\$\pm0.072$} & \shortstack{0.945\\$\pm0.043$} & \shortstack{0.510\\$\pm0.295$} & \shortstack{0.688\\$\pm0.089$} & \shortstack{0.927\\$\pm0.024$}\\
Depth RMSE (m)$\downarrow$ & 12 & \shortstack{0.338\\$\pm0.087$} & \shortstack{0.753\\$\pm0.086$} & \shortstack{0.522\\$\pm0.093$} & \shortstack{0.598\\$\pm0.041$} & \shortstack{0.697\\$\pm0.064$} & \shortstack{1.053\\$\pm0.080$} & \shortstack{1.059\\$\pm0.044$} & \shortstack{0.599\\$\pm0.302$} & \shortstack{0.814\\$\pm0.122$} & \shortstack{0.973\\$\pm0.024$}\\
Depth AbsRel$\downarrow$ & 12 & \shortstack{0.104\\$\pm0.029$} & \shortstack{0.229\\$\pm0.027$} & \shortstack{0.142\\$\pm0.027$} & \shortstack{0.163\\$\pm0.013$} & \shortstack{0.192\\$\pm0.021$} & \shortstack{0.312\\$\pm0.028$} & \shortstack{0.292\\$\pm0.013$} & \shortstack{0.169\\$\pm0.108$} & \shortstack{0.229\\$\pm0.025$} & \shortstack{0.299\\$\pm0.006$}\\
Displacement MAE (cm)$\downarrow$ & 40 & \shortstack{0.585\\$\pm0.092$} & \shortstack{0.973\\$\pm0.056$} & \shortstack{1.488\\$\pm0.163$} & \shortstack{1.515\\$\pm0.160$} & \shortstack{2.278\\$\pm0.170$} & \shortstack{2.480\\$\pm0.071$} & \shortstack{2.935\\$\pm0.082$} & \shortstack{1.164\\$\pm0.475$} & \shortstack{1.934\\$\pm0.318$} & \shortstack{1.309\\$\pm0.050$}\\
GP rotation ($^\circ$)$\downarrow$ & 24 & \shortstack{1.17\\$\pm0.38$} & \shortstack{2.54\\$\pm1.02$} & \shortstack{2.80\\$\pm0.94$} & \shortstack{2.21\\$\pm0.18$} & \shortstack{3.10\\$\pm0.48$} & \shortstack{7.55\\$\pm0.77$} & \shortstack{5.26\\$\pm0.97$} & \shortstack{3.05\\$\pm0.96$} & \shortstack{3.75\\$\pm0.57$} & \shortstack{2.71\\$\pm0.36$}\\
GP translation (cm)$\downarrow$ & 24 & \shortstack{5.36\\$\pm2.11$} & \shortstack{9.50\\$\pm2.71$} & \shortstack{9.24\\$\pm2.05$} & \shortstack{6.99\\$\pm0.96$} & \shortstack{7.85\\$\pm0.90$} & \shortstack{23.59\\$\pm1.80$} & \shortstack{12.05\\$\pm0.52$} & \shortstack{11.13\\$\pm1.48$} & \shortstack{14.52\\$\pm1.80$} & \shortstack{10.14\\$\pm0.98$}\\
Scale MAPE (\%)$\downarrow$ & 44 & \shortstack{1.90\\$\pm0.28$} & \shortstack{2.84\\$\pm0.21$} & \shortstack{3.76\\$\pm0.05$} & \shortstack{3.77\\$\pm0.16$} & \shortstack{3.93\\$\pm0.46$} & \shortstack{16.07\\$\pm1.73$} & \shortstack{16.45\\$\pm0.48$} & \shortstack{3.91\\$\pm2.74$} & \shortstack{7.62\\$\pm1.13$} & \shortstack{2.21\\$\pm0.09$}\\
Scale judgment Pass@1 (\%)$\uparrow$ & 12 & \shortstack{100.0\\$\pm0.0$} & \shortstack{100.0\\$\pm0.0$} & \shortstack{100.0\\$\pm0.0$} & \shortstack{100.0\\$\pm0.0$} & \shortstack{100.0\\$\pm0.0$} & \shortstack{100.0\\$\pm0.0$} & \shortstack{93.3\\$\pm7.0$} & \shortstack{100.0\\$\pm0.0$} & \shortstack{98.3\\$\pm3.7$} & \shortstack{100.0\\$\pm0.0$}\\
2D trace RMSE$\downarrow$ & 56 & 0.120 & 0.132 & 0.118 & 0.118 & 0.127 & 0.155 & 0.284 & 0.180 & 0.141 & 0.107\\
3D trace RMSE (m)$\downarrow$ & 56 & 0.382 & 0.463 & 0.435 & 0.469 & 0.426 & 0.629 & 0.682 & 0.683 & 0.470 & 0.310\\
3D constraint success (\%)$\uparrow$ & 56 & 33.9 & 21.4 & 25.0 & 17.9 & 37.5 & 8.9 & 7.1 & 8.9 & 14.3 & 42.9\\
\midrule
\multicolumn{12}{@{}l}{\textit{Spatial Reasoning (220 cases)}}\\*
Native Pass@1 (\%)$\uparrow$ & 220 & \shortstack{69.4\\$\pm1.0$} & \shortstack{60.7\\$\pm2.5$} & \shortstack{62.0\\$\pm4.6$} & \shortstack{68.9\\$\pm1.1$} & \shortstack{65.4\\$\pm1.3$} & \shortstack{50.6\\$\pm1.8$} & \shortstack{48.0\\$\pm0.9$} & \shortstack{67.5\\$\pm5.1$} & \shortstack{51.6\\$\pm1.5$} & \shortstack{71.6\\$\pm0.7$}\\
Categorical Pass@1 (\%)$\uparrow$ & 183 & \shortstack{82.2\\$\pm0.6$} & \shortstack{71.3\\$\pm3.0$} & \shortstack{72.7\\$\pm5.2$} & \shortstack{81.2\\$\pm1.3$} & \shortstack{76.7\\$\pm1.9$} & \shortstack{60.0\\$\pm2.2$} & \shortstack{56.8\\$\pm0.7$} & \shortstack{79.2\\$\pm5.6$} & \shortstack{61.3\\$\pm1.7$} & \shortstack{84.2\\$\pm1.3$}\\
Numerical AbsRel$\downarrow$ & 27 & \shortstack{0.214\\$\pm0.016$} & \shortstack{0.273\\$\pm0.016$} & \shortstack{0.185\\$\pm0.011$} & \shortstack{0.206\\$\pm0.033$} & \shortstack{0.161\\$\pm0.028$} & \shortstack{0.251\\$\pm0.027$} & \shortstack{0.288\\$\pm0.023$} & \shortstack{0.184\\$\pm0.040$} & \shortstack{0.309\\$\pm0.035$} & \shortstack{0.169\\$\pm0.023$}\\
\midrule
\multicolumn{12}{@{}l}{\textit{Affordance (70 cases)}}\\*
BBox--GT box IoU$\uparrow$ & 70 & \shortstack{0.390\\$\pm0.005$} & \shortstack{0.366\\$\pm0.012$} & \shortstack{0.386\\$\pm0.014$} & \shortstack{0.382\\$\pm0.022$} & \shortstack{0.324\\$\pm0.098$} & \shortstack{0.146\\$\pm0.015$} & \shortstack{0.021\\$\pm0.008$} & \shortstack{0.432\\$\pm0.021$} & \shortstack{0.278\\$\pm0.015$} & \shortstack{0.484\\$\pm0.012$}\\
BBox--GT mask IoU$\uparrow$ & 70 & \shortstack{0.286\\$\pm0.010$} & \shortstack{0.264\\$\pm0.010$} & \shortstack{0.284\\$\pm0.013$} & \shortstack{0.272\\$\pm0.017$} & \shortstack{0.251\\$\pm0.081$} & \shortstack{0.116\\$\pm0.014$} & \shortstack{0.013\\$\pm0.006$} & \shortstack{0.311\\$\pm0.020$} & \shortstack{0.217\\$\pm0.011$} & \shortstack{0.356\\$\pm0.009$}\\
Polygon--GT mask IoU$\uparrow$ & 70 & 0.379 & 0.325 & 0.328 & 0.307 & 0.291 & 0.092 & 0.051 & 0.397 & 0.249 & 0.441\\
BBox success@0.5 (\%)$\uparrow$ & 70 & \shortstack{46.6\\$\pm1.3$} & \shortstack{43.7\\$\pm1.6$} & \shortstack{46.2\\$\pm0.8$} & \shortstack{45.7\\$\pm1.7$} & \shortstack{33.1\\$\pm11.8$} & \shortstack{5.4\\$\pm3.7$} & \shortstack{0.3\\$\pm0.6$} & \shortstack{51.4\\$\pm3.0$} & \shortstack{28.9\\$\pm3.7$} & \shortstack{56.3\\$\pm2.2$}\\
Contact-point validity (\%)$\uparrow$ & 70 & \shortstack{60.3\\$\pm2.6$} & \shortstack{52.0\\$\pm3.6$} & \shortstack{57.1\\$\pm2.5$} & \shortstack{52.9\\$\pm4.2$} & \shortstack{44.6\\$\pm13.0$} & \shortstack{32.6\\$\pm4.7$} & \shortstack{6.3\\$\pm1.6$} & \shortstack{61.1\\$\pm1.9$} & \shortstack{50.9\\$\pm4.2$} & \shortstack{59.7\\$\pm1.2$}\\
\midrule
\multicolumn{12}{@{}l}{\textit{Task Planning (157 cases)}}\\*
Native Pass@1 (\%)$\uparrow$ & 157 & 80.3 & 44.6 & 63.7 & 36.9 & 48.4 & 19.1 & 19.1 & 48.4 & 31.2 & 70.1\\
\end{longtable}
}

%% file: sections/table_source_comparisons.tex
{\ArenaDenseTableFont
\setlength{\tabcolsep}{1.6pt}
\begin{longtable}{@{}lr*{10}{r}@{}}
\caption{Source results for all ten agents. Gemini~4 denotes Gemini~4~Argon and Gemini~3.8 denotes Gemini~3.8~Flash. Spatial/Affordance: mean above $\pm$ run SD (\%); Planning: solved/assigned cases; Manipulation: success (\%). $N$ counts eligible cases.}
\label{tab:source-comparisons}\label{tab:allmanip}\\
\toprule
Source & $N$ & Astra & Sol & Fable & Q-Max & Gemini~3.8 & Q-397B & Q-27B & Opus~5.5 & DS-1017 & Gemini~4\\\midrule\endfirsthead
\multicolumn{12}{l}{\textit{Source-level comparisons, continued.}}\\\toprule
Source & $N$ & Astra & Sol & Fable & Q-Max & Gemini~3.8 & Q-397B & Q-27B & Opus~5.5 & DS-1017 & Gemini~4\\\midrule\endhead
\bottomrule\endfoot
\multicolumn{12}{l}{\textit{Spatial Reasoning: categorical accuracy (\%)}}\\*
3DSRBench & 50 & \shortstack{73.6\\$\pm2.6$} & \shortstack{71.2\\$\pm1.8$} & \shortstack{72.7\\$\pm3.1$} & \shortstack{74.0\\$\pm3.7$} & \shortstack{70.0\\$\pm2.4$} & \shortstack{69.2\\$\pm3.6$} & \shortstack{57.6\\$\pm2.6$} & \shortstack{$76.8$\\$\pm9.9$} & \shortstack{61.6\\$\pm7.1$} & \shortstack{$88.0$\\$\pm1.4$}\\
BOP-ASK & 20 & \shortstack{96.0\\$\pm4.2$} & \shortstack{87.0\\$\pm4.5$} & \shortstack{88.3\\$\pm2.9$} & \shortstack{92.0\\$\pm2.7$} & \shortstack{97.0\\$\pm2.7$} & \shortstack{85.0\\$\pm3.5$} & \shortstack{84.0\\$\pm4.2$} & \shortstack{$95.0$\\$\pm0.0$} & \shortstack{88.0\\$\pm7.6$} & \shortstack{$98.0$\\$\pm2.7$}\\
MindCube & 50 & \shortstack{90.8\\$\pm1.1$} & \shortstack{68.0\\$\pm7.3$} & \shortstack{79.3\\$\pm8.1$} & \shortstack{90.8\\$\pm2.7$} & \shortstack{89.2\\$\pm3.0$} & \shortstack{46.0\\$\pm2.8$} & \shortstack{50.4\\$\pm4.1$} & \shortstack{$85.2$\\$\pm6.7$} & \shortstack{54.4\\$\pm4.3$} & \shortstack{$93.2$\\$\pm1.8$}\\
MMSI-Bench & 20 & \shortstack{77.0\\$\pm5.7$} & \shortstack{55.0\\$\pm11.7$} & \shortstack{66.7\\$\pm7.6$} & \shortstack{55.0\\$\pm8.7$} & \shortstack{56.0\\$\pm9.6$} & \shortstack{35.0\\$\pm9.4$} & \shortstack{29.0\\$\pm7.4$} & \shortstack{$59.0$\\$\pm8.2$} & \shortstack{53.0\\$\pm5.7$} & \shortstack{$72.0$\\$\pm5.7$}\\
RoboSpatial & 20 & \shortstack{85.0\\$\pm3.5$} & \shortstack{81.0\\$\pm8.9$} & \shortstack{88.3\\$\pm2.9$} & \shortstack{88.0\\$\pm5.7$} & \shortstack{76.0\\$\pm4.2$} & \shortstack{90.0\\$\pm3.5$} & \shortstack{92.0\\$\pm5.7$} & \shortstack{$92.0$\\$\pm6.7$} & \shortstack{87.0\\$\pm5.7$} & \shortstack{$85.0$\\$\pm3.5$}\\
VSI-Bench & 23 & \shortstack{72.2\\$\pm7.3$} & \shortstack{70.4\\$\pm5.7$} & \shortstack{36.2\\$\pm28.9$} & \shortstack{83.5\\$\pm8.4$} & \shortstack{65.2\\$\pm5.3$} & \shortstack{44.3\\$\pm6.4$} & \shortstack{39.1\\$\pm3.1$} & \shortstack{$64.3$\\$\pm3.6$} & \shortstack{37.4\\$\pm5.0$} & \shortstack{$53.9$\\$\pm9.0$}\\
\midrule
\multicolumn{12}{l}{\textit{Affordance: strict box success at IoU $\geq0.5$ (\%)}}\\*
ReasonAff-style & 20 & \shortstack{19.0\\$\pm5.5$} & \shortstack{6.0\\$\pm4.2$} & \shortstack{16.7\\$\pm7.6$} & \shortstack{13.0\\$\pm5.7$} & \shortstack{5.0\\$\pm0.0$} & \shortstack{0.0\\$\pm0.0$} & \shortstack{0.0\\$\pm0.0$} & \shortstack{$20.0$\\$\pm6.1$} & \shortstack{3.0\\$\pm4.5$} & \shortstack{$24.0$\\$\pm2.2$}\\
RAGNet-3DOI & 20 & \shortstack{14.0\\$\pm4.2$} & \shortstack{15.0\\$\pm5.0$} & \shortstack{13.3\\$\pm2.9$} & \shortstack{7.0\\$\pm2.7$} & \shortstack{6.0\\$\pm2.2$} & \shortstack{4.0\\$\pm2.2$} & \shortstack{1.0\\$\pm2.2$} & \shortstack{$18.0$\\$\pm4.5$} & \shortstack{5.0\\$\pm6.1$} & \shortstack{$32.0$\\$\pm5.7$}\\
UMD & 30 & \shortstack{86.7\\$\pm3.3$} & \shortstack{88.0\\$\pm4.5$} & \shortstack{87.8\\$\pm5.1$} & \shortstack{93.3\\$\pm2.4$} & \shortstack{70.0\\$\pm27.0$} & \shortstack{10.0\\$\pm7.8$} & \shortstack{0.0\\$\pm0.0$} & \shortstack{$94.7$\\$\pm1.8$} & \shortstack{62.0\\$\pm6.9$} & \shortstack{$94.0$\\$\pm2.8$}\\
\midrule
\multicolumn{12}{l}{\textit{Task Planning: terminal success (single run)}}\\*
ALFRED & 20 & 15/20 & 3/20 & 7/20 & 3/20 & 2/20 & 1/20 & 0/20 & 3/20 & 1/20 & 14/20\\
ALFWorld & 20 & 14/20 & 4/20 & 6/20 & 2/20 & 4/20 & 3/20 & 4/20 & 9/20 & 4/20 & 16/20\\
DiscoveryWorld & 10 & 10/10 & 4/10 & 7/10 & 3/10 & 5/10 & 1/10 & 1/10 & 3/10 & 2/10 & 6/10\\
HumanCLAW & 13 & 3/13 & 2/13 & 1/13 & 2/13 & 3/13 & 2/13 & 0/13 & 3/13 & 2/13 & 2/13\\
ScienceWorld & 73 & 64/73 & 38/73 & 59/73 & 29/73 & 45/73 & 11/73 & 14/73 & 43/73 & 28/73 & 52/73\\
VirtualHome & 21 & 20/21 & 19/21 & 20/21 & 19/21 & 17/21 & 12/21 & 11/21 & 15/21 & 12/21 & 20/21\\
\midrule
Total & 157 & 126/157 & 70/157 & 100/157 & 58/157 & 76/157 & 30/157 & 30/157 & 76/157 & 49/157 & 110/157\\
\midrule
\multicolumn{12}{l}{\textit{Manipulation: task success (\%)}}\\*
CALVIN & 10 & 30.0 & 0.0 & 0.0 & 0.0 & 0.0 & 0.0 & 0.0 & 10.0 & 10.0 & 40.0\\
CLIPort & 18 & 61.1 & 27.8 & 33.3 & 22.2 & 27.8 & 22.2 & 16.7 & 61.1 & 33.3 & 38.9\\
RLBench & 10 & 60.0 & 20.0 & 10.0 & 0.0 & 20.0 & 0.0 & 10.0 & 30.0 & 20.0 & 80.0\\
RoboCasa & 10 & 10.0 & 0.0 & 0.0 & 0.0 & 0.0 & 0.0 & 0.0 & 0.0 & 0.0 & 20.0\\
RoboCasa365 & 10 & 0.0 & 0.0 & 10.0 & 0.0 & 0.0 & 0.0 & 0.0 & 10.0 & 0.0 & 10.0\\
RoboTwin 2.0 & 50 & 38.0 & 10.0 & 14.0 & 4.0 & 8.0 & 6.0 & 8.0 & 46.0 & 4.0 & 28.0\\
RoboWits & 10 & 10.0 & 10.0 & 10.0 & 10.0 & 20.0 & 0.0 & 10.0 & 10.0 & 20.0 & 10.0\\
VIMA-Bench & 10 & 40.0 & 30.0 & 30.0 & 10.0 & 20.0 & 30.0 & 30.0 & 100.0 & 60.0 & 90.0\\
VLABench & 25 & 72.0 & 52.0 & 60.0 & 20.0 & 36.0 & 24.0 & 24.0 & 76.0 & 36.0 & 64.0\\
ManiSkill & 9 & 55.6 & 0.0 & -- & 0.0 & 11.1 & 0.0 & 0.0 & 55.6 & 11.1 & 55.6\\
LIBERO-PRO & 10 & 40.0 & 30.0 & 60.0 & 10.0 & 40.0 & 0.0 & 10.0 & 80.0 & 50.0 & 80.0\\
robosuite & 5 & 80.0 & 80.0 & 100.0 & 60.0 & 100.0 & 40.0 & 20.0 & 100.0 & 80.0 & 100.0\\
BEHAVIOR-1K & 5 & 0.0 & 0.0 & 0.0 & 0.0 & 0.0 & 0.0 & 0.0 & 0.0 & 0.0 & 0.0\\
\midrule
Total & 182 & 41.8 & 19.8 & 24.7$^\dagger$ & 9.3 & 18.7 & 9.9 & 11.0 & 47.8 & 20.9 & 44.0\\
\end{longtable}
\noindent $\dagger$See Appendix~\ref{app:w4-comparison} for coverage and aggregation.
}

%% file: sections/table_composition.tex
\begin{table}[!htbp]
\centering
\caption{Task-composition analysis (success \%). Spatial excludes VSI-Bench from the categorical set (160 remaining cases); Planning removes ScienceWorld and VirtualHome; Affordance removes UMD and uses strict box success. All columns retain the assigned-case denominator of the indicated subset. Spatial and Affordance show mean $\pm$ run SD; Planning uses one run.}
\label{tab:composition}
\ArenaTableFont
\begin{tabular*}{\linewidth}{@{\extracolsep{\fill}}lrrr@{}}
\toprule
Agent & \shortstack{Spatial without\\VSI} & \shortstack{Planning without\\text/symbolic} & \shortstack{Affordance without\\UMD} \\
\midrule
Astra & $83.6\!\pm\!1.2$ & 66.7 & $16.5\!\pm\!1.4$ \\
Sol & $71.4\!\pm\!3.3$ & 20.6 & $10.5\!\pm\!2.7$ \\
Fable & $77.9\!\pm\!3.1$ & 33.3 & $15.0\!\pm\!5.0$ \\
\mbox{Qwen-Max} & $80.9\!\pm\!1.4$ & 15.9 & $10.0\!\pm\!3.5$ \\
Gemini~3.8 & $78.4\!\pm\!2.2$ & 22.2 & $5.5\!\pm\!1.1$ \\
\mbox{Qwen-397B} & $62.2\!\pm\!1.6$ & 11.1 & $2.0\!\pm\!1.1$ \\
\mbox{Qwen-27B} & $59.4\!\pm\!0.9$ & 7.9 & $0.5\!\pm\!1.1$ \\
Opus~5.5 & $81.4\!\pm\!6.9$ & 28.6 & $19.0\!\pm\!4.9$ \\
\mbox{DeepSeek-1017} & $64.8\!\pm\!1.6$ & 14.3 & $4.0\!\pm\!3.8$ \\
Gemini~4 & $88.5\!\pm\!0.3$ & 60.3 & $28.0\!\pm\!2.1$\\
\bottomrule
\end{tabular*}
\end{table}

%% file: sections/appendix_opus_statistics.tex
\subsection{Opus~5.5 Repetitions, Dispersion, and Coverage}
\label{app:opus-statistics}
Opus~5.5 has five runs on 258 non-tracing Geometry cases, 220 Spatial Reasoning cases, and 70 point-and-box Affordance cases (2,740 assigned records). Four invalid InFlux references are excluded from eligible scoring. Repeated results use the equal-weight mean of per-run metrics and sample SD (divisor four). Tracing, direct polygons, Planning, and Manipulation each retain one run. Table~\ref{tab:opus-dispersion} gives metric-specific coverage.

\paragraph{Planning and Manipulation.}
Planning includes all 157 assignments. The five non-HumanCLAW sources contribute 73 successes, and HumanCLAW adds 3/13, for 76/157 (48.4\%). On HumanCLAW, Opus~5.5 succeeds on 3/5 navigation tasks and 0/8 sitting tasks. It finds the target and reaches within 20\,cm on 7/13 cases, including 4/8 sitting tasks; proximity does not establish interaction success. Opus~5.5 completes 87/182 Manipulation tasks, counting two RLBench exceptions and one RoboTwin exception as unsuccessful assignments. Its CALVIN reports use automatic per-round RGB and the ceilings in Appendix~\ref{app:w4-comparison}. The supplemental recordings in Appendix~\ref{app:latest-cases} illustrate individual episodes; source reports supply \mbox{their execution configurations.}

\paragraph{Sequential progress and final conditions.}
Opus~5.5 completes at least one through five consecutive CALVIN goals on 80\%, 70\%, 40\%, 40\%, and 10\% of the ten assignments. Summing these survival rates gives 2.40 consecutive goals on average. On the five ManiSkill diagnostic tasks, it satisfies all final goal conditions on PickCube, PlaceSphere, and StackCube, and none on PegInsertionSide or PlugCharger; their mean condition score is 60\%. These recorded diagnostics supplement the main evaluation; Opus~5.5 is outside the paired observation experiment. Opus~5.5 3D tracing has 8.9\% joint success across 56 assignments, alongside the 0.683\,m \mbox{RMSE in Table~\ref{tab:profile-full}.}

\begingroup
\ArenaTableFont
\begin{longtable}{@{}llrr@{}}
\caption{Opus~5.5 source results. Planning includes all six sources. Manipulation retains all binary assignments, including execution errors. Spatial values cover all questions in each source and show mean $\pm$ run SD across five repetitions.}\label{tab:opus-sources}\\
\toprule Domain & Source & Assigned & Success (\%)\\\midrule\endfirsthead
\toprule Domain & Source & Assigned & Success (\%)\\\midrule\endhead
\bottomrule\endfoot

Spatial & 3DSRBench & 50 & $76.8\!\pm\!9.9$\\
Spatial & BOP-ASK & 20 & $95.0\!\pm\!0.0$\\
Spatial & MindCube & 50 & $85.2\!\pm\!6.7$\\
Spatial & MMSI-Bench & 20 & $59.0\!\pm\!8.2$\\
Spatial & RoboSpatial-Home & 30 & $61.3\!\pm\!4.5$\\
Spatial & VSI-Bench & 50 & $36.4\!\pm\!1.7$\\
\midrule
Planning & ALFRED & 20 & 15.0\\
Planning & ALFWorld & 20 & 45.0\\
Planning & DiscoveryWorld & 10 & 30.0\\
Planning & ScienceWorld & 73 & 58.9\\
Planning & VirtualHome & 21 & 71.4\\
Planning & HumanCLAW & 13 & 23.1\\
Planning & All assigned & 157 & 48.4\\
\midrule
Manipulation & CALVIN & 10 & 10.0\\
Manipulation & CLIPort & 18 & 61.1\\
Manipulation & RLBench & 10 & 30.0\\
Manipulation & RoboCasa & 10 & 0.0\\
Manipulation & RoboCasa365 & 10 & 10.0\\
Manipulation & RoboWits & 10 & 10.0\\
Manipulation & RoboTwin 2.0 & 50 & 46.0\\
Manipulation & VIMA-Bench & 10 & 100.0\\
Manipulation & VLABench & 25 & 76.0\\
Manipulation & ManiSkill & 9 & 55.6\\
Manipulation & LIBERO-PRO & 10 & 80.0\\
Manipulation & Robosuite & 5 & 100.0\\
Manipulation & BEHAVIOR-1K & 5 & 0.0\\
Manipulation & All assigned & 182 & 47.8\\
\midrule
\end{longtable}
\begin{longtable}{@{}lrr@{}}
\caption{Opus~5.5 repeated metrics as mean $\pm$ run SD. Coverage counts valid continuous predictions or all assigned binary predictions across five runs.}\label{tab:opus-dispersion}\label{tab:opus-run-variance}\\
\toprule Metric & Mean $\pm$ SD & Scored/eligible\\\midrule\endfirsthead
\toprule Metric & Mean $\pm$ SD & Scored/eligible\\\midrule\endhead
\bottomrule\endfoot
InFlux focal MAPE & $60.27\!\pm\!1.49$ & 130/130\\
InFlux principal error & $3.82\!\pm\!0.00$ & 130/130\\
Angle relative L2 & $1.166\!\pm\!0.113$ & 60/60\\
Distance relative L2 & $0.455\!\pm\!0.069$ & 50/50\\
Vector relative L2 & $0.516\!\pm\!0.042$ & 40/40\\
Map-free rotation & $40.95\!\pm\!8.10$ & 142/150\\
Map-free translation & $243.13\!\pm\!35.82$ & 142/150\\
Map-free direction & $44.29\!\pm\!9.06$ & 142/150\\
GP focal MAPE & $16.48\!\pm\!6.13$ & 180/180\\
GP principal error & $60.46\!\pm\!4.13$ & 120/120\\
Depth MAE & $0.510\!\pm\!0.295$ & 60/60\\
Depth RMSE & $0.599\!\pm\!0.302$ & 60/60\\
Depth AbsRel & $0.169\!\pm\!0.108$ & 60/60\\
Displacement MAE & $1.164\!\pm\!0.475$ & 200/200\\
GP rotation & $3.05\!\pm\!0.96$ & 120/120\\
GP translation & $11.13\!\pm\!1.48$ & 120/120\\
Scale MAPE & $3.91\!\pm\!2.74$ & 220/220\\
Scale judgment Pass@1 & $100.0\!\pm\!0.0$ & 60/60\\
Numerical AbsRel & $0.184\!\pm\!0.040$ & 130/135\\
BBox--GT box IoU & $0.432\!\pm\!0.021$ & 350/350\\
BBox--GT mask IoU & $0.311\!\pm\!0.020$ & 350/350\\
BBox success@0.5 & $51.4\!\pm\!3.0$ & 350/350\\
Contact-point validity & $61.1\!\pm\!1.9$ & 350/350\\
Spatial Native Pass@1 & $67.5\!\pm\!5.1$ & 1100/1100\\
Categorical Pass@1 & $79.2\!\pm\!5.6$ & 915/915\\
\end{longtable}
\endgroup

%% file: sections/appendix_deepseek.tex
\subsection{\mbox{DeepSeek-1017}: Protocol, Dispersion, and Task Outcomes}
\label{app:deepseek}
\mbox{DeepSeek-1017} denotes \href{https://api-docs.deepseek.com/updates/}{DeepSeek-V4.1-Flash-1017}, the 17 October 2026 release served by \texttt{deepseek-flash}. Its main evaluation contains 3,192 assigned case--run slots covering all 1,000 cases; the added polygon experiment contributes 70 separate slots. Each case--run slot contributes one evaluation outcome under Appendix~\ref{app:latest-results}. Non-tracing Geometry, Spatial Reasoning, and point-and-box Affordance have five runs; tracing, Planning, and Manipulation have one. Four invalid InFlux references are excluded from eligible scoring. Failed or absent binary predictions retain their assigned denominator; continuous errors use finite valid predictions. Five-run SDs measure variation between run means.

\paragraph{Observations and execution budgets.}
Spatial runs use no YOLOE; 3DSRBench's corrected flipped inputs use matching native labels. In Manipulation, \mbox{DeepSeek-1017} uses native tools and receives current RGB views at every decision, completing 38/182 tasks. Its source-specific round, token, and in-episode time ceilings appear in Table~\ref{tab:deepseek-budgets}. Original Affordance uses native points and rectangles. A separate direct-polygon run scores all 70 assigned cases, yielding 0.249 IoU; its 67 valid outputs and three zero-scored failures are detailed in Appendix~\ref{app:polygon}.

\paragraph{Subgoal achievement and final task outcomes.}
The source breakdown in Table~\ref{tab:deepseek-sources} connects aggregate performance to task demands. \mbox{DeepSeek-1017} succeeds on 2/5 HumanCLAW navigation tasks and 0/8 sitting tasks. It solves 49/157 Planning assignments, falling to 9/63 when text/symbolic ScienceWorld and VirtualHome are removed. CALVIN averages 1.40 consecutive goals, while only 1/10 full five-goal sequences succeeds. In the five-task ManiSkill diagnostic, none of the final condition sets is fully satisfied. BEHAVIOR-1K has no complete success in five assignments: pre-termination logs confirm localization in four and a grasp candidate in three. Only stages confirmed before termination enter the diagnostic.

\begin{table}[!htbp]\centering\ArenaTableFont
\caption{\mbox{DeepSeek-1017} mean $\pm$ run SD and metric-specific coverage. Continuous metrics average valid predictions within each run; binary rates retain all assigned cases. Polygon, Planning, and Manipulation use one run.}\label{tab:deepseek-dispersion}
\begin{tabular*}{\linewidth}{@{\extracolsep{\fill}}lrrr@{}}\toprule
Metric & Mean & Run SD & Scored/eligible\\\midrule
Pose rotation ($^\circ$) & 30.2 & 3.9 & 259/270 \\
Pose translation (cm) & 182.0 & 29.0 & 259/270 \\
Spatial Pass@1 (\%) & 51.6 & 1.5 & 1100/1100 \\
Numerical AbsRel & 0.309 & 0.035 & 120/135 \\
Box--mask IoU & 0.217 & 0.011 & 348/350 \\
Contact validity (\%) & 50.9 & 4.2 & 350/350 \\
Polygon--mask IoU & 0.249 & --- & 70/70 \\
Planning success (\%) & 31.210 & --- & 157/157 \\
Manipulation success (\%) & 20.879 & --- & 182/182 \\
\bottomrule\end{tabular*}\end{table}

\begingroup\ArenaTableFont
\begin{longtable}{@{}llrr@{}}
\caption{\mbox{DeepSeek-1017} native source outcomes. Spatial includes all question types; Affordance uses the original point-or-box criterion. Manipulation measures complete-task success.}\label{tab:deepseek-sources}\\
\toprule Domain & Source & Cases & Success (\%)\\\midrule\endfirsthead
\toprule Domain & Source & Cases & Success (\%)\\\midrule\endhead
\bottomrule\endfoot
Spatial & 3DSRBench & 50 & $61.6\!\pm\!7.1$\\
Spatial & BOP-ASK & 20 & $88.0\!\pm\!7.6$\\
Spatial & MMSI-Bench & 20 & $53.0\!\pm\!5.7$\\
Spatial & MindCube & 50 & $54.4\!\pm\!4.3$\\
Spatial & RoboSpatial-Home & 30 & $58.0\!\pm\!3.8$\\
Spatial & VSI-Bench & 50 & $20.0\!\pm\!2.0$\\
Affordance & RAGNet-3DOI & 20 & $39.0\!\pm\!7.4$\\
Affordance & ReasonAff-style & 20 & $15.0\!\pm\!3.5$\\
Affordance & UMD & 30 & $86.0\!\pm\!6.4$\\
Planning & ALFRED & 20 & 5.0\\
Planning & ALFWorld & 20 & 20.0\\
Planning & DiscoveryWorld & 10 & 20.0\\
Planning & HumanCLAW & 13 & 15.4\\
Planning & ScienceWorld & 73 & 38.4\\
Planning & VirtualHome & 21 & 57.1\\
Manipulation & BEHAVIOR-1K & 5 & 0.0\\
Manipulation & CALVIN & 10 & 10.0\\
Manipulation & CLIPort & 18 & 33.3\\
Manipulation & LIBERO-PRO & 10 & 50.0\\
Manipulation & ManiSkill & 9 & 11.1\\
Manipulation & RLBench & 10 & 20.0\\
Manipulation & RoboCasa & 10 & 0.0\\
Manipulation & RoboCasa365 & 10 & 0.0\\
Manipulation & RoboTwin2 & 50 & 4.0\\
Manipulation & RoboWits & 10 & 20.0\\
Manipulation & Robosuite & 5 & 80.0\\
Manipulation & VIMA-Bench & 10 & 60.0\\
Manipulation & VLABench & 25 & 36.0\\
\end{longtable}\endgroup

\begin{table}[!htbp]\centering\ArenaTableFont
\caption{\mbox{DeepSeek-1017} Manipulation ceilings. Tokens are cumulative; minutes are in-episode wall-clock limits. Process wrappers allow additional startup and teardown time. Each assigned task is one episode. ManiSkill success counts nine binary tasks; its additional locomotion case \mbox{has a continuous reward.}}\label{tab:deepseek-budgets}
\begin{tabular*}{\linewidth}{@{\extracolsep{\fill}}lrrr@{}}\toprule
Source & Model rounds & Tokens & Minutes\\\midrule
CALVIN & 40 & 1280k & 120 \\
BEHAVIOR-1K & 24 & 640k & 180 \\
LIBERO-PRO & 24 & 640k & 180 \\
Robosuite & 24 & 640k & 180 \\
CLIPort & 24 & 960k & 60 \\
ManiSkill & 32 & 640k & 60 \\
RLBench & 16 & 640k & 60 \\
RoboCasa & 24 & 1120k & 60 \\
RoboCasa365 & 24 & 1120k & 60 \\
RoboTwin2 & 16 & 640k & 60 \\
RoboWits & 24 & 640k & 60 \\
VIMA-Bench & 16 & 640k & 60 \\
VLABench & 16 & 640k & 60 \\
\bottomrule\end{tabular*}\end{table}

\begin{table}[!htbp]\centering\ArenaTableFont
\caption{\mbox{DeepSeek-1017} execution stages. Fractions count confirmed events per assignment, with zero credit for unverified events. BEHAVIOR-1K uses pre-termination evidence.}\label{tab:deepseek-control}
\begin{tabularx}{\linewidth}{@{}lYr@{}}\toprule
Source & Diagnostic & Confirmed / assigned\\\midrule
CALVIN & All five ordered goals completed & 1/10\\
ManiSkill & All final conditions on the five-task subset & 0/5\\
RoboCasa & Approach; grasp and lift; stable release & 3/4; 3/4; 0/4\\
RoboCasa365 & Approach; grasp and lift; stable release & 3/7; 3/7; 0/7\\
BEHAVIOR-1K & Localization; grasp candidate & 4/5; 3/5\\
 & Arrival; holding; held motion; release & 0/5 each\\
\bottomrule\end{tabularx}\end{table}

%% file: sections/appendix_polygon.tex
\subsection{Direct Polygon Grounding}
\label{app:polygon}
The ten-model Affordance experiment retains the same 70 images and interaction instructions, with 20 RAGNet, 20 ReasonAff-style/3DOI, and 30 UMD cases. Each model directly predicts polygons in image coordinates alongside its point and bounding box. Rasterizing the polygons yields a predicted mask for IoU and Dice; no SAM, detector, or iterative segmentation refinement is added. The separate polygon run tests a more flexible output representation on the same tasks. Table~\ref{tab:profile} compares box and polygon overlap, while Table~\ref{tab:profile-full} reports point validity.

\paragraph{Coverage and aggregation.}
Table~\ref{tab:polygon-sources} reports single-run polygon scores by source. Invalid polygons and absent final submissions score zero. The original point/box pass flag is not treated as polygon success. Single-run scores have no run-to-run SD. \mbox{DeepSeek-1017} yields 67 valid polygons: two outputs have vertices outside the image, and one case reaches the 12-round limit without a submission. All three contribute zero to the 70-case mean (24.92\%). Its thinking-enabled, high-effort setting uses a 1,800-token visible-response ceiling; the 120k-token case budget also counts reasoning. The other limits remain 12 rounds and 900 seconds per case. Each of the 70 cases contributes one attempt.

\begin{table}[!htbp]
\centering\ArenaTableFont
\caption{Polygon--mask IoU (\%), with one run on 20 RAGNet, 20 ReasonAff-style, and 30 UMD cases. Invalid or absent submissions score zero; all 70 cases enter the overall mean.}\label{tab:polygon-sources}
\begin{tabular*}{\linewidth}{@{\extracolsep{\fill}}lcccc@{}}\toprule
Model & RAGNet & ReasonAff & UMD & Overall\\\midrule
Astra & 19.19 & 10.60 & 68.52 & 37.88\\
Sol & 13.22 & 7.40 & 62.17 & 32.54\\
Fable & 12.95 & 17.00 & 56.51 & 32.77\\
\mbox{Qwen-Max} & 5.64 & 7.01 & 63.28 & 30.74\\
Gemini~3.8 & 10.92 & 4.21 & 57.71 & 29.06\\
\mbox{Qwen-397B} & 5.86 & 2.45 & 16.02 & 9.24\\
\mbox{Qwen-27B} & 3.98 & 0.30 & 9.05 & 5.10\\
Opus~5.5 & 18.55 & 13.29 & 71.49 & 39.74\\
\mbox{DeepSeek-1017} & 8.67 & 9.22 & 46.21 & 24.92\\
Gemini~4~Argon & 14.03 & 30.71 & 73.08 & 44.10\\
\bottomrule\end{tabular*}\end{table}

Gemini~4~Argon has the highest aggregate at 44.10\%, followed by Opus~5.5 at 39.74\% and Astra at 37.88\%; all three score far lower on the interaction-conditioned scenes than on \mbox{UMD tools (Figure~\ref{fig:astra-affordance}b).}

%% file: sections/appendix_cost_profile.tex
\subsection{Interpreting the Capability Profiles}
\label{app:cost-profile}
Figure~\ref{fig:capability-cost} places complementary metrics on fixed visualization scales. G, S, A, P, and M denote Geometry, Spatial Reasoning, Affordance, Planning, and Manipulation. The first three indices are defined by
\begin{equation*}
\begin{aligned}
G &= 50\left[\exp\!\left(-\frac{r}{45}\right)+\exp\!\left(-\frac{t}{150}\right)\right],\\
S &= \frac{p+100\exp(-a/0.25)}{2},\\
A &= \frac{100u+c}{2}.
\end{aligned}
\end{equation*}
Here, $r$ and $t$ are the mean per-run rotation error in degrees and translation error in centimeters; $p$ is native Pass@1 (\%) and $a$ is numerical AbsRel; $u$ is original box--mask IoU and $c$ is contact validity (\%). Planning and Manipulation use their task success rates (\%). These monotone scales summarize complementary metrics; they are not native task tolerances or success probabilities.

Opus~5.5 includes all 157 Planning assignments, scoring 76/157 (48.4\%), and all 182 Manipulation assignments. The Affordance index uses box--mask IoU and contact validity from Table~\ref{tab:profile-full}; polygon IoU is reported as an additional metric.

\mbox{DeepSeek-1017} uses the same transformations, giving G/S/A/P indices of 40.4/40.4/36.3/31.2 and a Manipulation score of 20.9. Its Manipulation result uses the recorded per-round RGB schedule. Gemini~4~Argon gives 57.8/61.2/47.7/70.1/44.0 and also uses per-round RGB in Manipulation. Panel a plots all ten profiles, and panel b reports their domain indices separately.

%% file: sections/appendix_astra_subtasks.tex
\Needspace*{10\baselineskip}
\section{Task Requirements and Model Comparisons}
\label{app:astra-subtasks}
These diagnostics explain the task requirements behind Section~\ref{sec:astra} and Figures~\ref{fig:astra-geometry}--\ref{fig:astra-manipulation}. Source and task-group comparisons include all ten agents. Geometry and Affordance diagnostics include Opus~5.5 and Gemini~4~Argon; paired sensitivity checks use the specified shared repetitions. Each domain connects the aggregate score to specific estimation, grounding, or execution conditions; Appendix~\ref{app:latest-results} defines repetitions and valid coverage.

\subsection{Geometry: Task Requirements and Source Conditions}
\label{app:geometry-focus}
\noindent\textbf{Task-level comparisons.}
Table~\ref{tab:geometry-focus-stats}a supplies the absolute errors behind Figure~\ref{fig:astra-geometry}a. Its relative error reduction (\%) is
\begin{equation*}
100\left(1-\frac{e_{\mathrm{Astra}}}{e_{\mathrm{peer}}}\right),
\end{equation*}
using the lowest-error non-Astra peer in the original cohort for each target. Equal-run means retain task-specific units: MAPE for focal length, size, and height, and percentage AbsRel for depth. Astra leads the five GeoProbe targets. Gemini~4~Argon has the lowest InFlux focal error (34.8\%), improving on the strongest original peer, Sol (47.8\%). Its GeoProbe focal error is 24.3\%, however, versus Astra's 6.2\%: the advantage does not transfer uniformly between real-camera and rendered calibration tasks.

\noindent\textbf{Coverage sensitivity.}
Table~\ref{tab:geometry-focus-common}b restricts each Astra--peer comparison to cases valid in every shared run. All six comparison directions remain unchanged. The ranking is thus stable on the common valid cohort.
\begin{table}[htbp]
\centering
\caption{Geometry errors (\%), mean $\pm$ run SD; lower is better. (a) All ten agents. GP denotes GeoProbe; focal uses focal-only queries, and depth uses percentage AbsRel. (b) Original paired comparisons on common-valid cases and runs.}\label{tab:geometry-focus-stats}\label{tab:geometry-focus-common}
\noindent{\ArenaTableFont\textbf{(a) All valid cases}\par}\smallskip
\centering

\ArenaTableFont
\setlength{\tabcolsep}{3pt}
\begin{tabular*}{\linewidth}{@{\extracolsep{\fill}}l*{6}{r}@{}}\toprule
Model & GP focal & InFlux & GP depth & GP size & Ref. height & Height ratio\\\midrule
Astra & $6.2\!\pm\!2.1$ & $53.0\!\pm\!2.1$ & $10.4\!\pm\!2.9$ & $6.3\!\pm\!1.0$ & $0.31\!\pm\!0.11$ & $0.32\!\pm\!0.15$\\
Sol & $11.2\!\pm\!5.5$ & $47.8\!\pm\!2.0$ & $22.9\!\pm\!2.7$ & $8.5\!\pm\!0.7$ & $0.90\!\pm\!0.36$ & $0.95\!\pm\!0.41$\\
Fable & $29.5\!\pm\!1.8$ & $52.6\!\pm\!2.2$ & $14.2\!\pm\!2.7$ & $7.6\!\pm\!0.2$ & $2.31\!\pm\!0.08$ & $2.23\!\pm\!0.17$\\
Q-Max & $13.8\!\pm\!3.1$ & $55.5\!\pm\!1.0$ & $16.3\!\pm\!1.3$ & $9.6\!\pm\!0.8$ & $1.69\!\pm\!0.72$ & $2.17\!\pm\!0.22$\\
Gemini~3.8 & $23.7\!\pm\!12.9$ & $52.4\!\pm\!3.4$ & $19.2\!\pm\!2.1$ & $8.3\!\pm\!0.3$ & $2.73\!\pm\!0.75$ & $2.67\!\pm\!0.28$\\
Q-397B & $38.6\!\pm\!3.2$ & $67.8\!\pm\!1.8$ & $31.2\!\pm\!2.8$ & $11.4\!\pm\!0.6$ & $17.55\!\pm\!1.22$ & $20.35\!\pm\!0.87$\\
Q-27B & $38.4\!\pm\!5.0$ & $65.9\!\pm\!2.1$ & $29.2\!\pm\!1.3$ & $9.4\!\pm\!0.3$ & $13.69\!\pm\!0.86$ & $17.08\!\pm\!2.70$\\
Opus~5.5 & $15.9\!\pm\!6.3$ & $60.3\!\pm\!1.5$ & $16.9\!\pm\!10.8$ & $7.4\!\pm\!0.9$ & $3.97\!\pm\!5.35$ & $2.25\!\pm\!2.48$\\
DS-1017 & $38.1\!\pm\!10.5$ & $62.9\!\pm\!1.6$ & $22.9\!\pm\!2.5$ & $8.9\!\pm\!0.7$ & $5.71\!\pm\!1.49$ & $8.35\!\pm\!5.02$\\
Gemini~4 & $24.3\!\pm\!5.4$ & $34.8\!\pm\!1.6$ & $29.9\!\pm\!0.6$ & $7.0\!\pm\!0.3$ & $0.45\!\pm\!0.04$ & $0.60\!\pm\!0.10$\\
\bottomrule\end{tabular*}
\par\medskip
\noindent{\ArenaTableFont\textbf{(b) Matched common-valid cohorts}\par}\smallskip
\centering

\ArenaTableFont
\setlength{\tabcolsep}{3pt}
\begin{tabular*}{\linewidth}{@{\extracolsep{\fill}}lrrlrr@{}}
\toprule
Task & Cases & Rounds & Peer & Astra & Peer\\\midrule
GP focal-only & 12 & 5 & Sol & $6.21\!\pm\!2.10$ & $11.16\!\pm\!5.47$\\
InFlux focal & 26 & 5 & Sol & $53.02\!\pm\!2.15$ & $47.79\!\pm\!2.00$\\
GP depth & 9 & 3 & Fable & $10.16\!\pm\!3.49$ & $14.22\!\pm\!2.88$\\
GP object size & 12 & 3 & Fable & $6.44\!\pm\!1.36$ & $7.63\!\pm\!0.18$\\
GP ref. height & 12 & 5 & Sol & $0.31\!\pm\!0.11$ & $0.90\!\pm\!0.36$\\
GP height ratio & 12 & 5 & Sol & $0.32\!\pm\!0.15$ & $0.95\!\pm\!0.41$\\
\bottomrule\end{tabular*}
\end{table}

\noindent\textbf{Camera motion and real photographs.}
The rendered displacement subsets share 12 object interventions while changing the camera condition. Table~\ref{tab:geometry-focus-motion} compares Astra with Sol, the lowest-MAE original non-Astra peer under each displacement condition. The fixed-camera advantage persists when the two stationary Blender cases are excluded. Under the native direction-and-1\,cm criterion, both agents achieve 96.7\% fixed-camera success: the continuous error distinguishes precision that this tolerance does not resolve.

The four real-photo pairs form a separate tabletop subset. Their displacement labels are supplied annotations, without independent camera or metric calibration. The Blender comparison uses different scenes and object motions.

\begin{table}[htbp]
\centering
\caption{Displacement MAE (cm), mean $\pm$ run SD. Sol is the strongest original non-Astra comparator; Opus~5.5, \mbox{DeepSeek-1017}, and Gemini~4~Argon use the same task subsets. Moving-only excludes stationary cases.}
\label{tab:geometry-focus-motion}
\ArenaTableFont
\setlength{\tabcolsep}{3pt}
\begin{tabular*}{\linewidth}{@{\extracolsep{\fill}}lrrrrrr@{}}
\toprule
Condition & Cases & Astra & Sol & Opus~5.5 & DS-1017 & Gemini~4\\\midrule
Blender: fixed camera & 12 & $0.240\!\pm\!0.109$ & $0.467\!\pm\!0.267$ & $0.478\!\pm\!0.202$ & $1.517\!\pm\!0.105$ & $0.605\!\pm\!0.266$\\
Blender: fixed, moving-only & 10 & $0.288\!\pm\!0.131$ & $0.560\!\pm\!0.321$ & $0.574\!\pm\!0.243$ & $1.780\!\pm\!0.076$ & $0.726\!\pm\!0.319$\\
Blender: small view change & 12 & $0.567\!\pm\!0.158$ & $0.678\!\pm\!0.152$ & $1.208\!\pm\!0.917$ & $2.022\!\pm\!0.346$ & $1.442\!\pm\!0.181$\\
Blender: large view change & 12 & $1.125\!\pm\!0.144$ & $1.958\!\pm\!0.295$ & $1.992\!\pm\!0.567$ & $2.492\!\pm\!0.908$ & $2.183\!\pm\!0.168$\\
Real photographs & 4 & $0.050\!\pm\!0.068$ & $0.415\!\pm\!0.236$ & $0.600\!\pm\!0.347$ & $1.250\!\pm\!0.234$ & $0.400\!\pm\!0.229$\\
\bottomrule\end{tabular*}
\end{table}

\Needspace{4\baselineskip}
\noindent\textbf{Coupled motion and metric scale reveal distinct estimation errors.}
Table~\ref{tab:astra-geometry-subtasks} separates the GeoProbe pose modes: mixed rotation and translation produce 14.16\,cm translation error and zero joint success, despite much lower errors on simpler motions. The best non-Astra translation estimate on the mixed subset is \mbox{Qwen-Max} at 16.77\,cm; none of the original seven agents achieves joint success on this subset. Translation-only cases have zero target rotation. Depth also separates ordering from metric scale: Astra's ordinal accuracy is 96.3\%, whereas raw AbsRel is 0.104; oracle scale alignment reduces it to 0.038. Alignment uses reference depth.

Explicit scale references and regular rendered geometry support these estimates.
\begin{table}[!htbp]
\centering
\caption{GeoProbe camera-pose modes. Errors and joint success show mean $\pm$ run SD. Joint success uses $2^\circ$/2\,cm tolerances; failed predictions remain in its denominator.}\label{tab:astra-geometry-subtasks}
\ArenaTableFont
\setlength{\tabcolsep}{3pt}
\begin{tabular*}{\linewidth}{@{\extracolsep{\fill}}llrrrr@{}}
\toprule
Motion & Model & Cases & Rotation ($^\circ$) & Translation (cm) & Joint success (\%)\\\midrule
Stationary & Astra & 4 & $0.00\!\pm\!0.00$ & $0.00\!\pm\!0.00$ & $100.0\!\pm\!0.0$\\
 & \mbox{DeepSeek-1017} & 4 & $0.00\!\pm\!0.00$ & $0.00\!\pm\!0.00$ & $100.0\!\pm\!0.0$\\
 & Opus~5.5 & 4 & $0.00\!\pm\!0.00$ & $0.00\!\pm\!0.00$ & $100.0\!\pm\!0.0$\\
 & Gemini~4 & 4 & $0.00\!\pm\!0.00$ & $0.00\!\pm\!0.00$ & $100.0\!\pm\!0.0$\\
Rotation only & Astra & 6 & $0.35\!\pm\!0.78$ & $1.00\!\pm\!2.24$ & $96.7\!\pm\!7.5$\\
 & \mbox{DeepSeek-1017} & 6 & $5.43\!\pm\!1.57$ & $13.82\!\pm\!5.69$ & $13.3\!\pm\!13.9$\\
 & Opus~5.5 & 6 & $3.40\!\pm\!4.35$ & $2.40\!\pm\!3.57$ & $56.7\!\pm\!32.5$\\
 & Gemini~4 & 6 & $1.23\!\pm\!0.09$ & $0.00\!\pm\!0.00$ & $83.3\!\pm\!0.0$\\
Translation only & Astra & 6 & $0.00\!\pm\!0.00$ & $1.55\!\pm\!1.59$ & $76.7\!\pm\!25.3$\\
 & \mbox{DeepSeek-1017} & 6 & $1.18\!\pm\!0.91$ & $11.03\!\pm\!3.22$ & $20.0\!\pm\!13.9$\\
 & Opus~5.5 & 6 & $0.30\!\pm\!0.38$ & $3.48\!\pm\!3.20$ & $76.7\!\pm\!14.9$\\
 & Gemini~4 & 6 & $0.00\!\pm\!0.00$ & $2.63\!\pm\!0.51$ & $46.7\!\pm\!7.5$\\
Rotation + translation & Astra & 8 & $3.26\!\pm\!0.75$ & $14.16\!\pm\!3.94$ & $0.0\!\pm\!0.0$\\
 & \mbox{DeepSeek-1017} & 8 & $6.36\!\pm\!0.54$ & $25.16\!\pm\!2.73$ & $0.0\!\pm\!0.0$\\
 & Opus~5.5 & 8 & $6.37\!\pm\!1.15$ & $28.97\!\pm\!2.30$ & $0.0\!\pm\!0.0$\\
 & Gemini~4 & 8 & $7.22\!\pm\!1.05$ & $28.44\!\pm\!2.68$ & $0.0\!\pm\!0.0$\\
\bottomrule\end{tabular*}
\end{table}

\noindent\textbf{TraceSpatial-Bench: path matching and boundary constraints.}
Astra has lower 3D path RMSE than Gemini~3.8 (0.382 versus 0.426\,m), but fewer jointly valid paths (19/56 versus 21/56; Table~\ref{tab:profile-full}). Among Astra's trajectories, 27 fail initiation, 24 fail arrival, and ten fail collision checks; these sets overlap. Of its 37 joint failures, 36 violate initiation or arrival, while only one satisfies both boundaries and fails collision avoidance alone. This identifies initiation and arrival as the dominant failed constraints in Figure~\ref{fig:astra-geometry}c. The offline tolerances and sampled-position collision rule are defined in Appendix~\ref{app:geometry}.

\subsection{Spatial Reasoning: Cross-Source Task Contrasts}
\label{app:spatial-focus}
\noindent\textbf{Complete task coverage.}
Table~\ref{tab:astra-spatial-subtasks} covers all 43 groups across six sources. Source categories are retained; MindCube uses task axes and question templates, prioritizing composed rotation and turn--translation before individual motion and relation queries.

\noindent\textbf{Figure labels and comparisons.}
Figure~\ref{fig:astra-spatial}a uses six groups: BOP-ASK left/right (12 cases), MindCube object relations (12) and camera shift (13), 3DSRBench actor-relative side (7) and object facing (6), and VSI hard direction (4). Panel b uses BOP closer/farther (5/3), 3DSR camera/object distance (10/2), and VSI relative distance (6). Every model uses the same case IDs in each group. Values average five run accuracies, except Fable's three, and retain invalid predictions in the denominator. Gemini~3.8 Flash and Gemini~4~Argon have separate rows. The released CSV gives all 110 cell means, run SDs, and denominators. The groups isolate relation and reference-frame requirements within each source.

\noindent\textbf{Shared-round gaps.}
The original diagnostic below selects one task per source and its highest-mean original non-Astra peer; Figure~\ref{fig:astra-spatial}c shows native source scores for all ten agents. Gaps and SD then use shared run indices (Table~\ref{tab:spatial-source-gaps}; protocol in Appendix~\ref{app:scoring}). On object-facing queries, Fable's three-run mean is 72.2\% versus Astra's 50.0\%; Astra averages 56.7\% \mbox{across all five runs.}

\begin{table}[!htbp]\centering
\caption{Original source-specific Astra--peer comparisons. Accuracy and gaps use shared available runs and are in percentages and percentage points. Run-index alignment follows Appendix~\ref{app:scoring}.}\label{tab:spatial-source-gaps}
\ArenaTableFont\setlength{\tabcolsep}{3pt}
\begin{tabular*}{\linewidth}{@{\extracolsep{\fill}}llrrrr@{}}\toprule
Task & Peer & Rounds & Astra & Peer & Gap $\pm$ SD\\\midrule
MMSI camera--region & Fable & 3 & 100.0 & 77.8 & $+22.2\!\pm\!19.2$\\
VSI medium direction & \mbox{Qwen-Max} & 5 & 100.0 & 83.3 & $+16.7\!\pm\!11.8$\\
BOP-ASK left/right & Gemini~3.8 & 5 & 95.0 & 96.7 & $-1.7\!\pm\!7.0$\\
MindCube camera shift & \mbox{Qwen-Max} & 5 & 78.5 & 89.2 & $-10.8\!\pm\!14.0$\\
3DSR object facing & Fable & 3 & 50.0 & 72.2 & $-22.2\!\pm\!9.6$\\
RoboSpatial configuration & Q-397B & 5 & 78.0 & 100.0 & $-22.0\!\pm\!4.5$\\
\bottomrule\end{tabular*}\end{table}

\noindent\textbf{Numerical error and acceptance are complementary.}
The numerical rows in Table~\ref{tab:astra-spatial-subtasks} use the exact-match criterion from Appendix~\ref{app:scoring}. Table~\ref{tab:spatial-numeric-models}a instead measures continuous error on valid predictions. Fixing Astra's cohort across runs preserves low object-size AbsRel (0.091), while absolute-distance coverage is only 25/45 submissions (Table~\ref{tab:spatial-numeric-coverage}b). The table reports error together with valid-prediction coverage.

{\ArenaDenseTableFont
\setlength{\tabcolsep}{1.3pt}
\renewcommand{\arraystretch}{1.08}
\begin{longtable}{@{}>{\raggedright\arraybackslash}p{108pt}r*{10}{r}@{}}
\caption{All 43 Spatial task groups across ten agents. Entries show mean above $\pm$ run SD (\%); bold means mark the highest score in each row, including ties. $N$ counts distinct cases; each assigned case remains in the success denominator.}\label{tab:astra-spatial-subtasks}\\
\toprule
Task & $N$ & \textbf{Astra} & Sol & Fable & Q-Max & Gemini~3.8 & Q-397B & Q-27B & Opus~5.5 & DS-1017 & Gemini~4\\*
\midrule\endfirsthead
\multicolumn{12}{l}{\textit{Table \thetable\ continued. Values are mean $\pm$ run SD (\%).}}\\
\toprule
Task & $N$ & \textbf{Astra} & Sol & Fable & Q-Max & Gemini~3.8 & Q-397B & Q-27B & Opus~5.5 & DS-1017 & Gemini~4\\*
\midrule\endhead
\bottomrule\endfoot
\multicolumn{12}{@{}l@{}}{\colorbox[HTML]{E8EDF1}{\parbox{\dimexpr\linewidth-2\fboxsep\relax}{\textbf{3DSRBench}\hfill 50 cases}}}\\*
World height & 4 & \shortstack{$60.0$\\$\pm13.7$} & \shortstack{$50.0$\\$\pm0.0$} & \shortstack{$\mathbf{66.7}$\\$\pm14.4$} & \shortstack{$55.0$\\$\pm11.2$} & \shortstack{$25.0$\\$\pm0.0$} & \shortstack{$50.0$\\$\pm0.0$} & \shortstack{$50.0$\\$\pm0.0$} & \shortstack{$55.0$\\$\pm11.2$} & \shortstack{$25.0$\\$\pm0.0$} & \shortstack{$50.0$\\$\pm0.0$}\\
Above / below & 7 & \shortstack{$77.1$\\$\pm7.8$} & \shortstack{$60.0$\\$\pm12.0$} & \shortstack{$71.4$\\$\pm0.0$} & \shortstack{$77.1$\\$\pm12.8$} & \shortstack{$74.3$\\$\pm12.0$} & \shortstack{$74.3$\\$\pm6.4$} & \shortstack{$68.6$\\$\pm12.0$} & \shortstack{$80.0$\\$\pm7.8$} & \shortstack{$60.0$\\$\pm21.2$} & \shortstack{$\mathbf{85.7}$\\$\pm0.0$}\\
Camera depth & 10 & \shortstack{$90.0$\\$\pm7.1$} & \shortstack{$98.0$\\$\pm4.5$} & \shortstack{$93.3$\\$\pm5.8$} & \shortstack{$90.0$\\$\pm0.0$} & \shortstack{$92.0$\\$\pm4.5$} & \shortstack{$96.0$\\$\pm5.5$} & \shortstack{$\mathbf{100.0}$\\$\pm0.0$} & \shortstack{$98.0$\\$\pm4.5$} & \shortstack{$92.0$\\$\pm8.4$} & \shortstack{$\mathbf{100.0}$\\$\pm0.0$}\\
Next to & 1 & \shortstack{$0.0$\\$\pm0.0$} & \shortstack{$0.0$\\$\pm0.0$} & \shortstack{$0.0$\\$\pm0.0$} & \shortstack{$0.0$\\$\pm0.0$} & \shortstack{$0.0$\\$\pm0.0$} & \shortstack{$60.0$\\$\pm54.8$} & \shortstack{$0.0$\\$\pm0.0$} & \shortstack{$60.0$\\$\pm54.8$} & \shortstack{$0.0$\\$\pm0.0$} & \shortstack{$\mathbf{100.0}$\\$\pm0.0$}\\
Object distance & 2 & \shortstack{$70.0$\\$\pm27.4$} & \shortstack{$80.0$\\$\pm27.4$} & \shortstack{$83.3$\\$\pm28.9$} & \shortstack{$\mathbf{100.0}$\\$\pm0.0$} & \shortstack{$\mathbf{100.0}$\\$\pm0.0$} & \shortstack{$50.0$\\$\pm0.0$} & \shortstack{$80.0$\\$\pm27.4$} & \shortstack{$50.0$\\$\pm50.0$} & \shortstack{$50.0$\\$\pm0.0$} & \shortstack{$\mathbf{100.0}$\\$\pm0.0$}\\
Parallel / perpendicular & 1 & \shortstack{$\mathbf{100.0}$\\$\pm0.0$} & \shortstack{$\mathbf{100.0}$\\$\pm0.0$} & \shortstack{$\mathbf{100.0}$\\$\pm0.0$} & \shortstack{$\mathbf{100.0}$\\$\pm0.0$} & \shortstack{$20.0$\\$\pm44.7$} & \shortstack{$\mathbf{100.0}$\\$\pm0.0$} & \shortstack{$80.0$\\$\pm44.7$} & \shortstack{$\mathbf{100.0}$\\$\pm0.0$} & \shortstack{$40.0$\\$\pm54.8$} & \shortstack{$\mathbf{100.0}$\\$\pm0.0$}\\
Same/different facing & 4 & \shortstack{$55.0$\\$\pm11.2$} & \shortstack{$50.0$\\$\pm0.0$} & \shortstack{$50.0$\\$\pm0.0$} & \shortstack{$50.0$\\$\pm0.0$} & \shortstack{$50.0$\\$\pm0.0$} & \shortstack{$\mathbf{75.0}$\\$\pm17.7$} & \shortstack{$35.0$\\$\pm13.7$} & \shortstack{$50.0$\\$\pm0.0$} & \shortstack{$40.0$\\$\pm13.7$} & \shortstack{$50.0$\\$\pm0.0$}\\
Object-facing side & 6 & \shortstack{$56.7$\\$\pm9.1$} & \shortstack{$53.3$\\$\pm13.9$} & \shortstack{$72.2$\\$\pm9.6$} & \shortstack{$66.7$\\$\pm0.0$} & \shortstack{$66.7$\\$\pm0.0$} & \shortstack{$36.7$\\$\pm18.3$} & \shortstack{$23.3$\\$\pm9.1$} & \shortstack{$73.3$\\$\pm22.4$} & \shortstack{$56.7$\\$\pm19.0$} & \shortstack{$\mathbf{86.7}$\\$\pm13.9$}\\
In front of actor & 4 & \shortstack{$\mathbf{100.0}$\\$\pm0.0$} & \shortstack{$\mathbf{100.0}$\\$\pm0.0$} & \shortstack{$\mathbf{100.0}$\\$\pm0.0$} & \shortstack{$\mathbf{100.0}$\\$\pm0.0$} & \shortstack{$\mathbf{100.0}$\\$\pm0.0$} & \shortstack{$\mathbf{100.0}$\\$\pm0.0$} & \shortstack{$95.0$\\$\pm11.2$} & \shortstack{$\mathbf{100.0}$\\$\pm0.0$} & \shortstack{$90.0$\\$\pm13.7$} & \shortstack{$95.0$\\$\pm11.2$}\\
Actor left/right & 7 & \shortstack{$74.3$\\$\pm6.4$} & \shortstack{$71.4$\\$\pm14.3$} & \shortstack{$61.9$\\$\pm8.2$} & \shortstack{$71.4$\\$\pm10.1$} & \shortstack{$65.7$\\$\pm7.8$} & \shortstack{$45.7$\\$\pm12.0$} & \shortstack{$5.7$\\$\pm7.8$} & \shortstack{$71.4$\\$\pm17.5$} & \shortstack{$74.3$\\$\pm6.4$} & \shortstack{$\mathbf{100.0}$\\$\pm0.0$}\\
Object viewpoint & 4 & \shortstack{$70.0$\\$\pm11.2$} & \shortstack{$70.0$\\$\pm11.2$} & \shortstack{$50.0$\\$\pm0.0$} & \shortstack{$60.0$\\$\pm13.7$} & \shortstack{$70.0$\\$\pm11.2$} & \shortstack{$70.0$\\$\pm11.2$} & \shortstack{$65.0$\\$\pm28.5$} & \shortstack{$70.0$\\$\pm27.4$} & \shortstack{$30.0$\\$\pm11.2$} & \shortstack{$\mathbf{100.0}$\\$\pm0.0$}\\
\addlinespace
\multicolumn{12}{@{}l@{}}{\colorbox[HTML]{E8EDF1}{\parbox{\dimexpr\linewidth-2\fboxsep\relax}{\textbf{BOP-ASK}\hfill 20 cases}}}\\*
Closer to camera & 5 & \shortstack{$96.0$\\$\pm8.9$} & \shortstack{$96.0$\\$\pm8.9$} & \shortstack{$86.7$\\$\pm11.5$} & \shortstack{$\mathbf{100.0}$\\$\pm0.0$} & \shortstack{$96.0$\\$\pm8.9$} & \shortstack{$80.0$\\$\pm14.1$} & \shortstack{$84.0$\\$\pm8.9$} & \shortstack{$\mathbf{100.0}$\\$\pm0.0$} & \shortstack{$84.0$\\$\pm16.7$} & \shortstack{$92.0$\\$\pm11.0$}\\
Farther from camera & 3 & \shortstack{$\mathbf{100.0}$\\$\pm0.0$} & \shortstack{$\mathbf{100.0}$\\$\pm0.0$} & \shortstack{$\mathbf{100.0}$\\$\pm0.0$} & \shortstack{$\mathbf{100.0}$\\$\pm0.0$} & \shortstack{$\mathbf{100.0}$\\$\pm0.0$} & \shortstack{$66.7$\\$\pm0.0$} & \shortstack{$40.0$\\$\pm14.9$} & \shortstack{$93.3$\\$\pm14.9$} & \shortstack{$86.7$\\$\pm18.3$} & \shortstack{$\mathbf{100.0}$\\$\pm0.0$}\\
Left / right & 12 & \shortstack{$95.0$\\$\pm4.6$} & \shortstack{$80.0$\\$\pm4.6$} & \shortstack{$86.1$\\$\pm4.8$} & \shortstack{$86.7$\\$\pm4.6$} & \shortstack{$96.7$\\$\pm4.6$} & \shortstack{$91.7$\\$\pm0.0$} & \shortstack{$95.0$\\$\pm4.6$} & \shortstack{$93.3$\\$\pm3.7$} & \shortstack{$90.0$\\$\pm7.0$} & \shortstack{$\mathbf{100.0}$\\$\pm0.0$}\\
\addlinespace
\multicolumn{12}{@{}l@{}}{\colorbox[HTML]{E8EDF1}{\parbox{\dimexpr\linewidth-2\fboxsep\relax}{\textbf{MindCube}\hfill 50 cases}}}\\*
Behind observer & 7 & \shortstack{$97.1$\\$\pm6.4$} & \shortstack{$\mathbf{100.0}$\\$\pm0.0$} & \shortstack{$\mathbf{100.0}$\\$\pm0.0$} & \shortstack{$\mathbf{100.0}$\\$\pm0.0$} & \shortstack{$88.6$\\$\pm12.0$} & \shortstack{$80.0$\\$\pm16.3$} & \shortstack{$65.7$\\$\pm16.3$} & \shortstack{$\mathbf{100.0}$\\$\pm0.0$} & \shortstack{$\mathbf{100.0}$\\$\pm0.0$} & \shortstack{$\mathbf{100.0}$\\$\pm0.0$}\\
Camera translation & 13 & \shortstack{$78.5$\\$\pm10.0$} & \shortstack{$64.6$\\$\pm8.8$} & \shortstack{$56.4$\\$\pm16.0$} & \shortstack{$89.2$\\$\pm6.9$} & \shortstack{$84.6$\\$\pm5.4$} & \shortstack{$20.0$\\$\pm6.9$} & \shortstack{$35.4$\\$\pm12.9$} & \shortstack{$78.5$\\$\pm6.4$} & \shortstack{$27.7$\\$\pm14.0$} & \shortstack{$\mathbf{93.8}$\\$\pm3.4$}\\
Imagined observer & 2 & \shortstack{$\mathbf{90.0}$\\$\pm22.4$} & \shortstack{$70.0$\\$\pm27.4$} & \shortstack{$50.0$\\$\pm0.0$} & \shortstack{$80.0$\\$\pm27.4$} & \shortstack{$70.0$\\$\pm27.4$} & \shortstack{$30.0$\\$\pm27.4$} & \shortstack{$30.0$\\$\pm27.4$} & \shortstack{$60.0$\\$\pm22.4$} & \shortstack{$60.0$\\$\pm22.4$} & \shortstack{$60.0$\\$\pm22.4$}\\
Object relations & 12 & \shortstack{$96.7$\\$\pm4.6$} & \shortstack{$63.3$\\$\pm12.6$} & \shortstack{$91.7$\\$\pm8.3$} & \shortstack{$85.0$\\$\pm3.7$} & \shortstack{$95.0$\\$\pm7.5$} & \shortstack{$31.7$\\$\pm7.0$} & \shortstack{$35.0$\\$\pm3.7$} & \shortstack{$86.7$\\$\pm20.9$} & \shortstack{$45.0$\\$\pm9.5$} & \shortstack{$\mathbf{98.3}$\\$\pm3.7$}\\
Composed rotation & 5 & \shortstack{$\mathbf{100.0}$\\$\pm0.0$} & \shortstack{$96.0$\\$\pm8.9$} & \shortstack{$\mathbf{100.0}$\\$\pm0.0$} & \shortstack{$\mathbf{100.0}$\\$\pm0.0$} & \shortstack{$\mathbf{100.0}$\\$\pm0.0$} & \shortstack{$\mathbf{100.0}$\\$\pm0.0$} & \shortstack{$96.0$\\$\pm8.9$} & \shortstack{$\mathbf{100.0}$\\$\pm0.0$} & \shortstack{$96.0$\\$\pm8.9$} & \shortstack{$\mathbf{100.0}$\\$\pm0.0$}\\
Turn then translate & 11 & \shortstack{$\mathbf{90.9}$\\$\pm6.4$} & \shortstack{$43.6$\\$\pm7.6$} & \shortstack{$75.8$\\$\pm10.5$} & \shortstack{$\mathbf{90.9}$\\$\pm6.4$} & \shortstack{$87.3$\\$\pm10.4$} & \shortstack{$49.1$\\$\pm13.8$} & \shortstack{$58.2$\\$\pm10.4$} & \shortstack{$80.0$\\$\pm11.9$} & \shortstack{$47.3$\\$\pm4.1$} & \shortstack{$85.5$\\$\pm5.0$}\\
\addlinespace
\multicolumn{12}{@{}l@{}}{\colorbox[HTML]{E8EDF1}{\parbox{\dimexpr\linewidth-2\fboxsep\relax}{\textbf{MMSI}\hfill 20 cases}}}\\*
Attribute (Appr.) & 2 & \shortstack{$\mathbf{90.0}$\\$\pm22.4$} & \shortstack{$70.0$\\$\pm27.4$} & \shortstack{$50.0$\\$\pm0.0$} & \shortstack{$60.0$\\$\pm22.4$} & \shortstack{$50.0$\\$\pm0.0$} & \shortstack{$70.0$\\$\pm27.4$} & \shortstack{$40.0$\\$\pm22.4$} & \shortstack{$70.0$\\$\pm27.4$} & \shortstack{$60.0$\\$\pm41.8$} & \shortstack{$50.0$\\$\pm0.0$}\\
Attribute (Meas.) & 2 & \shortstack{$\mathbf{100.0}$\\$\pm0.0$} & \shortstack{$50.0$\\$\pm0.0$} & \shortstack{$66.7$\\$\pm57.7$} & \shortstack{$40.0$\\$\pm22.4$} & \shortstack{$50.0$\\$\pm0.0$} & \shortstack{$0.0$\\$\pm0.0$} & \shortstack{$40.0$\\$\pm22.4$} & \shortstack{$70.0$\\$\pm27.4$} & \shortstack{$80.0$\\$\pm27.4$} & \shortstack{$80.0$\\$\pm27.4$}\\
MSR & 2 & \shortstack{$40.0$\\$\pm41.8$} & \shortstack{$30.0$\\$\pm27.4$} & \shortstack{$\mathbf{50.0}$\\$\pm0.0$} & \shortstack{$40.0$\\$\pm22.4$} & \shortstack{$30.0$\\$\pm27.4$} & \shortstack{$30.0$\\$\pm44.7$} & \shortstack{$40.0$\\$\pm22.4$} & \shortstack{$\mathbf{50.0}$\\$\pm0.0$} & \shortstack{$\mathbf{50.0}$\\$\pm0.0$} & \shortstack{$40.0$\\$\pm41.8$}\\
Motion (Cam.) & 1 & \shortstack{$80.0$\\$\pm44.7$} & \shortstack{$80.0$\\$\pm44.7$} & \shortstack{$\mathbf{100.0}$\\$\pm0.0$} & \shortstack{$\mathbf{100.0}$\\$\pm0.0$} & \shortstack{$60.0$\\$\pm54.8$} & \shortstack{$0.0$\\$\pm0.0$} & \shortstack{$0.0$\\$\pm0.0$} & \shortstack{$60.0$\\$\pm54.8$} & \shortstack{$60.0$\\$\pm54.8$} & \shortstack{$\mathbf{100.0}$\\$\pm0.0$}\\
Motion (Obj.) & 1 & \shortstack{$\mathbf{100.0}$\\$\pm0.0$} & \shortstack{$80.0$\\$\pm44.7$} & \shortstack{$\mathbf{100.0}$\\$\pm0.0$} & \shortstack{$80.0$\\$\pm44.7$} & \shortstack{$80.0$\\$\pm44.7$} & \shortstack{$80.0$\\$\pm44.7$} & \shortstack{$20.0$\\$\pm44.7$} & \shortstack{$\mathbf{100.0}$\\$\pm0.0$} & \shortstack{$80.0$\\$\pm44.7$} & \shortstack{$\mathbf{100.0}$\\$\pm0.0$}\\
Camera--camera relation & 3 & \shortstack{$\mathbf{100.0}$\\$\pm0.0$} & \shortstack{$73.3$\\$\pm27.9$} & \shortstack{$\mathbf{100.0}$\\$\pm0.0$} & \shortstack{$66.7$\\$\pm0.0$} & \shortstack{$73.3$\\$\pm14.9$} & \shortstack{$26.7$\\$\pm14.9$} & \shortstack{$33.3$\\$\pm0.0$} & \shortstack{$66.7$\\$\pm40.8$} & \shortstack{$80.0$\\$\pm18.3$} & \shortstack{$73.3$\\$\pm14.9$}\\
Camera--object relation & 1 & \shortstack{$0.0$\\$\pm0.0$} & \shortstack{$40.0$\\$\pm54.8$} & \shortstack{$66.7$\\$\pm57.7$} & \shortstack{$0.0$\\$\pm0.0$} & \shortstack{$60.0$\\$\pm54.8$} & \shortstack{$40.0$\\$\pm54.8$} & \shortstack{$0.0$\\$\pm0.0$} & \shortstack{$0.0$\\$\pm0.0$} & \shortstack{$20.0$\\$\pm44.7$} & \shortstack{$\mathbf{80.0}$\\$\pm44.7$}\\
Camera--region relation & 3 & \shortstack{$\mathbf{100.0}$\\$\pm0.0$} & \shortstack{$60.0$\\$\pm14.9$} & \shortstack{$77.8$\\$\pm19.2$} & \shortstack{$66.7$\\$\pm0.0$} & \shortstack{$73.3$\\$\pm14.9$} & \shortstack{$66.7$\\$\pm23.6$} & \shortstack{$53.3$\\$\pm18.3$} & \shortstack{$66.7$\\$\pm23.6$} & \shortstack{$46.7$\\$\pm18.3$} & \shortstack{$\mathbf{100.0}$\\$\pm0.0$}\\
Object--object relation & 1 & \shortstack{$\mathbf{100.0}$\\$\pm0.0$} & \shortstack{$80.0$\\$\pm44.7$} & \shortstack{$\mathbf{100.0}$\\$\pm0.0$} & \shortstack{$60.0$\\$\pm54.8$} & \shortstack{$20.0$\\$\pm44.7$} & \shortstack{$80.0$\\$\pm44.7$} & \shortstack{$60.0$\\$\pm54.8$} & \shortstack{$\mathbf{100.0}$\\$\pm0.0$} & \shortstack{$80.0$\\$\pm44.7$} & \shortstack{$\mathbf{100.0}$\\$\pm0.0$}\\
Object--region relation & 4 & \shortstack{$\mathbf{50.0}$\\$\pm0.0$} & \shortstack{$30.0$\\$\pm11.2$} & \shortstack{$25.0$\\$\pm25.0$} & \shortstack{$45.0$\\$\pm20.9$} & \shortstack{$\mathbf{50.0}$\\$\pm30.6$} & \shortstack{$5.0$\\$\pm11.2$} & \shortstack{$0.0$\\$\pm0.0$} & \shortstack{$35.0$\\$\pm13.7$} & \shortstack{$15.0$\\$\pm22.4$} & \shortstack{$\mathbf{50.0}$\\$\pm0.0$}\\
\addlinespace
\multicolumn{12}{@{}l@{}}{\colorbox[HTML]{E8EDF1}{\parbox{\dimexpr\linewidth-2\fboxsep\relax}{\textbf{RoboSpatial}\hfill 30 cases}}}\\*
Placement compatibility & 10 & \shortstack{$92.0$\\$\pm4.5$} & \shortstack{$92.0$\\$\pm8.4$} & \shortstack{$\mathbf{96.7}$\\$\pm5.8$} & \shortstack{$90.0$\\$\pm0.0$} & \shortstack{$64.0$\\$\pm8.9$} & \shortstack{$80.0$\\$\pm7.1$} & \shortstack{$90.0$\\$\pm7.1$} & \shortstack{$94.0$\\$\pm13.4$} & \shortstack{$92.0$\\$\pm13.0$} & \shortstack{$82.0$\\$\pm8.4$}\\
Configuration & 10 & \shortstack{$78.0$\\$\pm4.5$} & \shortstack{$70.0$\\$\pm10.0$} & \shortstack{$80.0$\\$\pm0.0$} & \shortstack{$86.0$\\$\pm11.4$} & \shortstack{$88.0$\\$\pm4.5$} & \shortstack{$\mathbf{100.0}$\\$\pm0.0$} & \shortstack{$94.0$\\$\pm8.9$} & \shortstack{$90.0$\\$\pm0.0$} & \shortstack{$82.0$\\$\pm8.4$} & \shortstack{$88.0$\\$\pm4.5$}\\
Free-space point set & 10 & \shortstack{$\mathbf{0.0}$\\$\pm0.0$} & \shortstack{$\mathbf{0.0}$\\$\pm0.0$} & \shortstack{$\mathbf{0.0}$\\$\pm0.0$} & \shortstack{$\mathbf{0.0}$\\$\pm0.0$} & \shortstack{$\mathbf{0.0}$\\$\pm0.0$} & \shortstack{$\mathbf{0.0}$\\$\pm0.0$} & \shortstack{$\mathbf{0.0}$\\$\pm0.0$} & \shortstack{$\mathbf{0.0}$\\$\pm0.0$} & \shortstack{$\mathbf{0.0}$\\$\pm0.0$} & \shortstack{$\mathbf{0.0}$\\$\pm0.0$}\\
\addlinespace
\multicolumn{12}{@{}l@{}}{\colorbox[HTML]{E8EDF1}{\parbox{\dimexpr\linewidth-2\fboxsep\relax}{\textbf{VSI-Bench}\hfill 50 cases}}}\\*
Appearance order & 2 & \shortstack{$40.0$\\$\pm22.4$} & \shortstack{$80.0$\\$\pm27.4$} & \shortstack{$50.0$\\$\pm50.0$} & \shortstack{$\mathbf{100.0}$\\$\pm0.0$} & \shortstack{$40.0$\\$\pm41.8$} & \shortstack{$40.0$\\$\pm22.4$} & \shortstack{$80.0$\\$\pm27.4$} & \shortstack{$60.0$\\$\pm22.4$} & \shortstack{$10.0$\\$\pm22.4$} & \shortstack{$60.0$\\$\pm22.4$}\\
Absolute distance & 9 & \shortstack{$2.2$\\$\pm5.0$} & \shortstack{$2.2$\\$\pm5.0$} & \shortstack{$3.7$\\$\pm6.4$} & \shortstack{$0.0$\\$\pm0.0$} & \shortstack{$\mathbf{6.7}$\\$\pm6.1$} & \shortstack{$0.0$\\$\pm0.0$} & \shortstack{$2.2$\\$\pm5.0$} & \shortstack{$2.2$\\$\pm5.0$} & \shortstack{$0.0$\\$\pm0.0$} & \shortstack{$0.0$\\$\pm0.0$}\\
Object count & 5 & \shortstack{$32.0$\\$\pm11.0$} & \shortstack{$20.0$\\$\pm14.1$} & \shortstack{$40.0$\\$\pm20.0$} & \shortstack{$36.0$\\$\pm16.7$} & \shortstack{$36.0$\\$\pm16.7$} & \shortstack{$28.0$\\$\pm11.0$} & \shortstack{$12.0$\\$\pm11.0$} & \shortstack{$\mathbf{60.0}$\\$\pm20.0$} & \shortstack{$16.0$\\$\pm8.9$} & \shortstack{$32.0$\\$\pm11.0$}\\
Direction: easy & 4 & \shortstack{$\mathbf{100.0}$\\$\pm0.0$} & \shortstack{$85.0$\\$\pm22.4$} & \shortstack{$33.3$\\$\pm28.9$} & \shortstack{$\mathbf{100.0}$\\$\pm0.0$} & \shortstack{$85.0$\\$\pm13.7$} & \shortstack{$30.0$\\$\pm20.9$} & \shortstack{$45.0$\\$\pm11.2$} & \shortstack{$80.0$\\$\pm11.2$} & \shortstack{$50.0$\\$\pm17.7$} & \shortstack{$65.0$\\$\pm22.4$}\\
Direction: hard & 4 & \shortstack{$45.0$\\$\pm11.2$} & \shortstack{$30.0$\\$\pm27.4$} & \shortstack{$25.0$\\$\pm43.3$} & \shortstack{$\mathbf{65.0}$\\$\pm28.5$} & \shortstack{$60.0$\\$\pm22.4$} & \shortstack{$30.0$\\$\pm32.6$} & \shortstack{$5.0$\\$\pm11.2$} & \shortstack{$40.0$\\$\pm13.7$} & \shortstack{$0.0$\\$\pm0.0$} & \shortstack{$30.0$\\$\pm32.6$}\\
Direction: medium & 6 & \shortstack{$\mathbf{100.0}$\\$\pm0.0$} & \shortstack{$73.3$\\$\pm9.1$} & \shortstack{$38.9$\\$\pm38.5$} & \shortstack{$83.3$\\$\pm11.8$} & \shortstack{$66.7$\\$\pm11.8$} & \shortstack{$40.0$\\$\pm14.9$} & \shortstack{$20.0$\\$\pm13.9$} & \shortstack{$66.7$\\$\pm11.8$} & \shortstack{$43.3$\\$\pm14.9$} & \shortstack{$50.0$\\$\pm11.8$}\\
Relative distance & 6 & \shortstack{$50.0$\\$\pm20.4$} & \shortstack{$\mathbf{76.7}$\\$\pm22.4$} & \shortstack{$38.9$\\$\pm9.6$} & \shortstack{$\mathbf{76.7}$\\$\pm9.1$} & \shortstack{$56.7$\\$\pm14.9$} & \shortstack{$60.0$\\$\pm19.0$} & \shortstack{$53.3$\\$\pm13.9$} & \shortstack{$63.3$\\$\pm7.5$} & \shortstack{$46.7$\\$\pm21.7$} & \shortstack{$56.7$\\$\pm14.9$}\\
Object size & 9 & \shortstack{$4.4$\\$\pm6.1$} & \shortstack{$\mathbf{22.2}$\\$\pm7.9$} & \shortstack{$11.1$\\$\pm0.0$} & \shortstack{$13.3$\\$\pm5.0$} & \shortstack{$11.1$\\$\pm7.9$} & \shortstack{$2.2$\\$\pm5.0$} & \shortstack{$8.9$\\$\pm5.0$} & \shortstack{$2.2$\\$\pm5.0$} & \shortstack{$6.7$\\$\pm6.1$} & \shortstack{$\mathbf{22.2}$\\$\pm7.9$}\\
Room area & 4 & \shortstack{$\mathbf{0.0}$\\$\pm0.0$} & \shortstack{$\mathbf{0.0}$\\$\pm0.0$} & \shortstack{$\mathbf{0.0}$\\$\pm0.0$} & \shortstack{$\mathbf{0.0}$\\$\pm0.0$} & \shortstack{$\mathbf{0.0}$\\$\pm0.0$} & \shortstack{$\mathbf{0.0}$\\$\pm0.0$} & \shortstack{$\mathbf{0.0}$\\$\pm0.0$} & \shortstack{$\mathbf{0.0}$\\$\pm0.0$} & \shortstack{$\mathbf{0.0}$\\$\pm0.0$} & \shortstack{$\mathbf{0.0}$\\$\pm0.0$}\\
Route planning & 1 & \shortstack{$\mathbf{100.0}$\\$\pm0.0$} & \shortstack{$\mathbf{100.0}$\\$\pm0.0$} & \shortstack{$33.3$\\$\pm57.7$} & \shortstack{$\mathbf{100.0}$\\$\pm0.0$} & \shortstack{$\mathbf{100.0}$\\$\pm0.0$} & \shortstack{$\mathbf{100.0}$\\$\pm0.0$} & \shortstack{$\mathbf{100.0}$\\$\pm0.0$} & \shortstack{$\mathbf{100.0}$\\$\pm0.0$} & \shortstack{$\mathbf{100.0}$\\$\pm0.0$} & \shortstack{$\mathbf{100.0}$\\$\pm0.0$}\\
\end{longtable}}
\begin{table}[!htbp]
\centering
\caption{VSI-Bench numerical estimation. (a) AbsRel, equal-run mean $\pm$ SD on each model's valid predictions; lower is better. (b) Astra's coverage and fixed-cohort sensitivity: common cases have valid predictions in every recorded round.}\label{tab:spatial-numeric-models}\label{tab:spatial-numeric-coverage}
\noindent{\ArenaTableFont\textbf{(a) Numerical error across agents}\par}\smallskip
\centering

\ArenaTableFont\setlength{\tabcolsep}{2pt}
\begin{tabular*}{\linewidth}{@{\extracolsep{\fill}}l*{4}{r}@{}}\toprule
Model & Object count & Abs. distance & Object size & Room area\\\midrule
Astra & $0.35\!\pm\!0.05$ & $0.28\!\pm\!0.11$ & $0.09\!\pm\!0.01$ & $0.24\!\pm\!0.03$\\
Sol & $0.40\!\pm\!0.11$ & $0.42\!\pm\!0.05$ & $0.09\!\pm\!0.02$ & $0.19\!\pm\!0.03$\\
Fable & $0.20\!\pm\!0.11$ & $0.28\!\pm\!0.07$ & $0.10\!\pm\!0.01$ & $0.21\!\pm\!0.04$\\
Q-Max & $0.23\!\pm\!0.11$ & $0.26\!\pm\!0.02$ & $0.12\!\pm\!0.02$ & $0.26\!\pm\!0.07$\\
Gemini~3.8 & $0.31\!\pm\!0.08$ & $0.12\!\pm\!0.03$ & $0.05\!\pm\!0.01$ & $0.27\!\pm\!0.04$\\
Q-397B & $0.41\!\pm\!0.06$ & $0.29\!\pm\!0.09$ & $0.13\!\pm\!0.02$ & $0.22\!\pm\!0.05$\\
Q-27B & $0.42\!\pm\!0.04$ & $0.31\!\pm\!0.08$ & $0.19\!\pm\!0.02$ & $0.26\!\pm\!0.10$\\
Opus~5.5 & $0.16\!\pm\!0.09$ & $0.30\!\pm\!0.07$ & $0.09\!\pm\!0.01$ & $0.21\!\pm\!0.06$\\
DS-1017 & $0.34\!\pm\!0.09$ & $0.48\!\pm\!0.09$ & $0.14\!\pm\!0.07$ & $0.35\!\pm\!0.12$\\
Gemini~4 & $0.25\!\pm\!0.07$ & $0.27\!\pm\!0.06$ & $0.07\!\pm\!0.02$ & $0.19\!\pm\!0.03$\\
\bottomrule\end{tabular*}
\par\medskip
\noindent{\ArenaTableFont\textbf{(b) Astra coverage and common-cohort error}\par}\smallskip
\centering

\ArenaTableFont\setlength{\tabcolsep}{3pt}
\begin{tabular*}{\linewidth}{@{\extracolsep{\fill}}lrrrr@{}}\toprule
Task & Cases & Valid / assigned & Common cases & Common AbsRel\\\midrule
Object count & 5 & 25/25 & 5 & $0.350\!\pm\!0.053$\\
Absolute distance & 9 & 25/45 & 2 & $0.279\!\pm\!0.083$\\
Object size & 9 & 42/45 & 8 & $0.091\!\pm\!0.017$\\
Room area & 4 & 20/20 & 4 & $0.240\!\pm\!0.029$\\
\bottomrule\end{tabular*}
\end{table}
\noindent\textbf{Other task requirements.}
RoboSpatial separates placement compatibility (92.0\%), configuration (78.0\%), and free-space point sets (0.0\%) for Astra. All agents in this table fail the point-set criterion. Compatibility and configuration judgments succeed more often than \mbox{point-set predictions.}

\subsection{Affordance: Contact and Region Extent}
\label{app:affordance-focus}
\input{sections/supplement_figure6_affordance}

\noindent\textbf{Point validity and functional-region localization.}
Table~\ref{tab:astra-affordance-subtasks} reports all fifteen action/hand groups for Astra, \mbox{DeepSeek-1017}, Opus~5.5, and Gemini~4~Argon, including zero-scoring conditions. UMD contact validity averages 96.7\% for Astra and 94.4\% for Fable. Cutting combines 100\% valid contacts and 96\% strict box success with box--mask IoU of 0.361; the box IoU is 0.694. High point validity and box--box overlap can therefore coexist with much lower box--mask overlap on irregular regions. Containment adds a same-representation comparison: Astra's strict box success is 44\%, versus 100\% for both Opus~5.5 and Gemini~4~Argon (Table~\ref{tab:affordance-peers}).

\noindent\textbf{Contact, extent, and shape give different comparisons.}
In the original cohort, Astra leads box--mask IoU on UMD cutting, grasping, scooping, and support, while Sol leads containment and pounding. With all ten agents, Gemini~4~Argon leads mask IoU on cutting (0.399) and containment (0.755), yet its ReasonAff pulling boxes never cross the strict success threshold despite 45\% valid contacts (Table~\ref{tab:affordance-peers}). High overlap on tool regions therefore does not imply accurate localization in interaction-conditioned scenes. In the interaction-conditioned sources, one-handed pulling has zero strict box success for every original model despite nonzero contact validity. Table~\ref{tab:affordance-peers} identifies the best peer separately for each reported criterion. ReasonAff and RAGNet retain their original \mbox{scenes and interaction annotations.}

{\ArenaTableFont
\setlength{\tabcolsep}{5pt}
\begin{longtable}{@{}p{.26\linewidth}lrrrr@{}}
\caption{Astra, \mbox{DeepSeek-1017}, Opus~5.5, and Gemini~4~Argon Affordance metrics, mean $\pm$ run SD. IoU compares the native box to the GT mask; strict box success compares it to the GT box at IoU $\geq0.5$. Point and box \mbox{success are percentages.}}\label{tab:astra-affordance-subtasks}\\
\toprule
Action / hands & Model & Cases & Box--mask IoU & Point (\%) & Box (\%)\\\midrule\endfirsthead
\multicolumn{6}{l}{\textit{Continued from previous page.}}\\\toprule
Action / hands & Model & Cases & Box--mask IoU & Point (\%) & Box (\%)\\\midrule\endhead
\bottomrule\endfoot
\multicolumn{6}{l}{\textit{RAGNet-3DOI}}\\*
free movement / two hands & Astra & 2 & $0.046\!\pm\!0.063$ & $20.0\!\pm\!27.4$ & $20.0\!\pm\!27.4$\\
 & \mbox{DeepSeek-1017} & 2 & $0.179\!\pm\!0.070$ & $50.0\!\pm\!0.0$ & $30.0\!\pm\!27.4$\\
 & Opus~5.5 & 2 & $0.000\!\pm\!0.000$ & $0.0\!\pm\!0.0$ & $0.0\!\pm\!0.0$\\
 & Gemini~4 & 2 & $0.000\!\pm\!0.000$ & $0.0\!\pm\!0.0$ & $0.0\!\pm\!0.0$\\
free / one hand & Astra & 1 & $0.384\!\pm\!0.351$ & $60.0\!\pm\!54.8$ & $60.0\!\pm\!54.8$\\
 & \mbox{DeepSeek-1017} & 1 & $0.241\!\pm\!0.291$ & $40.0\!\pm\!54.8$ & $20.0\!\pm\!44.7$\\
 & Opus~5.5 & 1 & $0.000\!\pm\!0.000$ & $0.0\!\pm\!0.0$ & $0.0\!\pm\!0.0$\\
 & Gemini~4 & 1 & $0.669\!\pm\!0.010$ & $100.0\!\pm\!0.0$ & $100.0\!\pm\!0.0$\\
free / two hands & Astra & 5 & $0.249\!\pm\!0.070$ & $60.0\!\pm\!14.1$ & $36.0\!\pm\!16.7$\\
 & \mbox{DeepSeek-1017} & 5 & $0.083\!\pm\!0.030$ & $28.0\!\pm\!11.0$ & $4.0\!\pm\!8.9$\\
 & Opus~5.5 & 5 & $0.429\!\pm\!0.153$ & $80.0\!\pm\!24.5$ & $60.0\!\pm\!24.5$\\
 & Gemini~4 & 5 & $0.341\!\pm\!0.135$ & $80.0\!\pm\!20.0$ & $44.0\!\pm\!16.7$\\
pull / one hand & Astra & 12 & $0.008\!\pm\!0.002$ & $25.0\!\pm\!5.9$ & $0.0\!\pm\!0.0$\\
 & \mbox{DeepSeek-1017} & 12 & $0.053\!\pm\!0.019$ & $41.7\!\pm\!8.3$ & $0.0\!\pm\!0.0$\\
 & Opus~5.5 & 12 & $0.048\!\pm\!0.047$ & $31.7\!\pm\!3.7$ & $5.0\!\pm\!7.5$\\
 & Gemini~4 & 12 & $0.206\!\pm\!0.059$ & $23.3\!\pm\!3.7$ & $26.7\!\pm\!7.0$\\
\multicolumn{6}{l}{\textit{ReasonAff-style}}\\*
free movement / two hands & Astra & 5 & $0.270\!\pm\!0.055$ & $48.0\!\pm\!11.0$ & $48.0\!\pm\!11.0$\\
 & \mbox{DeepSeek-1017} & 5 & $0.026\!\pm\!0.028$ & $0.0\!\pm\!0.0$ & $0.0\!\pm\!0.0$\\
 & Opus~5.5 & 5 & $0.112\!\pm\!0.019$ & $20.0\!\pm\!0.0$ & $20.0\!\pm\!0.0$\\
 & Gemini~4 & 5 & $0.256\!\pm\!0.058$ & $40.0\!\pm\!14.1$ & $48.0\!\pm\!11.0$\\
free / one hand & Astra & 4 & $0.000\!\pm\!0.000$ & $0.0\!\pm\!0.0$ & $0.0\!\pm\!0.0$\\
 & \mbox{DeepSeek-1017} & 4 & $0.035\!\pm\!0.040$ & $0.0\!\pm\!0.0$ & $0.0\!\pm\!0.0$\\
 & Opus~5.5 & 4 & $0.085\!\pm\!0.090$ & $10.0\!\pm\!13.7$ & $10.0\!\pm\!13.7$\\
 & Gemini~4 & 4 & $0.065\!\pm\!0.090$ & $10.0\!\pm\!13.7$ & $10.0\!\pm\!13.7$\\
free / two hands & Astra & 4 & $0.001\!\pm\!0.002$ & $0.0\!\pm\!0.0$ & $0.0\!\pm\!0.0$\\
 & \mbox{DeepSeek-1017} & 4 & $0.004\!\pm\!0.008$ & $0.0\!\pm\!0.0$ & $0.0\!\pm\!0.0$\\
 & Opus~5.5 & 4 & $0.174\!\pm\!0.094$ & $20.0\!\pm\!11.2$ & $20.0\!\pm\!11.2$\\
 & Gemini~4 & 4 & $0.096\!\pm\!0.004$ & $25.0\!\pm\!0.0$ & $25.0\!\pm\!0.0$\\
pull / one hand & Astra & 4 & $0.098\!\pm\!0.033$ & $60.0\!\pm\!13.7$ & $0.0\!\pm\!0.0$\\
 & \mbox{DeepSeek-1017} & 4 & $0.086\!\pm\!0.047$ & $55.0\!\pm\!11.2$ & $0.0\!\pm\!0.0$\\
 & Opus~5.5 & 4 & $0.083\!\pm\!0.002$ & $50.0\!\pm\!0.0$ & $0.0\!\pm\!0.0$\\
 & Gemini~4 & 4 & $0.093\!\pm\!0.039$ & $45.0\!\pm\!20.9$ & $0.0\!\pm\!0.0$\\
push / one hand & Astra & 3 & $0.318\!\pm\!0.221$ & $46.7\!\pm\!29.8$ & $46.7\!\pm\!29.8$\\
 & \mbox{DeepSeek-1017} & 3 & $0.146\!\pm\!0.201$ & $26.7\!\pm\!27.9$ & $20.0\!\pm\!29.8$\\
 & Opus~5.5 & 3 & $0.448\!\pm\!0.108$ & $66.7\!\pm\!0.0$ & $60.0\!\pm\!14.9$\\
 & Gemini~4 & 3 & $0.280\!\pm\!0.048$ & $33.3\!\pm\!0.0$ & $33.3\!\pm\!0.0$\\
\multicolumn{6}{l}{\textit{UMD}}\\*
contain & Astra & 5 & $0.472\!\pm\!0.067$ & $100.0\!\pm\!0.0$ & $44.0\!\pm\!16.7$\\*
 & \mbox{DeepSeek-1017} & 5 & $0.593\!\pm\!0.045$ & $96.0\!\pm\!8.9$ & $84.0\!\pm\!16.7$\\
 & Opus~5.5 & 5 & $0.649\!\pm\!0.009$ & $100.0\!\pm\!0.0$ & $100.0\!\pm\!0.0$\\
 & Gemini~4 & 5 & $0.755\!\pm\!0.009$ & $100.0\!\pm\!0.0$ & $100.0\!\pm\!0.0$\\
cut & Astra & 5 & $0.361\!\pm\!0.009$ & $100.0\!\pm\!0.0$ & $96.0\!\pm\!8.9$\\*
 & \mbox{DeepSeek-1017} & 5 & $0.223\!\pm\!0.019$ & $48.0\!\pm\!30.3$ & $52.0\!\pm\!17.9$\\
 & Opus~5.5 & 5 & $0.349\!\pm\!0.012$ & $100.0\!\pm\!0.0$ & $100.0\!\pm\!0.0$\\
 & Gemini~4 & 5 & $0.399\!\pm\!0.009$ & $100.0\!\pm\!0.0$ & $92.0\!\pm\!11.0$\\
grasp & Astra & 5 & $0.491\!\pm\!0.007$ & $100.0\!\pm\!0.0$ & $100.0\!\pm\!0.0$\\*
 & \mbox{DeepSeek-1017} & 5 & $0.377\!\pm\!0.063$ & $92.0\!\pm\!11.0$ & $64.0\!\pm\!29.7$\\
 & Opus~5.5 & 5 & $0.477\!\pm\!0.005$ & $100.0\!\pm\!0.0$ & $100.0\!\pm\!0.0$\\
 & Gemini~4 & 5 & $0.430\!\pm\!0.012$ & $76.0\!\pm\!8.9$ & $100.0\!\pm\!0.0$\\
pound & Astra & 5 & $0.637\!\pm\!0.009$ & $100.0\!\pm\!0.0$ & $100.0\!\pm\!0.0$\\*
 & \mbox{DeepSeek-1017} & 5 & $0.490\!\pm\!0.034$ & $96.0\!\pm\!8.9$ & $44.0\!\pm\!16.7$\\
 & Opus~5.5 & 5 & $0.585\!\pm\!0.042$ & $96.0\!\pm\!8.9$ & $96.0\!\pm\!8.9$\\
 & Gemini~4 & 5 & $0.644\!\pm\!0.009$ & $100.0\!\pm\!0.0$ & $92.0\!\pm\!11.0$\\
scoop & Astra & 5 & $0.504\!\pm\!0.054$ & $80.0\!\pm\!0.0$ & $80.0\!\pm\!0.0$\\*
 & \mbox{DeepSeek-1017} & 5 & $0.337\!\pm\!0.059$ & $76.0\!\pm\!16.7$ & $52.0\!\pm\!17.9$\\
 & Opus~5.5 & 5 & $0.486\!\pm\!0.011$ & $80.0\!\pm\!0.0$ & $80.0\!\pm\!0.0$\\
 & Gemini~4 & 5 & $0.514\!\pm\!0.010$ & $80.0\!\pm\!0.0$ & $80.0\!\pm\!0.0$\\
support & Astra & 5 & $0.638\!\pm\!0.017$ & $100.0\!\pm\!0.0$ & $100.0\!\pm\!0.0$\\
 & \mbox{DeepSeek-1017} & 5 & $0.466\!\pm\!0.056$ & $88.0\!\pm\!17.9$ & $76.0\!\pm\!16.7$\\
 & Opus~5.5 & 5 & $0.608\!\pm\!0.030$ & $100.0\!\pm\!0.0$ & $92.0\!\pm\!11.0$\\
 & Gemini~4 & 5 & $0.644\!\pm\!0.011$ & $100.0\!\pm\!0.0$ & $100.0\!\pm\!0.0$\\
\end{longtable}
}

\begin{table}[!htbp]\centering
\caption{Criterion-specific Affordance comparisons. One highest-mean non-Astra peer is selected from all ten agents for each row; ties can share that score. Cells give mean $\pm$ run SD. Box success uses GT-box IoU $\geq0.5$; mask IoU uses the GT mask. 2H denotes two hands.}\label{tab:affordance-peers}
\ArenaTableFont\setlength{\tabcolsep}{3pt}
\begin{tabular*}{\linewidth}{@{\extracolsep{\fill}}lrrrr@{}}\toprule
Task / criterion & Astra & Best peer & Opus~5.5 & Gemini~4\\\midrule
UMD cutting: mask IoU & \shortstack{$0.361$\\$\pm0.009$} & \shortstack{Gemini~4\\$0.399\!\pm\!0.009$} & \shortstack{$0.349$\\$\pm0.012$} & \shortstack{$0.399$\\$\pm0.009$}\\
UMD cutting: box (\%) & \shortstack{$96.0$\\$\pm8.9$} & \shortstack{Fable\\$100.0\!\pm\!0.0$} & \shortstack{$100.0$\\$\pm0.0$} & \shortstack{$92.0$\\$\pm11.0$}\\
UMD containment: mask IoU & \shortstack{$0.472$\\$\pm0.067$} & \shortstack{Gemini~4\\$0.755\!\pm\!0.009$} & \shortstack{$0.649$\\$\pm0.009$} & \shortstack{$0.755$\\$\pm0.009$}\\
UMD containment: box (\%) & \shortstack{$44.0$\\$\pm16.7$} & \shortstack{Opus~5.5\\$100.0\!\pm\!0.0$} & \shortstack{$100.0$\\$\pm0.0$} & \shortstack{$100.0$\\$\pm0.0$}\\
ReasonAff pulling: point (\%) & \shortstack{$60.0$\\$\pm13.7$} & \shortstack{DS-1017\\$55.0\!\pm\!11.2$} & \shortstack{$50.0$\\$\pm0.0$} & \shortstack{$45.0$\\$\pm20.9$}\\
RAGNet pulling: point (\%) & \shortstack{$25.0$\\$\pm5.9$} & \shortstack{DS-1017\\$41.7\!\pm\!8.3$} & \shortstack{$31.7$\\$\pm3.7$} & \shortstack{$23.3$\\$\pm3.7$}\\
ReasonAff movement, 2H: point (\%) & \shortstack{$48.0$\\$\pm11.0$} & \shortstack{Gemini~4\\$40.0\!\pm\!14.1$} & \shortstack{$20.0$\\$\pm0.0$} & \shortstack{$40.0$\\$\pm14.1$}\\
RAGNet movement, 2H: point (\%) & \shortstack{$20.0$\\$\pm27.4$} & \shortstack{Sol\\$50.0\!\pm\!0.0$} & \shortstack{$0.0$\\$\pm0.0$} & \shortstack{$0.0$\\$\pm0.0$}\\
\bottomrule\end{tabular*}\end{table}

\subsection{Task Planning: Goal Families and Terminal States}
\label{app:planning-focus}
\noindent\textbf{Compound goals and source-specific requirements.}
Table~\ref{tab:astra-planning-subtasks} covers all 56 native families across the six sources. Each row compares all ten agents on identical assigned task IDs. Figure~\ref{fig:planning-detail}a compares five models by source; panels b--c hold Fable fixed for household and scientific goal comparisons. Heating and dual placement test state change and multiple-object goal retention; ScienceWorld distinguishes experimental topics and instrument use. This preserves the task-specific advantages and reversals behind the aggregate Planning lead.

{\ArenaDenseTableFont
\setlength{\tabcolsep}{1.3pt}
\begin{longtable}{@{}p{150pt}*{10}{r}@{}}
\caption{Task Planning by native goal family, with one episode per assigned case. Entries are solved/assigned cases. All ten models use the same assigned task IDs within each family. Source-level results appear in Table~\ref{tab:source-comparisons}.}\label{tab:astra-planning-subtasks}\\
\toprule
Goal family & Astra & Sol & Fable & Q-Max & Gemini~3.8 & Q-397B & Q-27B & Opus~5.5 & DS-1017 & Gemini~4\\\midrule\endfirsthead
\multicolumn{11}{l}{\textit{Continued from previous page.}}\\\toprule
Goal family & Astra & Sol & Fable & Q-Max & Gemini~3.8 & Q-397B & Q-27B & Opus~5.5 & DS-1017 & Gemini~4\\\midrule\endhead
\bottomrule\endfoot
\multicolumn{11}{l}{\textit{ALFRED}}\\*
Illumination & 1/2 & 0/2 & 1/2 & 1/2 & 0/2 & 1/2 & 0/2 & 0/2 & 0/2 & 1/2\\
Placement & 4/4 & 2/4 & 2/4 & 0/4 & 0/4 & 0/4 & 0/4 & 0/4 & 0/4 & 4/4\\
Container transport & 2/2 & 0/2 & 1/2 & 1/2 & 1/2 & 0/2 & 0/2 & 1/2 & 1/2 & 2/2\\
Cleaning & 2/2 & 1/2 & 1/2 & 1/2 & 1/2 & 0/2 & 0/2 & 1/2 & 0/2 & 2/2\\
Cooling & 2/4 & 0/4 & 2/4 & 0/4 & 0/4 & 0/4 & 0/4 & 1/4 & 0/4 & 2/4\\
Heating & 2/3 & 0/3 & 0/3 & 0/3 & 0/3 & 0/3 & 0/3 & 0/3 & 0/3 & 2/3\\
Dual placement & 2/3 & 0/3 & 0/3 & 0/3 & 0/3 & 0/3 & 0/3 & 0/3 & 0/3 & 1/3\\
\multicolumn{11}{l}{\textit{ALFWorld}}\\*
Illumination & 3/3 & 0/3 & 2/3 & 0/3 & 2/3 & 1/3 & 2/3 & 3/3 & 2/3 & 3/3\\
Placement & 3/4 & 2/4 & 2/4 & 1/4 & 1/4 & 1/4 & 1/4 & 1/4 & 1/4 & 4/4\\
Container transport & 2/2 & 0/2 & 0/2 & 0/2 & 1/2 & 0/2 & 1/2 & 1/2 & 1/2 & 1/2\\
Cleaning & 3/4 & 1/4 & 1/4 & 0/4 & 0/4 & 1/4 & 0/4 & 2/4 & 0/4 & 2/4\\
Cooling & 0/1 & 0/1 & 0/1 & 0/1 & 0/1 & 0/1 & 0/1 & 0/1 & 0/1 & 1/1\\
Heating & 1/1 & 0/1 & 0/1 & 0/1 & 0/1 & 0/1 & 0/1 & 1/1 & 0/1 & 1/1\\
Dual placement & 2/5 & 1/5 & 1/5 & 1/5 & 0/5 & 0/5 & 0/5 & 1/5 & 0/5 & 4/5\\
\multicolumn{11}{l}{\textit{DiscoveryWorld}}\\*
Archaeology dating & 1/1 & 1/1 & 1/1 & 0/1 & 1/1 & 0/1 & 0/1 & 1/1 & 0/1 & 1/1\\
Combinatorial chemistry & 1/1 & 1/1 & 0/1 & 0/1 & 1/1 & 0/1 & 0/1 & 0/1 & 0/1 & 1/1\\
It's not rocket science & 1/1 & 0/1 & 1/1 & 0/1 & 0/1 & 0/1 & 0/1 & 0/1 & 0/1 & 0/1\\
Lost in translation & 1/1 & 0/1 & 1/1 & 0/1 & 0/1 & 0/1 & 0/1 & 0/1 & 0/1 & 0/1\\
Plant nutrients & 1/1 & 0/1 & 0/1 & 0/1 & 0/1 & 0/1 & 0/1 & 0/1 & 0/1 & 0/1\\
Discovery feed & 1/1 & 1/1 & 1/1 & 1/1 & 1/1 & 1/1 & 0/1 & 1/1 & 1/1 & 1/1\\
Doors & 1/1 & 0/1 & 0/1 & 1/1 & 0/1 & 0/1 & 0/1 & 0/1 & 0/1 & 0/1\\
Navigation in a house & 1/1 & 0/1 & 1/1 & 0/1 & 1/1 & 0/1 & 0/1 & 0/1 & 0/1 & 1/1\\
Pick and give & 1/1 & 1/1 & 1/1 & 1/1 & 1/1 & 0/1 & 1/1 & 1/1 & 1/1 & 1/1\\
Space sick & 1/1 & 0/1 & 1/1 & 0/1 & 0/1 & 0/1 & 0/1 & 0/1 & 0/1 & 1/1\\
\multicolumn{11}{l}{\textit{HumanCLAW}}\\*
Navigate to contact & 3/5 & 2/5 & 1/5 & 2/5 & 3/5 & 2/5 & 0/5 & 3/5 & 2/5 & 2/5\\
Sit on target & 0/8 & 0/8 & 0/8 & 0/8 & 0/8 & 0/8 & 0/8 & 0/8 & 0/8 & 0/8\\
\multicolumn{11}{l}{\textit{ScienceWorld}}\\*
Find animal & 3/3 & 1/3 & 3/3 & 1/3 & 2/3 & 0/3 & 0/3 & 1/3 & 0/3 & 3/3\\
Find living thing & 2/2 & 2/2 & 2/2 & 0/2 & 2/2 & 0/2 & 0/2 & 2/2 & 2/2 & 2/2\\
Find non living thing & 3/3 & 3/3 & 3/3 & 3/3 & 3/3 & 2/3 & 2/3 & 3/3 & 3/3 & 3/3\\
Find plant & 3/3 & 3/3 & 3/3 & 1/3 & 3/3 & 0/3 & 3/3 & 0/3 & 2/3 & 3/3\\
Grow fruit & 2/3 & 0/3 & 2/3 & 1/3 & 1/3 & 0/3 & 0/3 & 0/3 & 0/3 & 0/3\\
Inclined plane determine angle & 2/2 & 0/2 & 2/2 & 0/2 & 1/2 & 0/2 & 0/2 & 2/2 & 2/2 & 2/2\\
Inclined plane friction named surfaces & 15/16 & 7/16 & 16/16 & 5/16 & 8/16 & 4/16 & 3/16 & 13/16 & 8/16 & 11/16\\
Inclined plane friction unnamed surfaces & 1/1 & 0/1 & 1/1 & 1/1 & 1/1 & 1/1 & 1/1 & 1/1 & 0/1 & 1/1\\
Lifespan longest lived & 2/2 & 2/2 & 2/2 & 2/2 & 2/2 & 1/2 & 0/2 & 2/2 & 2/2 & 2/2\\
Lifespan longest lived then shortest lived & 1/1 & 1/1 & 1/1 & 1/1 & 1/1 & 0/1 & 0/1 & 1/1 & 1/1 & 1/1\\
Lifespan shortest lived & 1/1 & 1/1 & 1/1 & 1/1 & 1/1 & 0/1 & 1/1 & 1/1 & 1/1 & 1/1\\
Measure melting point known substance & 8/8 & 5/8 & 5/8 & 4/8 & 6/8 & 2/8 & 2/8 & 5/8 & 5/8 & 8/8\\
Measure melting point unknown substance & 1/1 & 1/1 & 1/1 & 1/1 & 0/1 & 0/1 & 0/1 & 0/1 & 0/1 & 1/1\\
Mendelian genetics known plant & 1/1 & 1/1 & 1/1 & 1/1 & 1/1 & 0/1 & 1/1 & 1/1 & 1/1 & 0/1\\
Mendelian genetics unknown plant & 3/5 & 0/5 & 0/5 & 1/5 & 1/5 & 1/5 & 0/5 & 1/5 & 0/5 & 0/5\\
Test conductivity & 6/7 & 4/7 & 5/7 & 2/7 & 6/7 & 0/7 & 0/7 & 3/7 & 0/7 & 4/7\\
Test conductivity of unknown substances & 6/8 & 3/8 & 6/8 & 3/8 & 1/8 & 0/8 & 0/8 & 3/8 & 1/8 & 4/8\\
Use thermometer & 4/6 & 4/6 & 5/6 & 1/6 & 5/6 & 0/6 & 1/6 & 4/6 & 0/6 & 6/6\\
\multicolumn{11}{l}{\textit{VirtualHome}}\\*
Grab & 1/1 & 1/1 & 1/1 & 1/1 & 1/1 & 0/1 & 0/1 & 1/1 & 0/1 & 1/1\\
Greet & 1/1 & 1/1 & 1/1 & 1/1 & 1/1 & 1/1 & 1/1 & 1/1 & 1/1 & 1/1\\
Lie & 1/1 & 1/1 & 1/1 & 1/1 & 1/1 & 1/1 & 1/1 & 1/1 & 1/1 & 1/1\\
Navigation or observation & 1/1 & 1/1 & 1/1 & 1/1 & 1/1 & 0/1 & 0/1 & 0/1 & 1/1 & 1/1\\
Read & 2/2 & 1/2 & 2/2 & 2/2 & 0/2 & 0/2 & 0/2 & 1/2 & 0/2 & 2/2\\
Rinse & 1/1 & 1/1 & 1/1 & 0/1 & 1/1 & 1/1 & 1/1 & 1/1 & 0/1 & 1/1\\
Sit & 3/3 & 3/3 & 3/3 & 3/3 & 3/3 & 3/3 & 3/3 & 2/3 & 1/3 & 3/3\\
Switchoff & 1/1 & 1/1 & 1/1 & 1/1 & 1/1 & 1/1 & 1/1 & 1/1 & 1/1 & 1/1\\
Switchon & 7/7 & 7/7 & 7/7 & 7/7 & 7/7 & 4/7 & 2/7 & 6/7 & 5/7 & 7/7\\
Touch & 1/1 & 1/1 & 1/1 & 1/1 & 1/1 & 1/1 & 1/1 & 1/1 & 1/1 & 1/1\\
Type & 1/1 & 0/1 & 0/1 & 0/1 & 0/1 & 0/1 & 0/1 & 0/1 & 0/1 & 0/1\\
Wash & 0/1 & 1/1 & 1/1 & 1/1 & 0/1 & 0/1 & 1/1 & 0/1 & 1/1 & 1/1\\
\end{longtable}

\input{sections/supplement_figure7_planning}
}

\begin{table}[t]\centering
\caption{HumanCLAW sitting diagnostics for eight assigned cases per model. Counts pool independently recorded conditions. $^\dagger$ denotes lower bounds from six Gemini~3.8 records. $^\ddagger$ includes two \mbox{Qwen-Max} episodes ending in provider errors: both record target finding, neither the native navigation criterion nor a sit action. The other six record one target found. Dashes denote unavailable process metrics. Terminal success retains all eight assignments. Opus~5.5 lacks near-plus-sit action counts. \mbox{DeepSeek-1017} and Gemini~4~Argon use the same native success fields. Nav.-plus-sit counts require the native 20\,cm navigation flag and a sit action; they do not establish a seated state. Navigation flags are not inferred from distance alone.}\label{tab:humanclaw-stages}
\ArenaTableFont\setlength{\tabcolsep}{3pt}
\begin{tabular*}{\linewidth}{@{\extracolsep{\fill}}lrrrrr@{}}\toprule
Model & Process coverage & Target found & 20\,cm nav. & Nav. + sit issued & Success\\\midrule
Astra & 8/8 & 6 & 5 & 5 & 0\\
Sol & 8/8 & 3 & 1 & 1 & 0\\
Fable & 8/8 & 0 & 0 & 0 & 0\\
\mbox{Qwen-Max} & 8/8$^\ddagger$ & 3$^\ddagger$ & 0 & 0 & 0\\
Gemini~3.8 & 6/8 & 3$^\dagger$ & 3$^\dagger$ & 3$^\dagger$ & 0\\
Q-397B & 8/8 & 4 & 1 & 1 & 0\\
Q-27B & 8/8 & 4 & 2 & 2 & 0\\
Opus~5.5 & 8/8 & 4 & 4 & -- & 0\\
\mbox{DeepSeek-1017} & 8/8 & 5 & 3 & 3 & 0\\
Gemini~4~Argon & 8/8 & 7 & 7 & 7 & 0\\
\bottomrule\end{tabular*}\end{table}

\noindent\textbf{Attempted actions and terminal goals.}
All eleven Astra household failures exhaust their assigned budget. ScienceWorld's nine failures have scored task-failure outcomes with no recorded parser-no-match, invalid-action, or code-exception events; negative score sentinels are not physical-progress measures. HumanCLAW identifies a different limit (Table~\ref{tab:humanclaw-stages}): six Astra sitting episodes find the target, five record both proximity within 20\,cm and a sit action, yet none succeeds. Gemini~4~Argon finds the target in 7/8 cases and combines 20\,cm proximity with a sit action in 7/8, yet completes none; \mbox{DeepSeek-1017} records the same conjunction in 3/8, also without success. Fable records no target-finding success on these episodes. Both \mbox{Qwen-397B} and \mbox{Qwen-27B} find the target in 4/8 cases, but record proximity plus a sit action in only 1/8 and 2/8, respectively. The same zero sitting-success rate thus accompanies different recorded target-finding and approach outcomes.

\subsection{Manipulation: Subgoals, Control Events, and Task Success}
\label{app:manipulation-focus}
Table~\ref{tab:w4-protocol-success} breaks down Manipulation results to all thirteen sources and ten agents. Astra records 76 successes out of 182 assignments; Fable records 45, with nine missing ManiSkill records receiving no success credit. Table~\ref{tab:w4-protocol-stages} includes auxiliary metrics for all ten agents wherever matching stage evidence exists. Each task and time window retains the definition below.

\input{sections/table_w4_protocol_success}
\input{sections/table_w4_protocol_stages}
\input{sections/calvin_recovery}

\input{sections/appendix_w4_metrics}

%% file: sections/supplement_figure6_affordance.tex
\begin{figure}[!htbp]
\centering
\includegraphics[width=\linewidth]{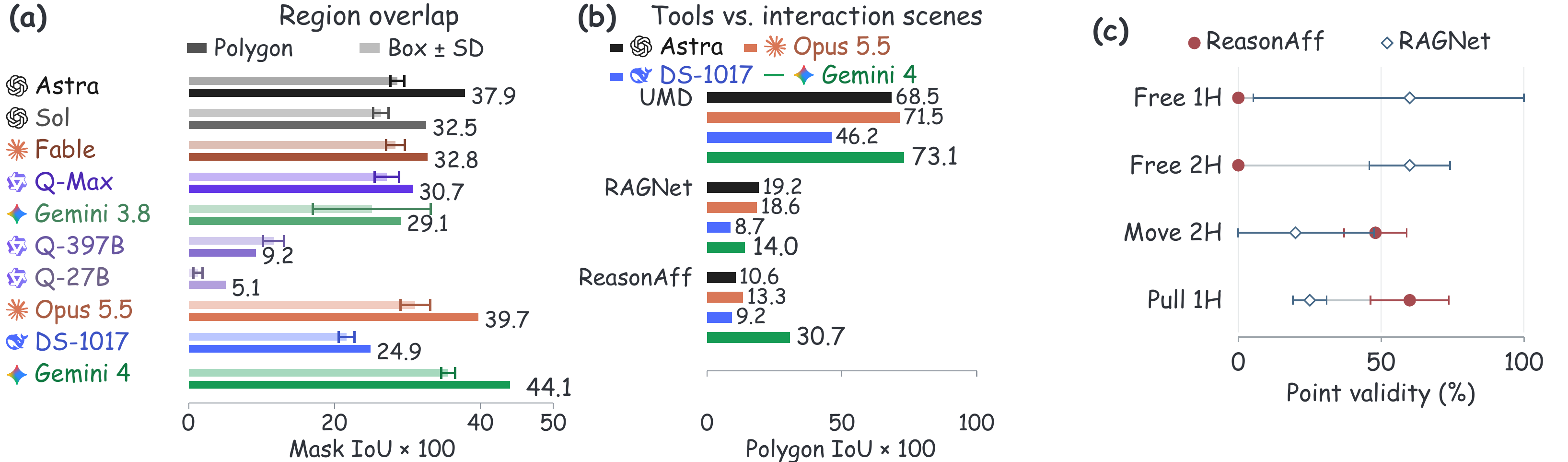}
\caption{Flexible outlines help, but interaction scenes remain difficult. (a) Aggregate mask IoU for native boxes and direct polygons across all ten agents. Box whiskers show sample SD over five runs (three for Fable); polygon bars use one separate run on the same 70 cases. (b) Polygon IoU on tools (UMD) and interaction-conditioned scenes (RAGNet and ReasonAff) for Astra, Opus~5.5, \mbox{DeepSeek-1017}, and Gemini~4~Argon. (c) Astra's original contact validity by action, with run SD; 1H/2H denotes one/two hands. Invalid or absent polygon outputs score zero. Appendix~\ref{app:polygon} defines the polygon protocol.}
\label{fig:affordance-detail}
\end{figure}

%% file: sections/supplement_figure7_planning.tex
\begin{figure}[!htbp]
\centering
\setlength{\abovecaptionskip}{5pt}
\setlength{\belowcaptionskip}{0pt}
\includegraphics[width=\linewidth]{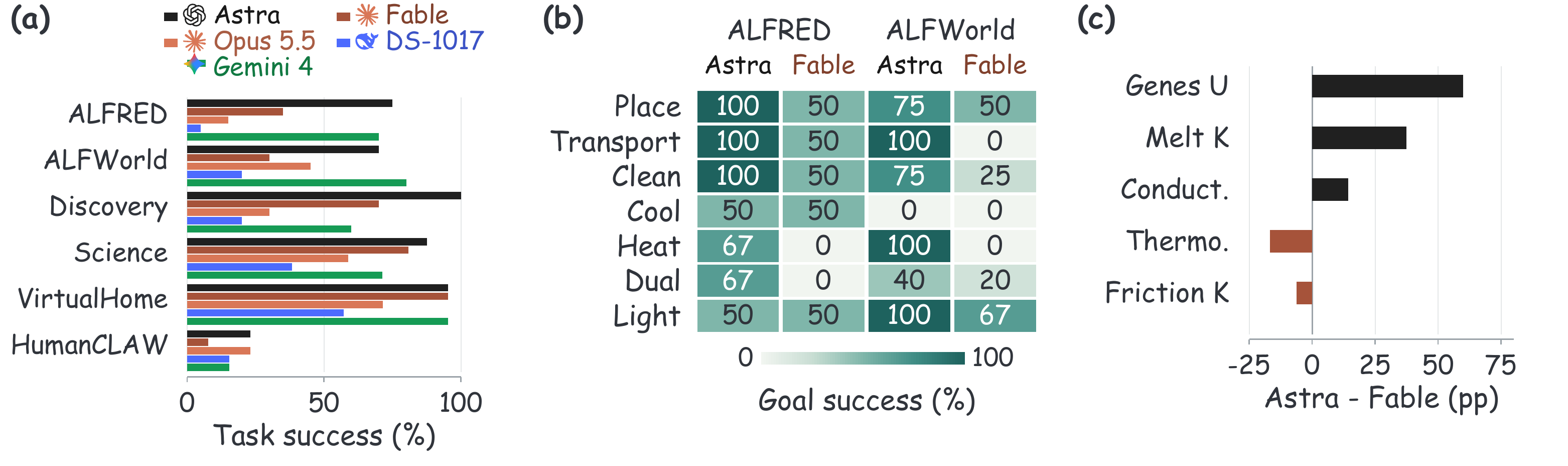}
\caption{Planning success varies across sources and goals. (a) Source-level success for Astra, Fable, Opus~5.5, \mbox{DeepSeek-1017}, and Gemini~4~Argon. \mbox{DeepSeek-1017} and Gemini~4~Argon each complete 2/13 HumanCLAW tasks, both navigation. Opus~5.5 completes 3/13 HumanCLAW tasks, all navigation; none of its eight sitting tasks succeeds. (b) Astra/Fable success on seven household goal families. (c) Their five nonzero ScienceWorld gaps; thirteen families tie, and K/U denotes known/unknown conditions. Each assignment uses one run. Appendix~\ref{app:planning-focus} gives the task-family breakdown; Table~\ref{tab:opus-sources} \mbox{gives Opus~5.5 coverage.}}
\label{fig:planning-detail}
\end{figure}

%% file: sections/table_w4_protocol_success.tex
\begin{table}[!htbp]\centering
\caption{Adopted Manipulation successes across thirteen sources and ten agents. Headers give assigned tasks; missing Fable ManiSkill records (--) receive no success credit.}\label{tab:w4-protocol-success}
\ArenaTableFont\setlength{\tabcolsep}{3pt}
\begin{tabular*}{\linewidth}{@{\extracolsep{\fill}}lrrrrrrr@{}}\toprule
Model & CALVIN & CLIPort & RLBench & RoboCasa & RC365 & RoboTwin & RoboWits\\
Assigned & 10 & 18 & 10 & 10 & 10 & 50 & 10\\\midrule
Astra & 3 & 11 & 6 & 1 & 0 & 19 & 1\\
Sol & 0 & 5 & 2 & 0 & 0 & 5 & 1\\
Fable & 0 & 6 & 1 & 0 & 1 & 7 & 1\\
Q-Max & 0 & 4 & 0 & 0 & 0 & 2 & 1\\
Gemini~3.8 & 0 & 5 & 2 & 0 & 0 & 4 & 2\\
Q-397B & 0 & 4 & 0 & 0 & 0 & 3 & 0\\
Q-27B & 0 & 3 & 1 & 0 & 0 & 4 & 1\\
Opus~5.5 & 1 & 11 & 3 & 0 & 1 & 23 & 1\\
DS-1017 & 1 & 6 & 2 & 0 & 0 & 2 & 2\\
Gemini~4 & 4 & 7 & 8 & 2 & 1 & 14 & 1\\
\bottomrule\end{tabular*}
\par\medskip
\begin{tabular*}{\linewidth}{@{\extracolsep{\fill}}lrrrrrr@{}}\toprule
Model & VIMA & VLABench & ManiSkill & LIBERO & robosuite & BEHAVIOR\\
Assigned & 10 & 25 & 9 & 10 & 5 & 5\\\midrule
Astra & 4 & 18 & 5 & 4 & 4 & 0\\
Sol & 3 & 13 & 0 & 3 & 4 & 0\\
Fable & 3 & 15 & -- & 6 & 5 & 0\\
Q-Max & 1 & 5 & 0 & 1 & 3 & 0\\
Gemini~3.8 & 2 & 9 & 1 & 4 & 5 & 0\\
Q-397B & 3 & 6 & 0 & 0 & 2 & 0\\
Q-27B & 3 & 6 & 0 & 1 & 1 & 0\\
Opus~5.5 & 10 & 19 & 5 & 8 & 5 & 0\\
DS-1017 & 6 & 9 & 1 & 5 & 4 & 0\\
Gemini~4 & 9 & 16 & 5 & 8 & 5 & 0\\
\bottomrule\end{tabular*}
\end{table}

%% file: sections/table_w4_protocol_stages.tex
\begin{table}[!htbp]\centering
\caption{Auxiliary Manipulation metrics for all ten agents under their recorded settings. CALVIN reports consecutive goals (0--5); other columns report task-normalized stage evidence (\%). Subset sizes are in parentheses. Definitions and evidence windows follow in Table~\ref{tab:w4-metric-scope}. -- denotes unavailable comparable evidence, not zero task success.}\label{tab:w4-protocol-stages}
\ArenaTableFont\setlength{\tabcolsep}{2pt}
\begin{tabular*}{\linewidth}{@{\extracolsep{\fill}}l*{7}{r}@{}}\toprule
Model & \shortstack{CALVIN\\(10)} & \shortstack{RoboCasa\\(4)} & \shortstack{Casa365\\(7)} & \shortstack{ManiSkill\\(5)} & \shortstack{LIBERO\\(10)} & \shortstack{robosuite\\(5)} & \shortstack{BEHAVIOR\\(5)}\\\midrule
Astra & 3.10 & -- & -- & 60.00 & 20.00 & 60.00 & 13.33\\
Sol & 1.40 & -- & -- & 0.00 & 20.00 & 60.00 & 16.67\\
Fable & 0.90 & -- & -- & -- & 30.00 & 80.00 & 30.00\\
Q-Max & 0.80 & -- & -- & 0.00 & 5.00 & 40.00 & 16.67\\
Gemini~3.8 & 1.80 & -- & -- & 0.00 & 20.00 & 80.00 & 6.67\\
Q-397B & 0.00 & -- & -- & 0.00 & 0.00 & 30.00 & 6.67\\
Q-27B & 0.10 & -- & -- & 0.00 & 5.00 & 16.67 & 6.67\\
Opus~5.5 & 2.40 & 83.33 & 61.90 & 60.00 & 45.00 & 80.00 & --\\
DS-1017 & 1.40 & 50.00 & 28.57 & 0.00 & 25.00 & 60.00 & 23.33\\
Gemini~4 & 3.30 & 83.33 & 57.14 & 60.00 & 40.00 & 80.00 & 23.33\\
\bottomrule\end{tabular*}
\end{table}

%% file: sections/calvin_recovery.tex
\subsubsection{CALVIN Subgoal Completion and Execution Errors}
\label{app:calvin-recovery}
The analysis uses the same ten sequence IDs for all ten models, giving 100 single-run episodes. Figure~\ref{fig:astra-manipulation}a counts episodes with at least $k$ verified consecutive goals, for $k=1,\ldots,5$.

\noindent\textbf{Attrition along the sequence.}
Summing the ten model rows gives 60, 40, 25, 18, and nine episodes reaching at least one through five goals. Thus nine of the 60 episodes with a verified first goal finish the sequence, and nine of the 18 reaching four goals finish the fifth. These are nested counts on the recorded sequences, not independent per-step probabilities. Unknown prefixes receive no verified credit as specified below.

\noindent\textbf{Resumed execution and subsequent task success.}
An execution-error episode contains a round with a code-execution failure, including syntax, interface-contract, or runtime/tool errors. API transport failures, timeouts, and budget guards are excluded. Resumed execution requires a later round marked \texttt{execution\_ok=true}; subsequent task success requires a later successful native task verification. An episode contributes once to each count, regardless of the number of errors. Thus, the three episodes with subsequent task success are a subset of the 62 resumed episodes and the 65 error-containing episodes (Table~\ref{tab:calvin-recovery}). These counts describe the recorded settings in Appendix~\ref{app:w4-comparison}.

\begin{table}[!htbp]\centering
\caption{CALVIN: subgoal completion and outcomes after execution errors on ten assigned sequences per model. Error, Resumed, and Then complete are nested episode counts. Full success includes episodes with and without an error. Mean prefix retains the ten-task denominator.}\label{tab:calvin-recovery}
\ArenaTableFont\setlength{\tabcolsep}{4pt}
\begin{tabular*}{\linewidth}{@{\extracolsep{\fill}}lrrrrr@{}}\toprule
Model & Mean prefix & Error & Resumed & Then complete & Full success\\\midrule
Astra & 3.1 & 3 & 3 & 1 & 3\\
Sol & 1.4 & 3 & 3 & 0 & 0\\
Fable & 0.9 & 9 & 9 & 0 & 0\\
Q-Max & 0.8 & 10 & 9 & 0 & 0\\
Gemini~3.8 & 1.8 & 8 & 8 & 0 & 0\\
Q-397B & 0.0 & 2 & 2 & 0 & 0\\
Q-27B & 0.1 & 7 & 6 & 0 & 0\\
Opus~5.5 & 2.4 & 9 & 9 & 1 & 1\\
DS-1017 & 1.4 & 10 & 10 & 1 & 1\\
Gemini~4 & 3.3 & 4 & 3 & 0 & 4\\
\midrule
All episodes & -- & 65 & 62 & 3 & 9\\
\bottomrule\end{tabular*}\end{table}

\noindent\textbf{Incomplete prefix evidence.}
One Qwen-Max episode has no verified prefix; five Qwen-397B episodes stop without a reported prefix. These six assignments receive zero verified goal credit within the ten-task denominator. The other 94 prefixes are explicit. Complete-sequence success requires the native verifier for every model. The released analysis data provide per-episode error rounds, prefix values, and aggregate counts.

%% file: sections/appendix_w4_metrics.tex
\noindent\textbf{Metric definitions and coverage.}
\label{app:w4-metrics}
The 183-case Manipulation inventory contains 182 binary tasks and one continuous-return AnymalC-Spin task. AnymalC-Spin is evaluated separately through continuous return. For each model and configuration, native acceptance gives $s_i=1$ and task success is
\begin{equation*}
\mathrm{SR}=\frac{100}{N}\sum_{i=1}^{N}s_i.
\end{equation*} Runs without native acceptance receive zero success credit, including timeouts, execution or infrastructure failures, and missing records. Each task contributes one run to the assigned denominator.

CALVIN scores a five-goal sequence as successful only when all five goals are completed in order. Let $L_i\in\{0,\ldots,5\}$ be the verified consecutive prefix. The average number of consecutive goals completed is
\begin{equation*}
\bar L=\frac{1}{10}\sum_{i=1}^{10}L_i.
\end{equation*} Unverified prefixes receive no goal credit; Appendix~\ref{app:calvin-recovery} gives the case-level coverage.

For the other auxiliary metrics, case $i$ has $K_i$ scoring items and verified indicators $z_{ij}$. The per-case score and reported mean are
\begin{equation*}
u_i=\frac{1}{K_i}\sum_{j=1}^{K_i}z_{ij},\qquad
\bar u=\frac{100}{N}\sum_{i=1}^{N}u_i.
\end{equation*} Failed or unknown items contribute zero. This weights tasks equally even when $K_i$ differs. The diagnostics summarize recorded control events on fixed task subsets using the scoring items and evidence windows in Table~\ref{tab:w4-metric-scope}.

\begin{table}[!htbp]\centering
\caption{Six auxiliary metrics, averaging task-normalized evidence within each subset.}
\label{tab:w4-metric-scope}
\ArenaTableFont\setlength{\tabcolsep}{4pt}
\begin{tabularx}{\linewidth}{@{}lrYY@{}}
\toprule Source & $N$ & Scoring items & Evidence window\\\midrule
RoboCasa & 4 & Approach, grasp and lift, stable release & Any verified event during execution\\
RoboCasa365 & 7 & Same three transport milestones & Any verified event during execution\\
ManiSkill & 5 & Two or three native conditions per task & Final verification state\\
LIBERO-PRO & 10 & Two task-specific conditions & Final state and limited process records\\
robosuite & 5 & Lifting: 2; stacking: 3; wiping: 4 & Native success and reward-derived evidence\\
BEHAVIOR-1K & 5 & Six control-evidence items & Before the first episode-end signal\\
\bottomrule\end{tabularx}
\end{table}

\noindent\textbf{Transport control versus destination success.}
RoboCasa uses ArrangeBreadBasket, LoadDishwasher, PickPlaceToasterToCounter, and RecycleBottlesByType. RoboCasa365 uses BreadAndCheese, MakeBananaMilkshake, MicrowaveThawingFridge, MixCakeFrosting, PlaceMeatInMarinade, PlaceVegetablesEvenly, and RemoveCuttingBoardItems. Door, appliance, and navigation-only tasks are outside these subsets. Approach requires executed motion and confirmed arrival; grasp-and-lift requires both grasp and lift confirmation. Stable release requires prior grasp-and-lift of the same canonically named object, followed by confirmed release and stability. Each item scores at most once per case. These events need not cover every task object, and release does not establish the requested destination relation.

\begin{table}[!htbp]\centering
\caption{ManiSkill's five-task diagnostic. Items are evaluated at final verification; grasp history is separate. The other four binary tasks remain in complete-task success.}
\label{tab:w4-maniskill-definition}
\ArenaTableFont\setlength{\tabcolsep}{4pt}
\begin{tabularx}{\linewidth}{@{}lrY@{}}
\toprule Task & $K_i$ & Scored conditions\\\midrule
PickCube & 2 & Object placed at its goal; placed and robot static. The suspended goal requires no release.\\
PlaceSphere & 3 & Sphere on its support; on support and released; both with the sphere static.\\
StackCube & 3 & Cube A on B; on B and released; both with cube A static.\\
PlugCharger & 2 & Goal distance $\leq0.005$\,m; distance satisfied and angular error $\leq0.2$\,rad.\\
PegInsertionSide & 2 & Peg-head $x\geq-0.015$\,m in the hole frame; official insertion success.\\
\bottomrule\end{tabularx}
\end{table}

LiftPegUpright, PullCube, PushCube, and RollBall contribute to the nine-task success rate but not this diagnostic. Dashes mark unavailable Fable baseline records in both sets.

\noindent\textbf{Task-specific conditions in LIBERO-PRO and robosuite.}
LIBERO-PRO uses fixed init-0 instances: spatial 1/7, object 2/5, goal 1/7, libero\_10 1/9, spatial\_swap 0, and object\_swap 4. Ordinary placement scores recorded grasp and the official target relation; stove opening scores operation of the correct control and an open stove. The two longer tasks score a cup inside a microwave followed by the door being closed, or one food item followed by both items in a basket. Missing grasp/control records score zero; for the longer tasks, necessary preceding relations are confirmed only on official success.

robosuite uses cube lifting with seeds 99/50, cube stacking and restacking with seed 14, and spill wiping with seed 48. Lifting scores recorded grasp and native success; grasp history is missing. Stacking uses the fixed local reward $q$: $q>0.125$ supports a grasp/lift/stack condition, $q\geq0.5$ supports the top cube being lifted, and native success supplies the third item. A restacking zero can mask intermediate events under the two-cube airborne rule; these items remain unknown and receive zero credit. Wiping checks 25\%, 50\%, 75\%, and complete removal. The first three thresholds receive credit only when native success confirms complete wiping; total reward does not measure the fraction wiped. Consequently, Fable's 100\% complete-task success and 80\% auxiliary score are compatible: missing intermediate evidence lowers the latter.

\begin{table}[!htbp]\centering
\caption{BEHAVIOR-1K control evidence. Each item records one type of control event and scores once per case; terminal BDDL goal success is evaluated separately.}
\label{tab:w4-behavior-definition}
\ArenaTableFont\setlength{\tabcolsep}{4pt}
\begin{tabularx}{\linewidth}{@{}lY@{}}
\toprule Item & Positive evidence\\\midrule
Localization & Finite coordinates or pose from public RGB-D localization or grasp sampling.\\
Candidate & A nonempty pregrasp or grasp candidate set.\\
Arrival & Measured arrival from approach or recovery feedback for the same arm.\\
Holding & An explicit in-hand check or holding confirmation inside checked placement.\\
Held motion & Associated measured motion of at least 5\,cm while holding, or movement with a passed holding check inside checked placement.\\
Release & Holding, opening, then no longer holding; or release verified in checked placement.\\
\bottomrule\end{tabularx}
\end{table}

The five tasks are outfit\_a\_basic\_toolbox, can\_meat, tidying\_bedroom, putting\_shoes\_on\_rack, and chop\_an\_onion. Records are linked by arm, name, and temporal order; physical-instance IDs are not consistently recorded. Object identity and inside/ontop relations are checked separately by the terminal goal evaluator. Held motion uses a 5\,cm threshold. In the original seven-agent cohort, control-event records cover four tasks for each of four models and the toolbox task for each of the other three. All retain the five-task denominator. Scoring uses the evaluated episodes up to their first end signal and requires direct evidence for each item. Fable's nine verified items yield
\begin{equation*}
100\times\frac{9}{5\times6}=30\%,
\end{equation*}
versus Astra's four items and 13.33\%. RoboWits reports task success.

\noindent\textbf{Later-model stage coverage.}
Opus~5.5, \mbox{DeepSeek-1017}, and Gemini~4~Argon provide matching task-stage records for the RoboCasa, RoboCasa365, ManiSkill, LIBERO-PRO, and robosuite subsets. Their scores retain the same task-normalized definition; complete-task success remains separate. Gemini~4~Argon records 83.33\% transport evidence on the four RoboCasa tasks but completes none of these four tasks: approaching, lifting, and releasing an object does not ensure that every requested destination condition holds. Its two RoboCasa successes belong to other task types. For BEHAVIOR-1K, the DeepSeek and Gemini~4 logs support the six pre-end-signal control items; Opus supplies different diagnostics, so this column is unavailable.

%% file: sections/appendix_setting_comparisons.tex
\setlength{\parskip}{0.2\baselineskip}
\section{Execution Settings and Multi-Round Review}
\label{app:setting-comparisons}
\input{sections/table_review}
\subsection{Recorded Manipulation Settings}
\label{app:w4-comparison}
The main evaluation contains 182 binary assignments per model, with one episode per assignment. Nine missing Fable ManiSkill records receive no success credit in the domain total and remain unavailable in source-level comparisons. Table~\ref{tab:w4-protocol-success} lists all thirteen sources for all ten agents.

\input{sections/appendix_w4_settings}

\subsection{Fixed-Input Multi-Round Review}
\label{app:multi-round-review}
Multi-round review lets agents revisit their initial answers using the same task observations and scoring criteria. Self-review adds reconsideration, Crop review adds image crops, and Tool-assisted review adds specialist visual and geometric measurements. The three conditions permit up to two, two, and six additional responses, respectively, across Geometry, Spatial Reasoning, and Affordance. Table~\ref{tab:review} reports results; sampling and tools are specified below.

\noindent\textbf{Shared cases.}
Astra and \mbox{Qwen-27B} share 46 cases: 19 Geometry, 18 Spatial Reasoning, and nine Affordance. A 72-case pilot covers 24 benchmark/task strata selected by task category and SHA-256 ordering with seed 20260923, independently of scores and answers. It excludes the 60 exact cases from an earlier performance-selected study. One case per stratum forms a 24-case development set; withholding two more cases that share development scenes leaves the 46 evaluation cases. Some scenes recur, and some cases were inspected during development.

\begin{table}[!ht]
\centering
\caption{Review conditions for the same cases and native scoring interfaces. Budgets are additional to native execution; response limits count output tokens.}
\label{tab:review-conditions}
\ArenaTableFont
\begin{tabularx}{\linewidth}{@{}lYY@{}}
\toprule
Condition & Evidence available during review & Additional response budget\\
\midrule
Initial & Original task observations & None\\
Self-review & Original observations and initial answer & Two responses, up to 1,800 tokens each\\
Crop review & Self-review evidence plus image crops & Two responses, up to 1,800 tokens each\\
Tool-assisted & RGB, crops, specialist measurements & Six responses, each up to 4,096 tokens; 16 requests\\
\bottomrule
\end{tabularx}
\end{table}

\noindent\textbf{Evidence construction.}
Tool-assisted review adapts GCA's measurement and reference-frame formalization~\citep{gca} and SpatialClaw's inspectable execution~\citep{spatialclaw}. It formulates the measurement before consulting the candidate answer, then uses a persistent Python session. SAM~3~\citep{sam3} supplies regions; SuperPoint~\citep{superpoint} and LightGlue~\citep{lightglue} supply correspondences; GeoCalib~\citep{geocalib} estimates calibration. Numerical routines compute relations and estimates. Finite-value and coordinate checks validate the computed outputs. Remaining semantic errors include selecting a whole-object mask for a functional part and ambiguous translation scale.

\noindent\textbf{Observation and execution controls.}
Specialist tools receive only previously exposed RGB, without predicted depth, learned 3D reconstruction, private calibration, scoring annotations, or extra video frames. Public depth or masks originally supplied by a task remain available to the base model. Review reuses native candidates where available; one Astra Spatial case starts from its public observation because the candidate is missing. Temperature is zero, and total-token and wall-time accounting includes native execution. Each condition contributes one final outcome per case, selected independently of correctness; any error-terminated outcome receives no success credit.

\noindent\textbf{Outcome interpretation.}
Geometry counts improved, tied, and worsened results across 24 metric entries from 19 cases; one \mbox{Qwen-27B} crop entry is unavailable. Spatial accuracy retains 18 assigned cases, with missing answers counted as incorrect. Affordance reports box--box IoU and success at IoU$\geq0.5$ over nine cases.

\subsection{Native Capability and Complete Task Execution}
\label{app:capability-attribution}
Native capability concerns the decisions a model makes through an execution interface. Comparing robot-agent studies therefore requires identifying the task requirements, the perception and control capabilities supplied by the interface, and the resources available for completion. Together, these factors determine which parts of task completion depend on the model's decisions.

\noindent\textbf{Context and supplied control capabilities.}
Robotic in-context learning uses demonstrations and interaction through context-conditioned policies, geometric transfer, world models, or skill- and agent-based control~\citep{huang2026roboticl}. Fixed model weights can therefore coexist with substantial execution support. Learned skill selection~\citep{saycan}, programmatic composition of perception and control APIs~\citep{codepolicies}, and spatial value maps~\citep{voxposer} assign different responsibilities to the model and executor. GPT-Policy~\citep{cheng2026gptpolicy} conditions a fixed VLM on demonstrations, action references, and feedback, while a constrained controller checks and executes its requests. Our baseline leaves task interpretation and action selection to the model through source-native helpers; the supplementary review study separately introduces specialist measurements.

\noindent\textbf{Task composition and execution support change the comparison.}
Our RoboCasa sample contains seven atomic and three composite tasks; RoboCasa365 contains three atomic and seven composite tasks (Appendix~\ref{app:inventory}). PyRUA-Lean~\citep{si2026pyrualean} evaluates 18 atomic and 32 composite RoboCasa365 task types at five seeds each, with 86.7\% and 39.4\% success. Its Target50 selection differs from our pretrain-designated instances. Removing its frozen RLDX-1 policy lowers atomic and composite success to 65.6\% and 20.0\%, while retaining programmatic composition of other primitives. Execution allowances also differ: our RoboCasa protocols use 24 model rounds and 60 minutes of trial time; PyRUA-Lean uses 40 calls for atomic tasks, 100 for composite tasks, and a two-hour watchdog. Both allow multiple actions per Python response. These settings evaluate different combinations of model reasoning, supplied control, and task demands.

\noindent\textbf{Interaction count and task success measure different outcomes.}
GPT-Policy~\citep{cheng2026gptpolicy} reports 0/3 towel-pickup successes without demonstrations, averaging 96.3 decisions and 24.6 minutes. Human video raises success to 2/3 and lowers those averages to 76.7 decisions and 18.9 minutes, including failed trials. Our CALVIN analysis distinguishes a further requirement: 62 error-containing episodes resume code execution, but only three complete the sequence (Appendix~\ref{app:calvin-recovery}). Resumed code execution and subsequent task success must therefore be measured separately.

\noindent\textbf{The required contact state must hold at verification.}
GPT-Policy's plug task requires the plug to remain fully seated after release. Robot video yields 0/3 successes; adding aligned action references yields 2/3~\citep{cheng2026gptpolicy}. Our within-task diagnostics expose related final-state requirements: initiation or arrival fails in 36 of Astra's 37 unsuccessful traces, and five Astra HumanCLAW episodes record both proximity to the seat and a sit action without satisfying the sitting condition. These examples distinguish a predicted path or attempted action from satisfaction of its required endpoint or contact condition. The arena evaluates these requirements alongside final task success, interaction count, and elapsed time.

%% file: sections/table_review.tex
\begin{table}[!htbp]
\centering
\setlength{\abovecaptionskip}{0pt}
\caption{Fixed-input multi-round review on the shared 46-case subset. I/T/W counts improved/tied/worsened Geometry metric entries; SR$_{50}$ requires box IoU$\geq0.5$. Appendix~\ref{app:multi-round-review} specifies the cases and tools.}
\label{tab:review}
\ArenaTableFont
\setlength{\tabcolsep}{3pt}
\renewcommand{\arraystretch}{1.18}

\begin{tabular*}{\linewidth}{@{\extracolsep{\fill}}llcccc@{}}
\toprule
 & & Geometry & Spatial Reasoning & \multicolumn{2}{c}{Affordance}\\
\cmidrule(lr){3-3}\cmidrule(lr){4-4}\cmidrule(l){5-6}
Model & Condition & I/T/W & Acc.\ (\%)$\uparrow$ & Box IoU$\uparrow$ & SR$_{50}$\ (\%)$\uparrow$\\
\midrule
\multirow{4}{*}{Astra}
 & Initial & -- & \textbf{72.2} & 0.2903 & 22.2\\
 & Self-review & 3/20/1 & 66.7 & 0.2914 & 22.2\\
 & Crop review & 4/19/1 & 66.7 & 0.2912 & 22.2\\
 & Tool-assisted & 5/15/4 & \textbf{72.2} & \textbf{0.3174} & \textbf{33.3}\\
\midrule
\multirow{4}{*}{\mbox{Qwen-27B}}
 & Initial & -- & 55.6 & 0.0336 & 0.0\\
 & Self-review & 6/15/3 & 50.0 & \textbf{0.1524} & 0.0\\
 & Crop review & 3/19/1 & 44.4 & 0.0937 & 0.0\\
 & Tool-assisted & 10/6/8 & \textbf{61.1} & 0.1452 & \textbf{11.1}\\
\bottomrule
\end{tabular*}
\end{table}

%% file: sections/appendix_w4_settings.tex
\noindent\textbf{Observations and execution ceilings.}
All agents use native observations and tools. Observation schedules and resource limits vary by evaluation batch. CALVIN's original seven-model reports allow 640k tokens and 60 minutes; Fable allows 16 model rounds and the other six allow 20. The Opus~5.5, \mbox{DeepSeek-1017}, and Gemini~4~Argon reports allow 40 rounds, 1.28M tokens, and 120 minutes, with automatic per-round RGB. A model round may execute several native actions. The \href{https://github.com/embodied-agent-arena/embodied-agent-arena/blob/main/docs/W4_PROTOCOL_AUDIT.md}{released protocol documentation} specifies these configurations.

The CALVIN progress and recovery analysis covers these 100 episodes. Differences in observation frequency and budget are part of the evaluated configurations. Other source-specific limits appear in Table~\ref{tab:baseline-budgets}, Table~\ref{tab:deepseek-budgets}, and Table~\ref{tab:argon-budgets}. Unreported configuration fields are marked as unspecified. Tool calls, native physical steps, model rounds, and elapsed time are separate counters; video frame rate does not specify the model's observation frequency.

%% file: sections/appendix_case_studies.tex
\setlength{\parskip}{0.2\baselineskip}
\setlength{\intextsep}{8pt plus 1pt minus 1pt}
\section{Case Studies}
\label{app:case-studies}
\begin{figure}[!htbp]
\centering
\captionsetup{font={normalsize,sengokuserif}}
\includegraphics[width=\linewidth]{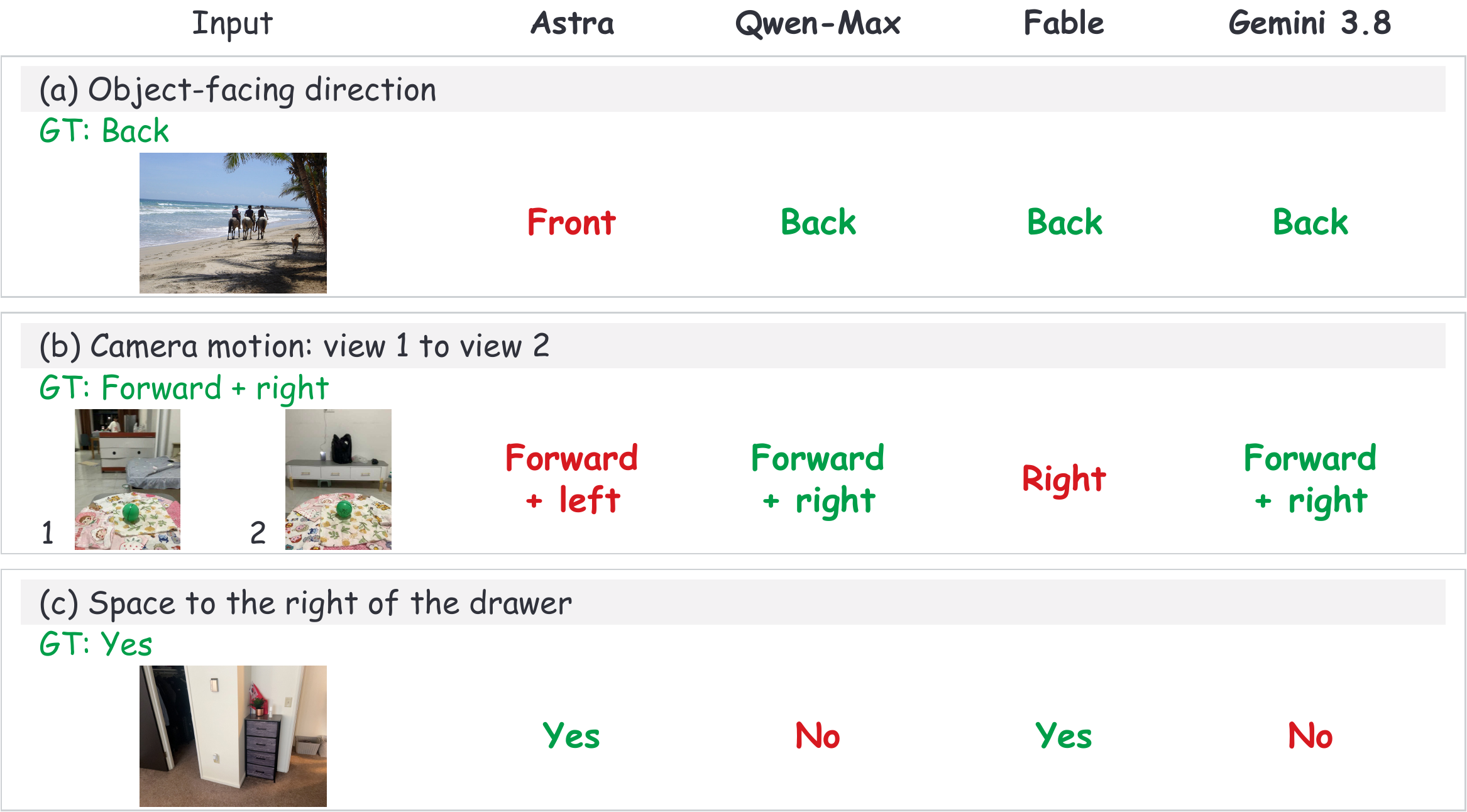}
\caption{Spatial judgments from the first evaluation run: 3DSRBench, MindCube, and RoboSpatial. The questions concern the dog's facing direction, camera motion, and whether the plant fits beside the drawer. Green answers match the reference; red answers are incorrect.}
\label{fig:spatial-cases}
\end{figure}

\begin{figure}[!htbp]
\centering
\includegraphics[width=\linewidth]{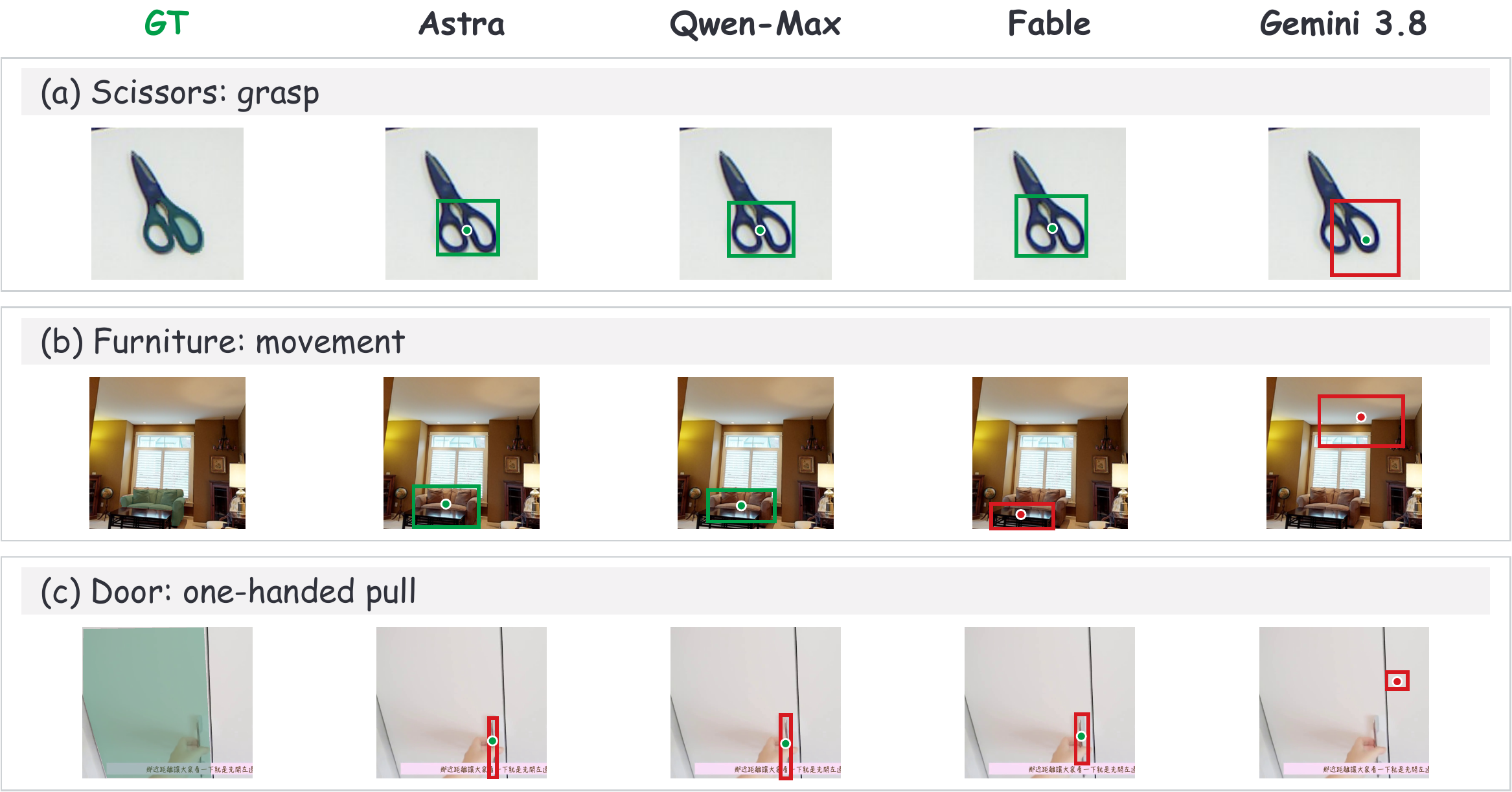}
\caption{Affordance localization in UMD, ReasonAff, and RAGNet. Green points lie inside the reference mask; green boxes have mask IoU$\geq0.5$. Red denotes failure. A point inside the reference mask can accompany a box that covers only a small part of that region, as the handle-versus-door example shows.}
\label{fig:affordance-cases}
\end{figure}

\noindent\textbf{Spatial judgments and contact-point localization.}
Figures~\ref{fig:spatial-cases} and~\ref{fig:affordance-cases} illustrate Section~\ref{sec:astra} with six cases from the first evaluation run.

\inlinepoint{172}{Spatial references.}
In these runs, Astra correctly judges placement compatibility but misidentifies object-facing direction and camera translation.
\inlinepoint{173}{Contact versus coverage.}
Astra localizes annotated tool and furniture regions. For pulling, three models predict points on the handle that lie inside the reference mask, but their boxes cover only a fraction of the reference door region.

\noindent\textbf{Motion and task completion.}
Figure~\ref{fig:execution} contrasts visible motion with verified completion. Figure~\ref{fig:astra-manipulation} complements these examples with the full CALVIN progress and recovery counts.

%% file: sections/appendix_benchmark_resources.tex
\setlength{\parskip}{0.2\baselineskip}
\section{Evaluation Scales and Public Resources}
\label{app:benchmark-resources}
Table~\ref{tab:benchmark-resources} expands Table~\ref{tab:benchmark-comparison} with instance counts and released artifacts. Task categories, training demonstrations, and source datasets are excluded from the instance count. Our release provides a fixed 1,000-case index across five capability domains, together with observations, episode initializations, and evaluation code.

OpenEQA adds navigation to episodic QA in its active track; Memory denotes cross-session persistence. VLABench's non-interactive track scores action sequences from fixed multi-view images against reference graphs. Hard Vision, Easy Vision varies reasoning effort and tool use; its 55 source benchmarks do not specify an instance count.

\begingroup
\ArenaTableFont
\renewcommand{\arraystretch}{1.10}
\setlength{\tabcolsep}{5pt}
\begin{longtable}{@{}>{\raggedright\arraybackslash}p{.25\linewidth}>{\raggedright\arraybackslash}p{.32\linewidth}>{\raggedright\arraybackslash}p{\dimexpr.43\linewidth-20pt\relax}@{}}
\caption{\textbf{Reported evaluation populations and released resources.} Counts follow the cited papers; artifact availability is checked on 3--4 October 2026. Task-type and source counts describe coverage rather than tested instances. Links lead to author-maintained resources.}
\label{tab:benchmark-resources}\\
\toprule
Study & Reported evaluation population & Released evaluation resources\\
\midrule\endfirsthead
\multicolumn{3}{l}{\textit{Evaluation populations and released resources (continued).}}\\
\toprule
Study & Reported evaluation population & Released evaluation resources\\
\midrule\endhead
OpenEQA~\citeyearpar{majumdar2024openeqa} & 1,636 QA entries; the 557 HM3D questions also support active navigation. GPT-4V's active evaluation uses a 184-question subset. & \href{https://github.com/facebookresearch/open-eqa}{Baseline and scoring code}; \href{https://github.com/facebookresearch/open-eqa/blob/main/data/open-eqa-v0.json}{QA manifest} and episode-history download instructions.\\
\addlinespace
BEAR~\citeyearpar{bear} & 4,469 multimodal QA entries; robot online validation is additional. & \href{https://github.com/yqi19/BEAR-official}{Agent code}; \href{https://huggingface.co/datasets/yqi19/BEAR-benchmark/tree/main}{benchmark data and scoring scripts}.\\
\addlinespace
EmbodiedBench~\citeyearpar{embodiedbench} & 1,128 testing instances in four environments. & \href{https://github.com/EmbodiedBench/EmbodiedBench}{Code and configurations}; \href{https://huggingface.co/EmbodiedBench}{benchmark datasets}.\\
\addlinespace
EmbodiedEval~\citeyearpar{embodiedevel} & 328 tasks in 125 scenes. & \href{https://github.com/thunlp/EmbodiedEval}{Code}; \href{https://huggingface.co/datasets/EmbodiedEval/EmbodiedEval}{328-entry test set and scene assets}.\\
\addlinespace
Embodied Arena~\citeyearpar{embodiedarena} & 22 source benchmarks; a combined evaluated-instance count is not reported. & \href{https://github.com/EmbodiedArena/embodied-arena}{Platform and configurations}; source datasets are downloaded separately.\\
\addlinespace
CaP-X~\citeyearpar{capx} & Core: seven task types, each with 100 trials per tier. Extensions include 30 LIBERO-PRO task types and two BEHAVIOR tasks with 25 trials each. & \href{https://github.com/capgym/cap-x}{Code, environments, and evaluation scripts}. The repository lists 39 task types; the paper describes 187 platform tasks.\\
\addlinespace
VLABench~\citeyearpar{vlabench} & 100 task categories: 60 primitive and 40 composite; episode counts depend on the experiment. & \href{https://github.com/OpenMOSS/VLABench}{Code and evaluation tracks}; \href{https://huggingface.co/datasets/VLABench/vlm_evaluation_v1.0}{fixed-image VLM evaluation data}.\\
\addlinespace
WorldLines~\citeyearpar{worldlines} & 310 memory QA entries and 21 planning cases; the 62-entry ablation uses a QA subset. & The \href{https://arxiv.org/html/2606.18847v2}{paper} supplies prompts and examples; an author code/data release was not located.\\
\addlinespace
Hard/Easy Vision~\citeyearpar{rasheed2026frontiervision} & 55 benchmarks, 34 capabilities, nine areas; total and per-benchmark evaluated-instance counts are not reported. & \href{https://mbzuai-oryx.github.io/frontier-vision/}{Results and qualitative examples}; the \href{https://github.com/mbzuai-oryx/frontier-vision}{repository} contains a README and overview image, without evaluation scripts or a test manifest.\\
\midrule
\rowcolor[HTML]{EEF5F8}
\textbf{Embodied Agent Arena} & \textbf{1,000 cases} from 32 established sources and GeoProbe, including 168 GeoProbe cases. & \href{https://github.com/embodied-agent-arena/embodied-agent-arena}{Harness and evaluation code}; \href{https://huggingface.co/datasets/uuu-Quant/Embodied-Agent-Arena}{case index, observations, and episode initializations}.\\
\bottomrule
\end{longtable}
\endgroup

%% file: sections/appendix_latest_cases.tex
\section{Additional Cases Across the Five Capabilities}
\label{app:latest-cases}

\begin{figure}[!htbp]
\centering
\includegraphics[width=\linewidth]{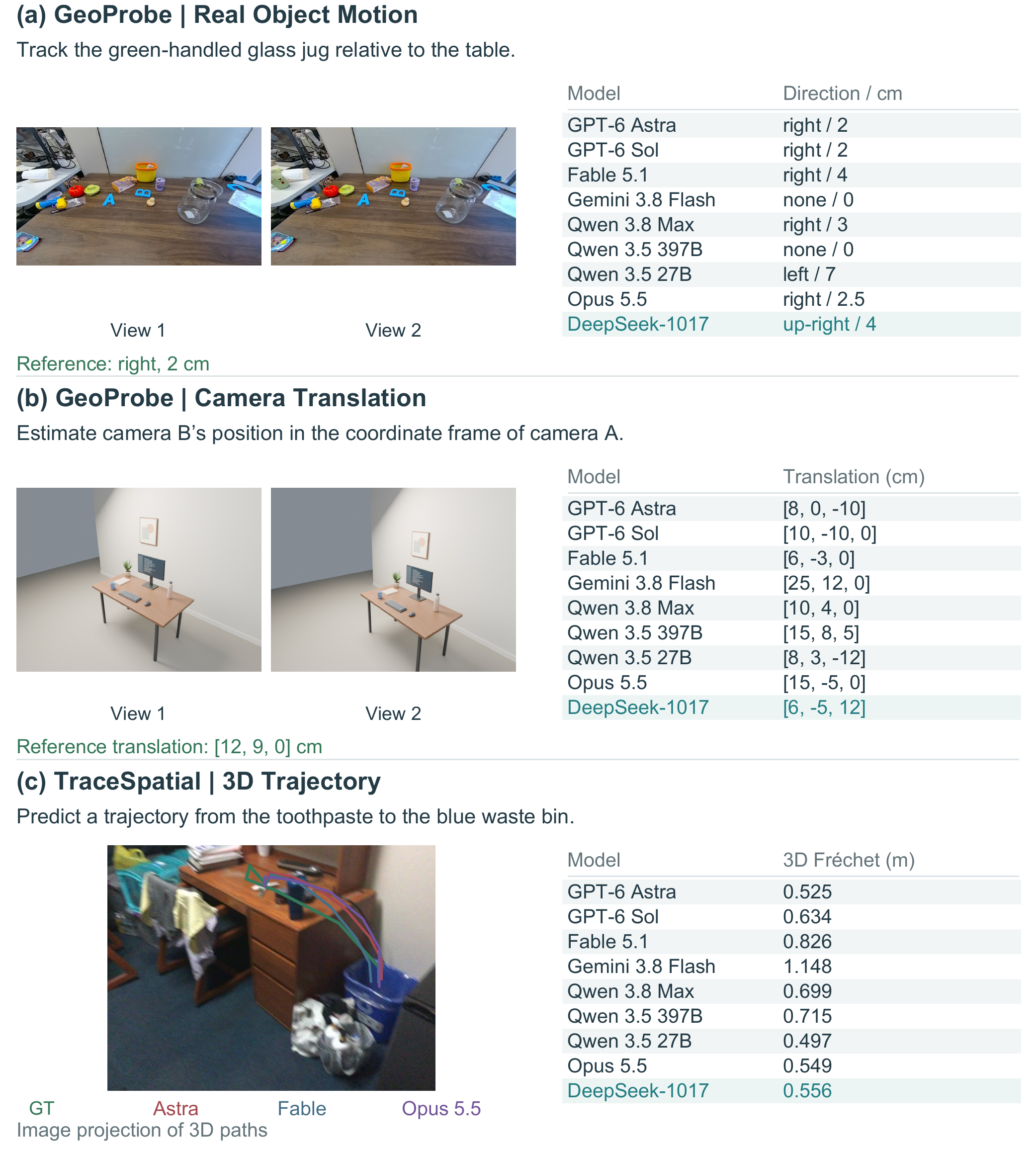}
\caption{\textbf{Geometry: displacement, reference frames, and trajectories.} Selected first-run predictions. (a) The jug moves right by 2\,cm relative to the table. (b) Camera translation uses camera A's frame, with reference $[12,9,0]$\,cm; orientation is omitted. (c) TraceSpatial paths are projected into the image; the errors are 3D Fr\'echet distances in meters, distinct from aggregate trajectory RMSE. Reference, Astra, Fable, and Opus~5.5 paths are drawn, with all nine errors listed.}
\label{fig:latest-geometry}
\end{figure}

\begin{figure}[!htbp]
\centering
\includegraphics[width=\linewidth]{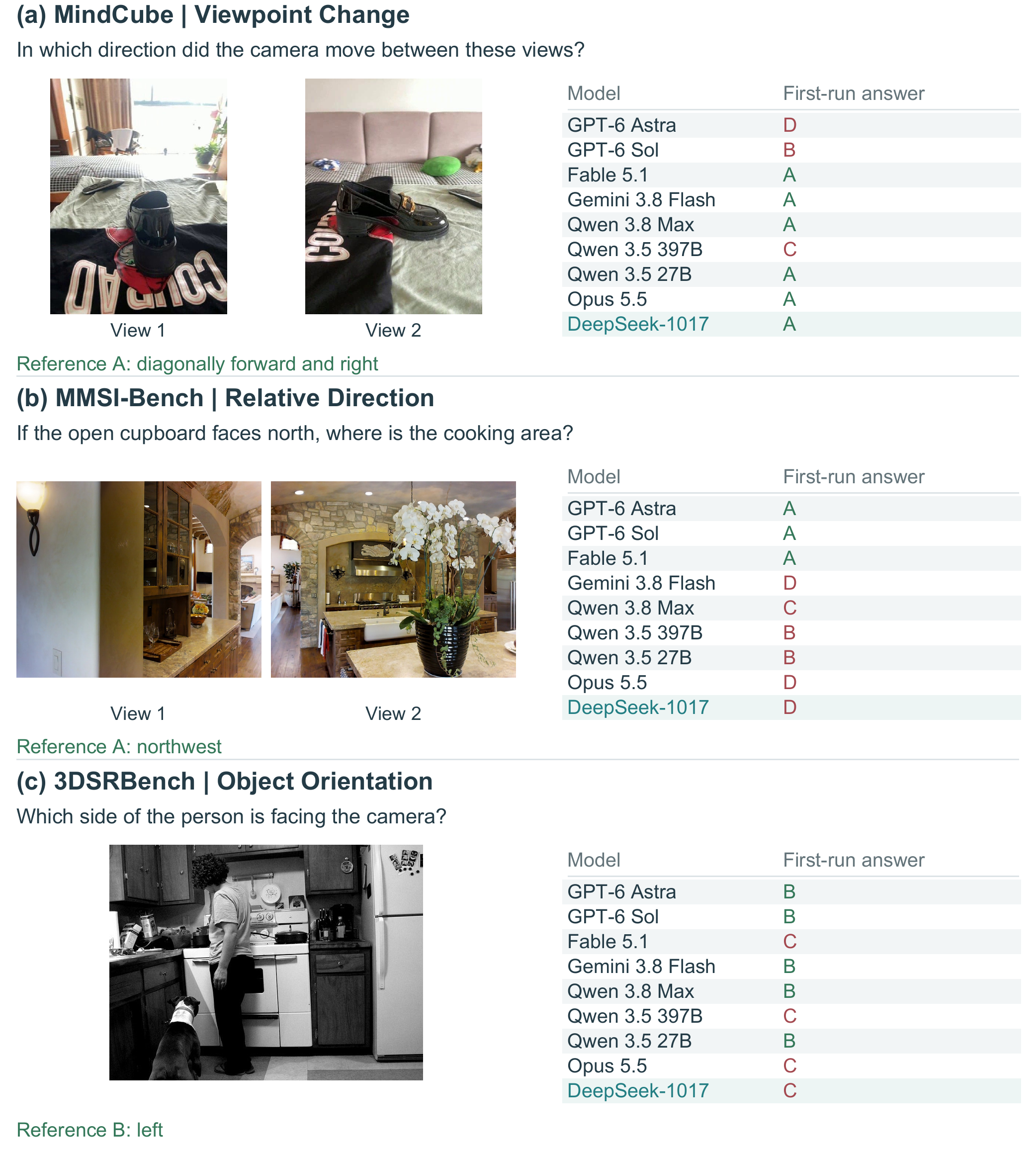}
\caption{\textbf{Spatial Reasoning: errors depend on the reference frame.} Selected first-run answers without YOLOE. (a) MindCube asks about camera movement between two views: A is diagonally forward-right, B right, C left, and D diagonally forward-left. (b) MMSI-Bench displays two selected views from its multi-view input and asks for a direction anchored to the open cupboard: A is northwest, B southwest, C north, and D northeast. (c) 3DSRBench asks which side of the person faces the camera: A is front, B left, C back, and D right. Green letters match the source reference; red letters do not. Different agents succeed on different frame transformations.}
\label{fig:latest-spatial}
\end{figure}

\begin{figure}[!htbp]
\centering
\includegraphics[width=\linewidth]{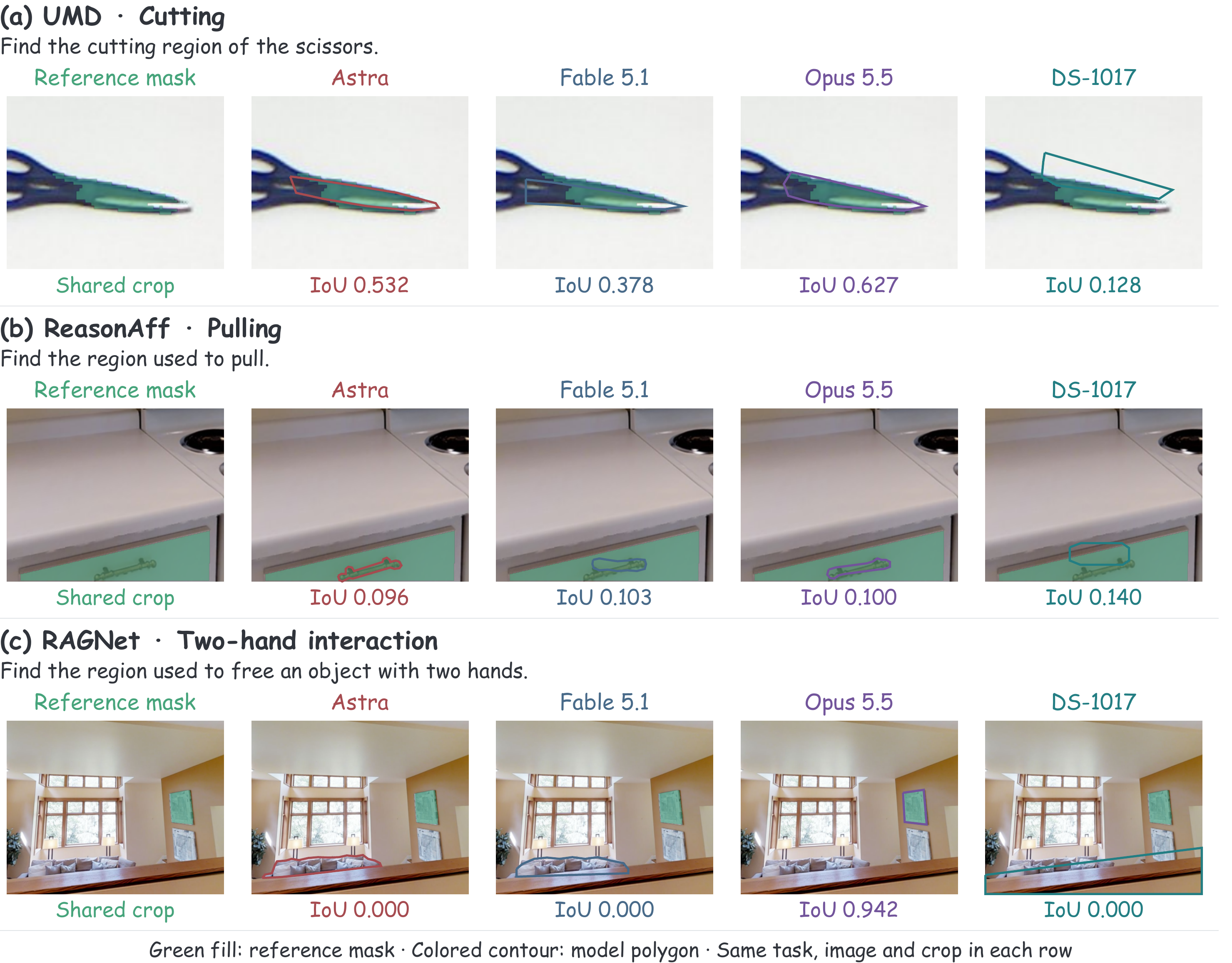}
\caption{\textbf{Affordance: region shape and target identity are separate challenges.} Direct polygon outputs on the same UMD, ReasonAff, and RAGNet tasks. Green fill is the reference mask; colored native contours are model predictions. Each row uses one shared crop for the reference, Astra, Fable~5.1, Opus~5.5, and \mbox{DeepSeek-1017}. IoU is computed over the full original image, not the crop. (a) Scissor outlines differ in extent and alignment. (b) Handle contours cover only part of the annotated pulling region. (c) Agents select different objects under the two-hand instruction. These are single-run examples; the project-page gallery provides all ten models, full images, and recorded polygon coordinates.}
\label{fig:latest-affordance}
\par\medskip
\begin{minipage}{\linewidth}
\textbf{Boundary precision and task interpretation fail differently.}
On the scissors, Opus~5.5 reaches 0.627 IoU while \mbox{DeepSeek-1017} reaches 0.128: both identify the tool, but their outlines recover different parts of its cutting region. On the RAGNet image, Opus~5.5 reaches 0.942 while the other three displayed models score zero, reflecting different target selections rather than a small boundary error.

\smallskip
\textbf{A plausible contact is not a complete region.}
All four pulling predictions concentrate on the handle, while the annotation covers a larger region; IoU remains 0.096--0.140. This example explains why contact validity and region overlap measure different aspects of affordance, and why neither alone establishes successful physical execution.
\end{minipage}
\end{figure}

\begin{figure}[!htbp]
\centering
\includegraphics[width=\linewidth]{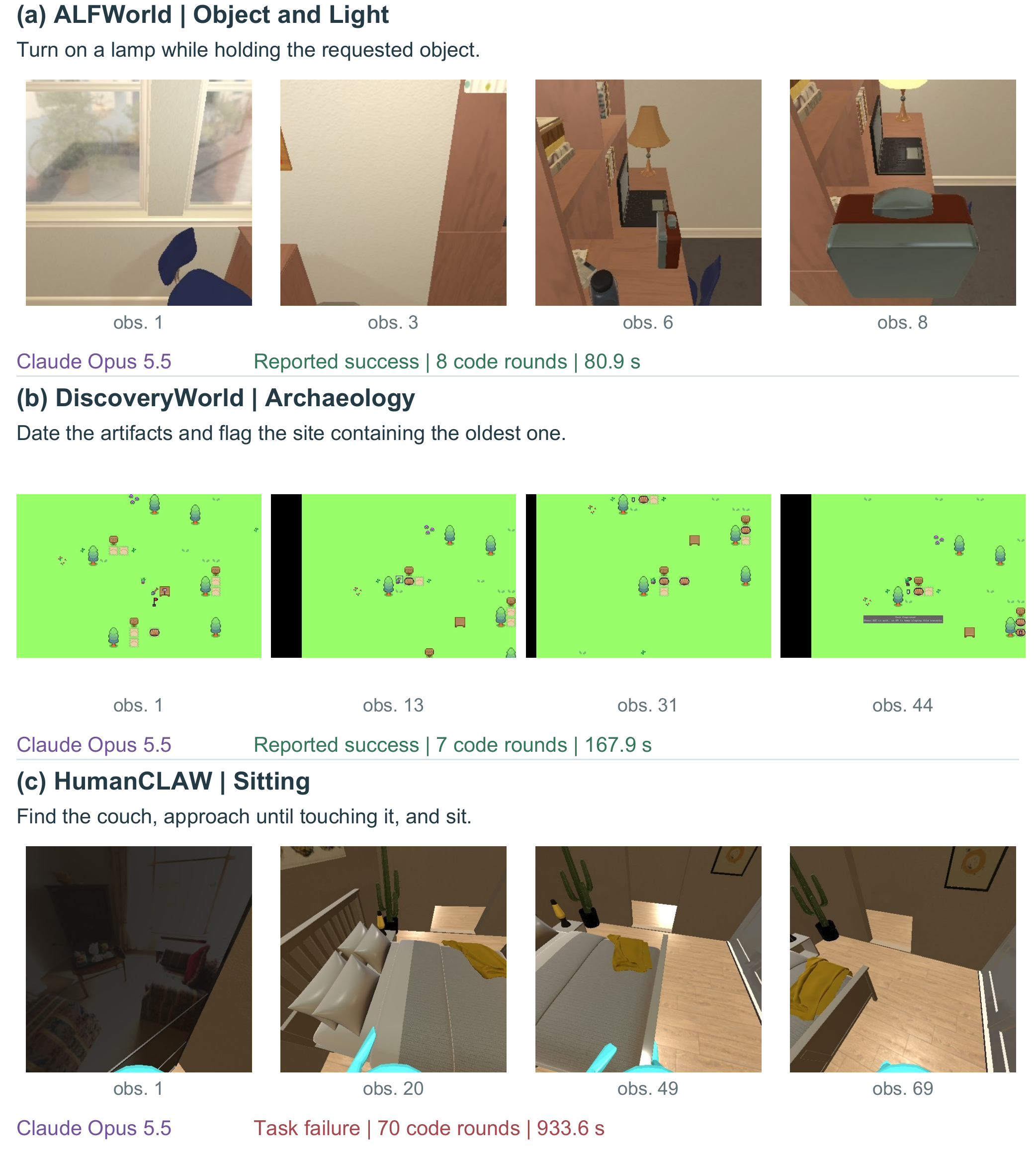}
\caption{\textbf{Task Planning: a terminal episode need not be a completed task.} Additional Opus~5.5 runs. (a) ALFWorld reports success after 8 code rounds under ceilings of 20 minutes, 8 rounds, 120 environment steps, 540 primitive calls, and 320k tokens. (b) DiscoveryWorld reports success after 7 rounds under 20 minutes, 8 rounds, 100 steps, 400 calls, and 320k tokens; its text/JSON interaction is illustrated with environment snapshots. (c) HumanCLAW stops after 70 rounds (933.6\,s) without completing the couch-sitting task, under ceilings of 180 minutes, 101 rounds, 100 steps, 400 calls, and 2M tokens. The HumanCLAW episode ends in task failure before the time ceiling. Frames sample saved observations at discrete times.}
\label{fig:latest-planning}
\end{figure}

\begin{figure}[!htbp]
\centering
\includegraphics[width=\linewidth]{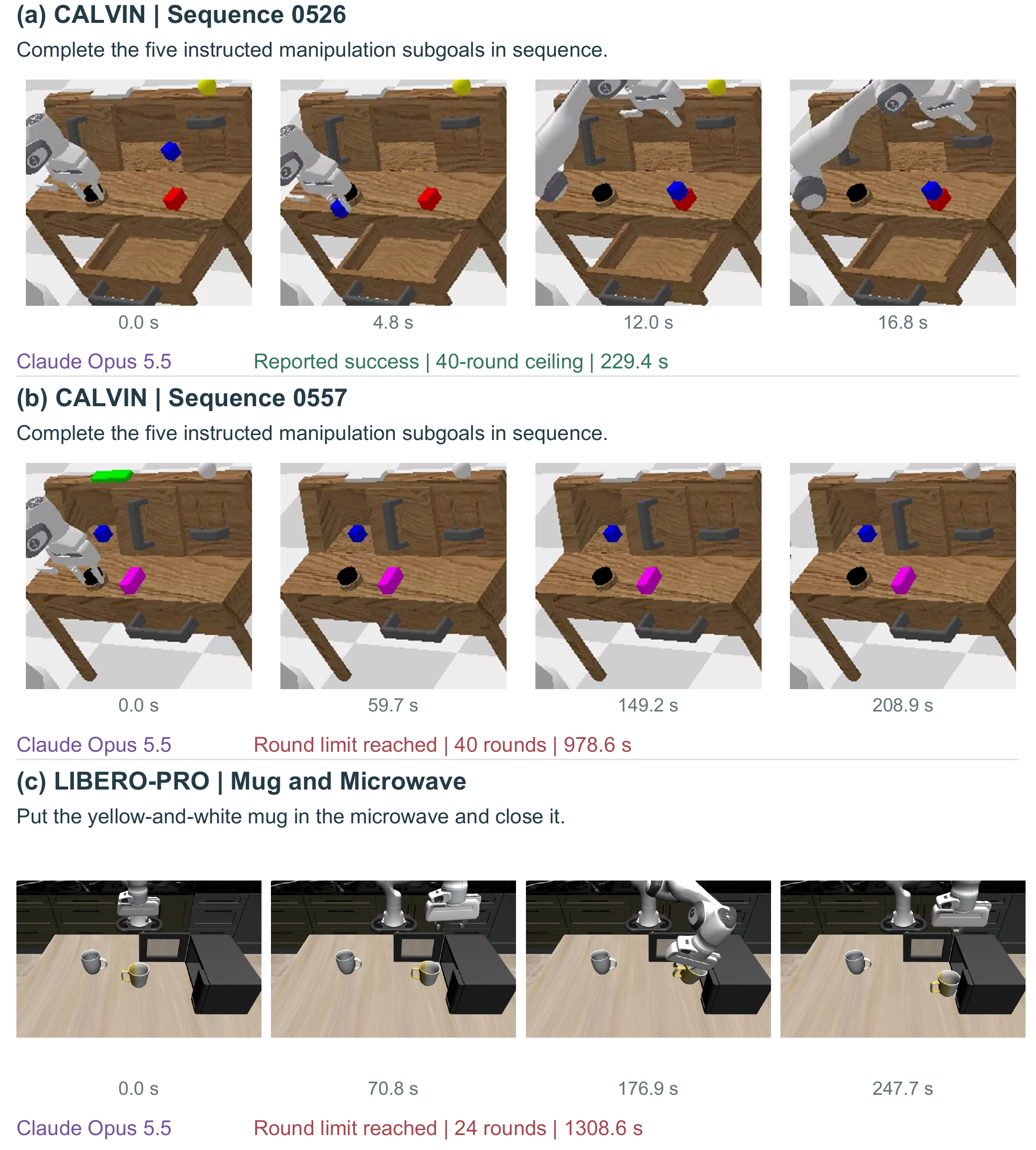}
\caption{\textbf{Manipulation: motion can continue without completing all goals.} Additional Opus~5.5 runs through recorded source-native interfaces. (a) CALVIN sequence 0526 reports success; (b) sequence 0557 exhausts 40 model rounds. Both allow 7,000 primitive calls and 1.28M tokens. (c) LIBERO-PRO requires placing a mug inside the microwave and closing it; the run exhausts 24 model rounds, with ceilings of 256 primitive calls and 640k tokens. Outcomes come from the terminal reports. Frame labels index the encoded recording, whereas the seconds beside each outcome are wall-clock runtime. The three panels illustrate distinct task instances and source interfaces.}
\label{fig:latest-manipulation}
\end{figure}

Figures~\ref{fig:latest-geometry}--\ref{fig:latest-manipulation} connect instructions to predictions and execution. The public \href{https://embodied-agent-arena.github.io/embodied-agent-arena/case-atlas.html}{case atlas} covers 181 tasks across 43 benchmark entries and subsets, with ten-model results on each task. The execution explorer links 460 episode records in 46 same-task groups across all ten models, including 189 detailed action/feedback traces (Appendix~\ref{app:execution-analysis}). The selection spans task types, model disagreements, and successful and unsuccessful outcomes. Recordings, observation replays, and scene-only context are labeled separately.


\input{sections/appendix_multimodel_cases}

%% file: sections/appendix_multimodel_cases.tex
\FloatBarrier
\begin{figure}[!htbp]
\centering
\begin{minipage}{\linewidth}
\subsection{Execution Across Models}
\label{app:multimodel-cases}
The ALFRED montage pairs instructions and observations with terminal outcomes. Tasks are selected separately for each model.
\end{minipage}\par\medskip
\includegraphics[width=.90\linewidth]{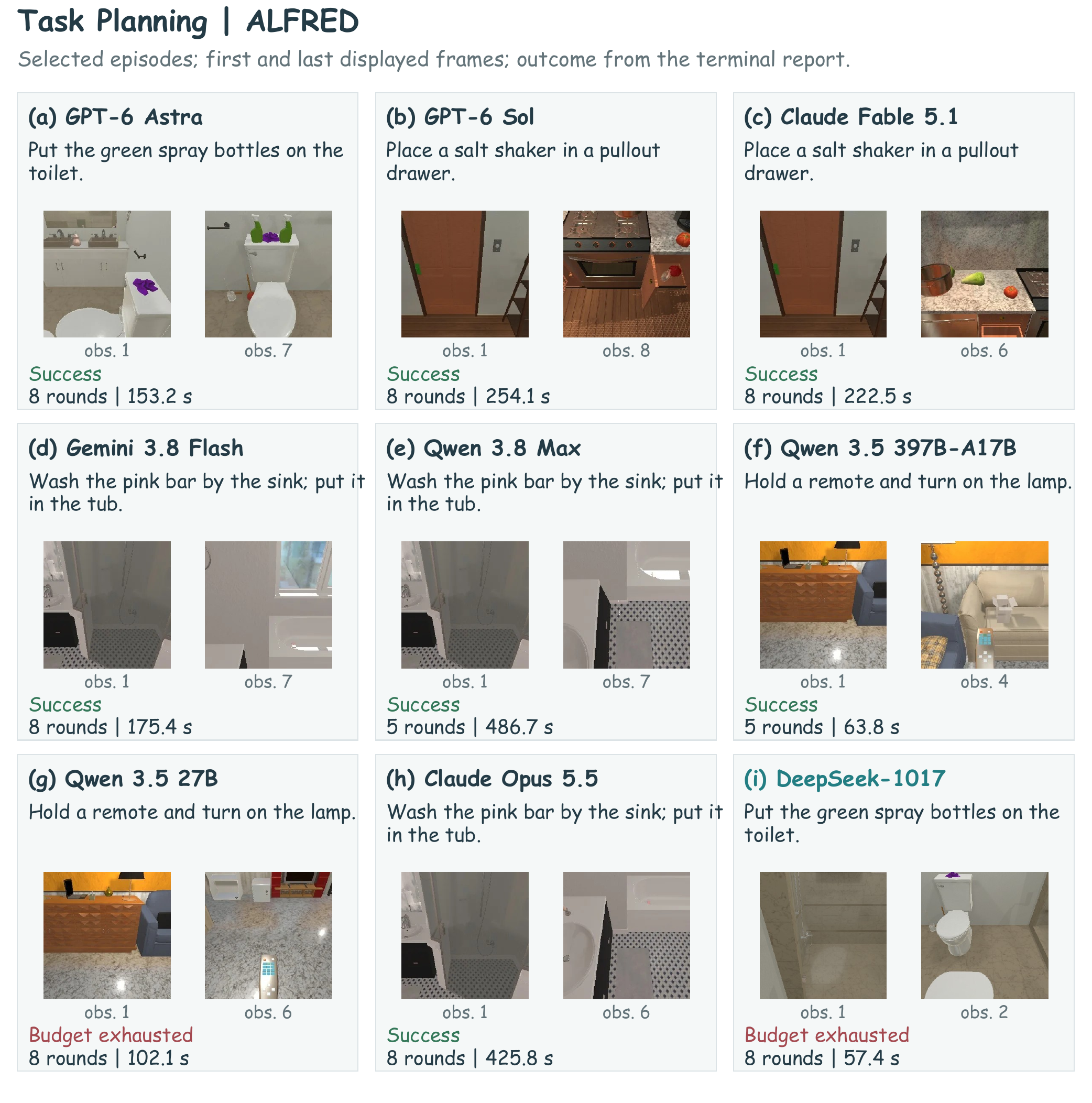}
\caption{\textbf{Nine-model task execution in ALFRED.} Each panel shows the first and last displayed observations from an individual Baseline episode, with a shortened instruction, terminal outcome, rounds, and wall-clock runtime. All runs allow 8 model rounds, 120 environment steps, and 20 minutes. \mbox{DeepSeek-1017} exhausts 8 rounds on the same spray-bottle task as Astra. The two Qwen models in panels (f,g) share the remote-and-lamp task: \mbox{Qwen-397B} reports success after 5 rounds, whereas \mbox{Qwen-27B} exhausts its budget at 8. Other panels cover separately selected tasks. Saved observations preserve temporal order with unequal intervals.}
\label{fig:eight-model-planning}
\par\medskip\begin{minipage}{\linewidth}
\noindent\textbf{Interaction count and elapsed time measure different costs.}
On the shared ALFWorld task (Table~\ref{tab:shared-execution-cases}), Sol succeeds in six rounds versus Fable's seven, yet takes more than twice as long (433.8 versus 197.5\,s). In LIBERO-PRO, three displayed agents complete the placement while Qwen-397B exhausts its budget. Completion, interactions, and elapsed time must therefore be read together. Appendix~\ref{app:matched-latest-cases} extends this comparison to identical archaeology, drawer, and charger instances.
\end{minipage}
\end{figure}
\input{sections/appendix_matched_cases}

%% file: sections/appendix_matched_cases.tex
\begin{figure}[p]\centering
\begin{minipage}{\linewidth}
\subsection{Matched Cases: Recovery, Interaction, and Contact}
\label{app:matched-latest-cases}
Figures~\ref{fig:matched-recovery}--\ref{fig:matched-contact} compare three matched execution cases. Each uses one task identifier and seed across all ten agents. Matching visual records show Opus~5.5, \mbox{DeepSeek-1017}, and Gemini~4~Argon; accompanying tables retain every agent's outcome. Saved frames, terminal reports, and the public \href{https://embodied-agent-arena.github.io/embodied-agent-arena/execution-guide.html}{execution explorer} are linked by model and task. The preceding ALFRED montage instead illustrates separately selected tasks.

\end{minipage}
\medskip

\includegraphics[width=\linewidth]{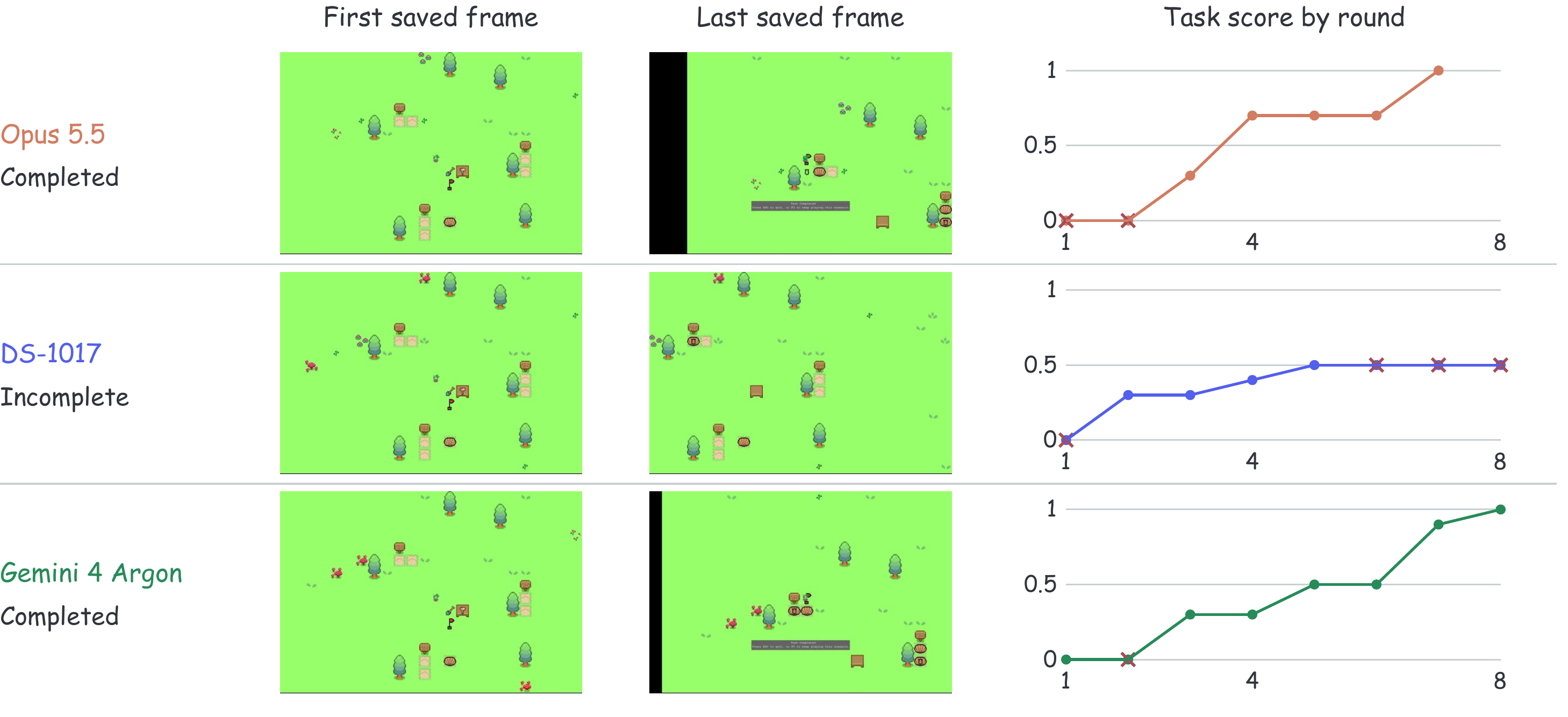}
\caption{\textbf{Subgoal scores and task outcomes after code errors.} Matching saved observations and native score histories for one DiscoveryWorld archaeology task. Crosses mark rounds containing code errors; a score of 1 accompanies native success in these records. The two images are the first and last saved frames of each observation replay; they are not synchronized with the adjacent round axis. All three runs allow 8 model rounds, 100 environment steps, 400 primitive calls, 320k tokens, and 20 minutes.}
\label{fig:matched-recovery}
\par\medskip
\begin{minipage}{\linewidth}
\captionof{table}{\textbf{DiscoveryWorld archaeology, Normal, seed 4.} Gallery episodes on the same task instance, retaining recorded interfaces. Rounds count execution rounds; time includes model and tool execution.}
\label{tab:matched-latest-cases}
\centering\ArenaTableFont\setlength{\tabcolsep}{4pt}
\begin{tabular*}{\linewidth}{@{\extracolsep{\fill}}llrr@{}}\toprule
Model & Terminal outcome & Rounds & Time (s)\\\midrule
Astra & Success & 6 & 319.0\\
Sol & Success & 7 & 391.5\\
Fable & Success & 5 & 89.4\\
Q-Max & Task failure & 8 & 581.0\\
Gemini~3.8 & Success & 8 & 53.6\\
Q-397B & Budget exhausted & 8 & 77.9\\
Q-27B & Budget exhausted & 8 & 151.3\\
Opus~5.5 & Success & 7 & 167.9\\
DS-1017 & Budget exhausted & 8 & 51.3\\
Gemini~4 & Success & 8 & 134.3\\
\bottomrule\end{tabular*}
\end{minipage}
\par\medskip\begin{minipage}{\linewidth}
\noindent\textbf{Task outcomes diverge after code errors.}
The task requires excavating three sites, dating their artifacts, and placing a red flag at the oldest site (Archaeology\_Dating, Normal, seed 4). Opus~5.5 and Gemini~4~Argon encounter code errors but succeed in seven and eight rounds. \mbox{DeepSeek-1017} reaches score 0.5 at round five; subsequent errors leave it unchanged through the eighth-round limit. Its 51.3\,s episode is shorter than either successful run, but ends without task success. Six of ten agents complete this instance. The score histories distinguish eventual task success after an error from further errors without additional task achievement.
\end{minipage}
\end{figure}

\begin{figure}[p]\centering
\includegraphics[width=\linewidth]{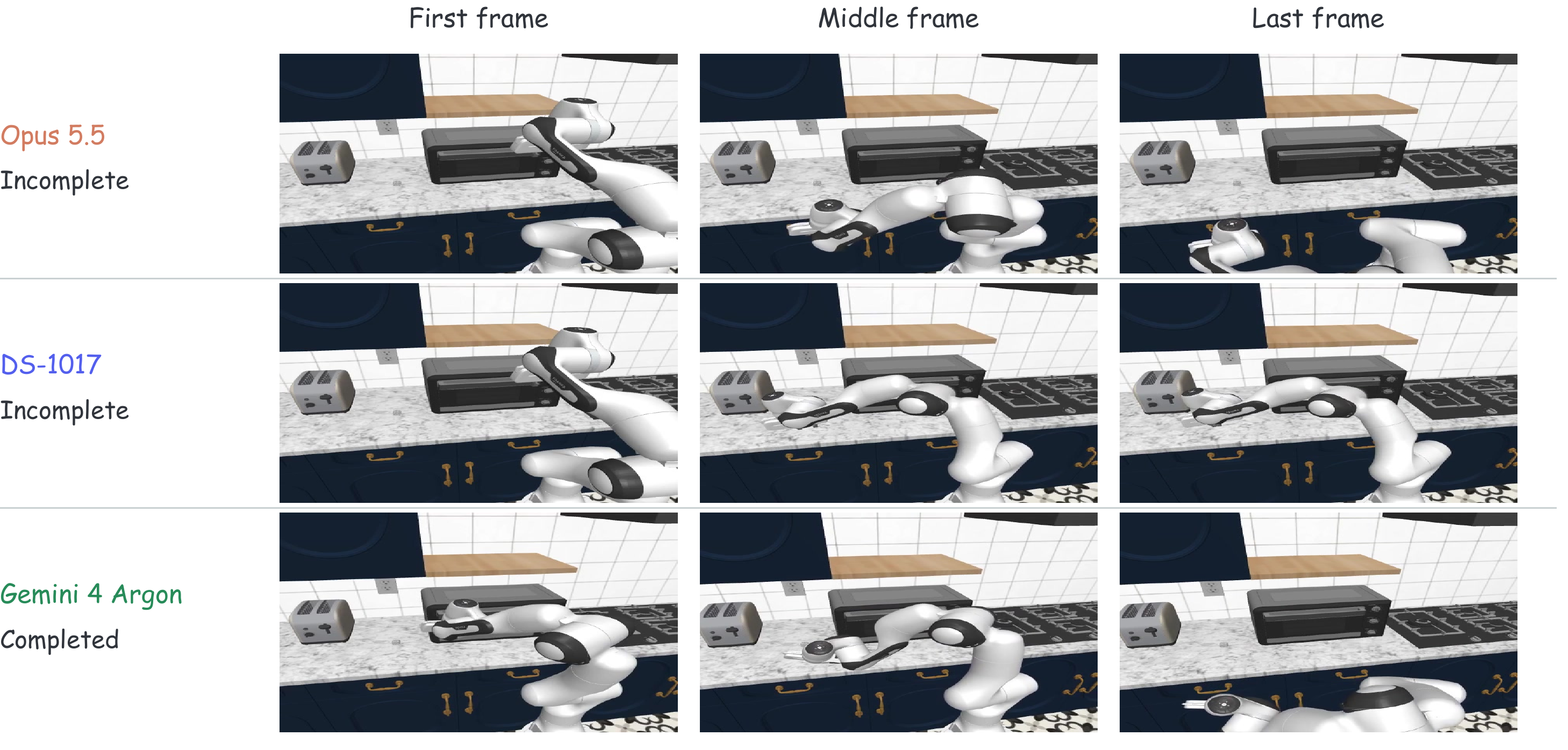}
\caption{\textbf{The same drawer task separates activity from completion.} First, middle, and last frames from each model's matched RoboCasa clip, in temporal order. Opus~5.5 and \mbox{DeepSeek-1017} exhaust their budgets; Gemini~4~Argon completes the task. All three allow 24 model rounds, 256 primitive calls, and 1.12M tokens. The common task timeout is 60 minutes; Gemini~4 also records an 80-minute outer timeout. Clips cover selected portions of execution, not the complete wall-clock runtime. Outcomes use the terminal reports.}
\label{fig:matched-drawer}
\par\medskip
\begin{minipage}{\linewidth}
\captionof{table}{\textbf{RoboCasa OpenDrawer, seed 0.} Gallery episodes on the same task instance, retaining recorded interfaces. Rounds count execution rounds; time includes model and tool execution.}
\label{tab:matched-drawer}
\centering\ArenaTableFont\setlength{\tabcolsep}{4pt}
\begin{tabular*}{\linewidth}{@{\extracolsep{\fill}}llrr@{}}\toprule
Model & Terminal outcome & Rounds & Time (s)\\\midrule
Astra & Budget exhausted & 24 & 470.2\\
Sol & Budget exhausted & 24 & 735.5\\
Fable & Success & 20 & 618.1\\
Q-Max & Budget exhausted & 24 & 3,223.2\\
Gemini~3.8 & Budget exhausted & 24 & 246.3\\
Q-397B & Budget exhausted & 24 & 197.4\\
Q-27B & Budget exhausted & 24 & 215.5\\
Opus~5.5 & Budget exhausted & 24 & 511.8\\
DS-1017 & Budget exhausted & 24 & 195.0\\
Gemini~4 & Success & 13 & 348.0\\
\bottomrule\end{tabular*}
\end{minipage}
\par\medskip\begin{minipage}{\linewidth}
\noindent\textbf{Recorded arm motion does not establish drawer-opening success.}
On RoboCasa's OpenDrawer instance (seed 0), Fable's supplementary episode and Gemini~4~Argon's evaluation episode succeed in 20 and 13 rounds. Fable's supplementary episode is separate from the baseline in Table~\ref{tab:w4-protocol-success}. The other eight agents exhaust their budgets at round 24. Among the three illustrated runs, \mbox{DeepSeek-1017} terminates sooner than Gemini~4~Argon (195.0 versus 348.0\,s) but never meets the drawer-opening condition; Opus~5.5 also remains unsuccessful after 511.8\,s. This reverses the apparent efficiency ordering if one looks only at runtime. Task efficiency must be assessed together with native success: a shorter unsuccessful episode does not demonstrate faster task completion.
\end{minipage}
\end{figure}

\begin{figure}[p]\centering
\includegraphics[width=\linewidth]{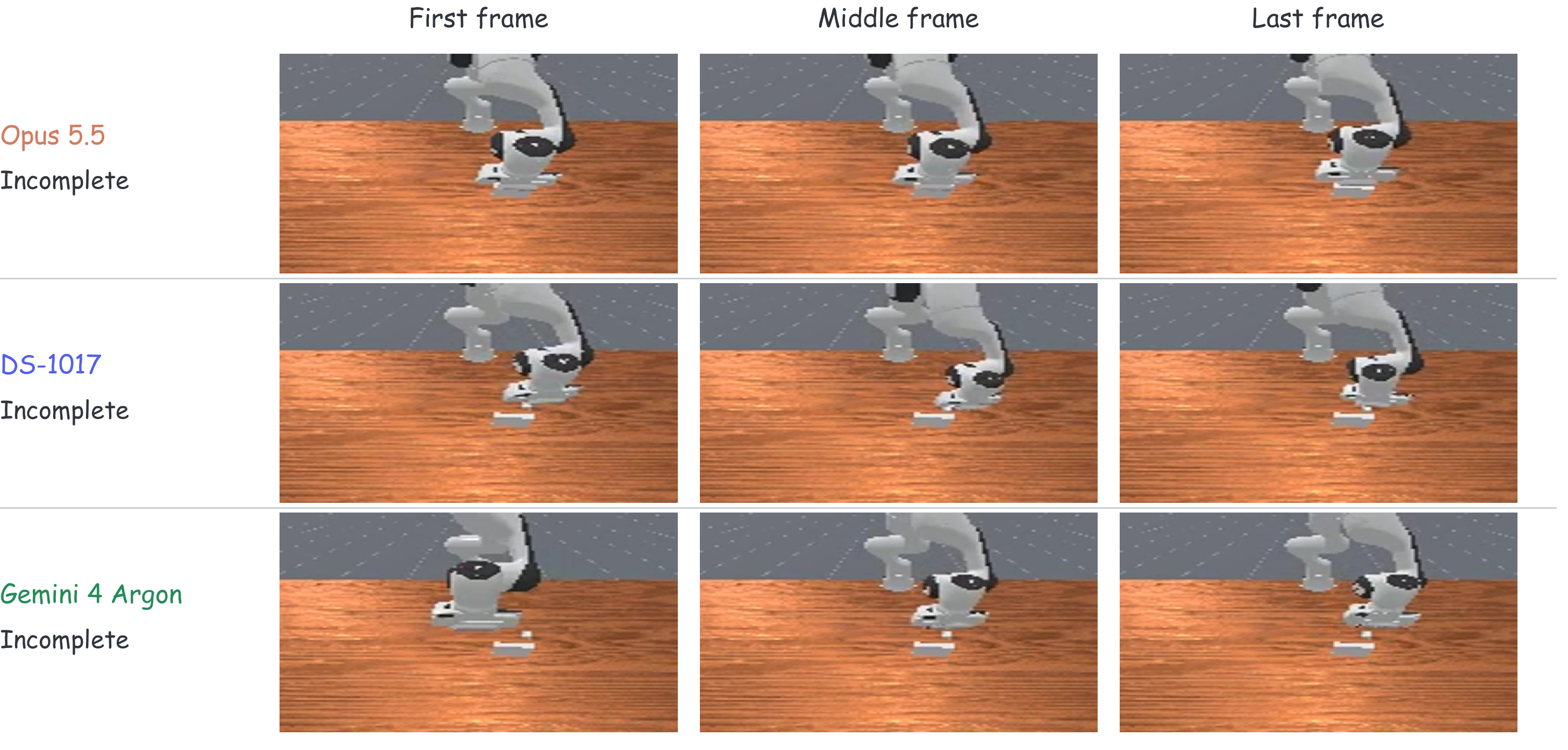}
\caption{\textbf{Satisfying the orientation tolerance does not ensure insertion.} Matching ManiSkill PlugCharger-v1 clips, sampled at their first, middle, and last frames. Opus~5.5 ends in task failure; \mbox{DeepSeek-1017} and Gemini~4~Argon exhaust their budgets. All allow 32 model rounds, 1,024 primitive calls, 640k tokens, and a 60-minute task timeout; Gemini~4 records a 74-minute outer timeout. Opus reaches the native 400-step episode limit. The frames illustrate execution; the numerical distances and angles in the text come from terminal verification.}
\label{fig:matched-contact}
\par\medskip
\begin{minipage}{\linewidth}
\captionof{table}{\textbf{ManiSkill PlugCharger-v1, seed 0.} Gallery episodes on the same task instance, retaining recorded interfaces. Rounds count execution rounds; time includes model and tool execution.}
\label{tab:matched-precision}
\centering\ArenaTableFont\setlength{\tabcolsep}{4pt}
\begin{tabular*}{\linewidth}{@{\extracolsep{\fill}}llrr@{}}\toprule
Model & Terminal outcome & Rounds & Time (s)\\\midrule
Astra & Task failure & 25 & 744.9\\
Sol & Budget exhausted & 29 & 761.6\\
Fable & Budget exhausted & 29 & 1,521.0\\
Q-Max & Task failure & 13 & 1,532.5\\
Gemini~3.8 & Budget exhausted & 24 & 391.6\\
Q-397B & Task failure & 19 & 196.8\\
Q-27B & Budget exhausted & 22 & 164.8\\
Opus~5.5 & Task failure & 15 & 270.0\\
DS-1017 & Budget exhausted & 32 & 479.8\\
Gemini~4 & Budget exhausted & 21 & 547.5\\
\bottomrule\end{tabular*}
\end{minipage}
\par\medskip\begin{minipage}{\linewidth}
\noindent\textbf{Insertion requires both position and orientation tolerances.}
None of these ten gallery episodes completes ManiSkill's PlugCharger-v1 instance (seed 0). The Fable episode supplements the unavailable baseline record; it does not change the aggregate denominator. At final verification, \mbox{DeepSeek-1017} has 9.76$^\circ$ angular error and Gemini~4~Argon has 3.97$^\circ$: both satisfy the angular threshold of 0.2\,rad. Their position errors remain 12.40 and 15.94\,cm, respectively, far beyond the required 0.5\,cm. Qwen-Max and Qwen-397B also meet the angular threshold but remain 15.94\,cm from the position target. These episodes satisfy the orientation criterion but fail the position criterion. Evaluating both components of the joint acceptance rule (Appendix~\ref{app:w4-metrics}) identifies the remaining position error rather than attributing failure to orientation from the video alone.
\end{minipage}
\end{figure}

\FloatBarrier

%% file: sections/appendix_argon.tex
\section{Gemini~4~Argon: Coverage and Execution Settings}
\label{app:argon}
Gemini~4~Argon (\texttt{gemini-4-argon}; 20--21 October 2026) contributes 3,192 trials covering 1,000 cases, plus 70 polygon submissions. Non-tracing Geometry, Spatial Reasoning, and point/box Affordance use five repetitions; tracing, polygons, Planning, and Manipulation use one. Continuous metrics equally weight within-run means over valid outputs. Spatial inputs use corrected 3DSRBench flips and native labels without YOLOE. Tables~\ref{tab:profile} and~\ref{tab:profile-full} compare all ten models.

\begin{table}[!htbp]
\centering\ArenaTableFont
\caption{Gemini~4~Argon Manipulation: task success and per-episode ceilings. Tokens include reasoning and context; rounds are decision interactions. One episode per binary assignment.}\label{tab:argon-budgets}
\begin{tabular*}{\linewidth}{@{\extracolsep{\fill}}lrrrrr@{}}\toprule
Source & Success & SR (\%) & Rounds & Tokens (k) & Time (s)\\\midrule
CALVIN & 4/10 & 40.0 & 40 & 1,280 & 7,200\\
BEHAVIOR-1K & 0/5 & 0.0 & 24 & 640 & 10,800\\
LIBERO-PRO & 8/10 & 80.0 & 24 & 640 & 10,800\\
robosuite & 5/5 & 100.0 & 24 & 640 & 10,800\\
CLIPort & 7/18 & 38.9 & 24 & 960 & 3,600\\
ManiSkill & 5/9 & 55.6 & 32 & 640 & 3,600\\
RLBench & 8/10 & 80.0 & 16 & 640 & 3,600\\
RoboCasa & 2/10 & 20.0 & 24 & 1,120 & 3,600\\
RoboCasa365 & 1/10 & 10.0 & 24 & 1,120 & 3,600\\
RoboTwin 2.0 & 14/50 & 28.0 & 16 & 640 & 3,600\\
RoboWits & 1/10 & 10.0 & 24 & 640 & 3,600\\
VIMA & 9/10 & 90.0 & 16 & 640 & 3,600\\
VLABench & 16/25 & 64.0 & 16 & 640 & 3,600\\
\bottomrule\end{tabular*}\end{table}

Full results appear in Tables~\ref{tab:profile} and~\ref{tab:profile-full}. Planning succeeds on 110/157 tasks: 14/20 ALFRED, 16/20 ALFWorld, 6/10 DiscoveryWorld, 2/13 HumanCLAW, 52/73 ScienceWorld, and 20/21 VirtualHome. Both HumanCLAW successes are navigation; all eight sitting tasks fail. With current RGB at every decision and the listed budgets, Manipulation succeeds on 80/182 binary tasks; Anymal's continuous return remains separate. CALVIN averages 3.30 consecutive goals with 4/10 complete sequences (Figure~\ref{fig:astra-manipulation}).

%% file: sections/appendix_execution_analysis.tex
\section{From Recorded Actions to Complete Tasks}
\label{app:execution-analysis}
This analysis connects the aggregate scores to task overlap, resource allocation, and the state changes inside an episode. We analyze 1,570 Planning episodes: one episode per model on the same 157 assignments for all ten models. The \href{https://embodied-agent-arena.github.io/embodied-agent-arena/technical-report.html}{online analysis} provides source filters, model-pair comparisons, and downloadable episode metrics. Its execution explorer links 460 episode records in 46 same-task groups, including 189 detailed action/feedback traces, to model decisions, primitive actions, returned feedback, and terminal outcomes.

\begin{figure}[!htbp]
\centering
\includegraphics[width=\linewidth]{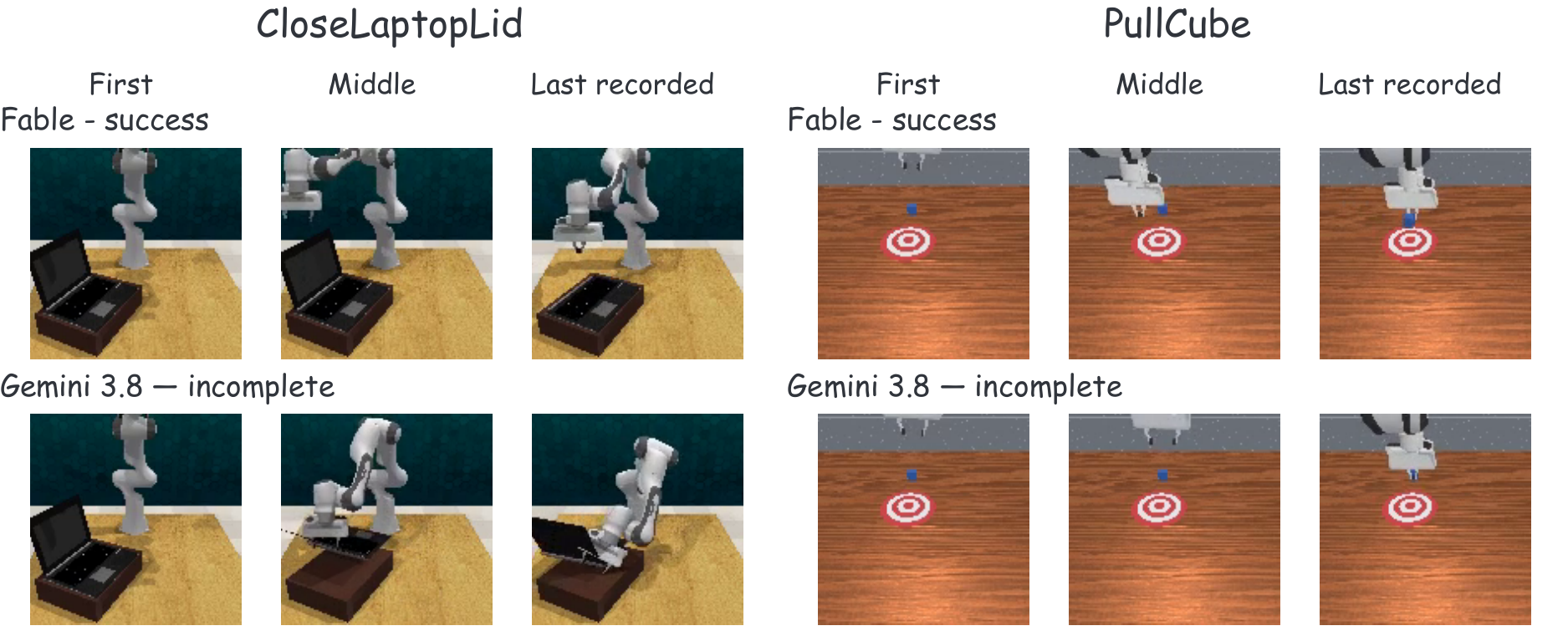}
\caption{Execution contrasts under matched task settings and budgets. CloseLaptopLid uses RLBench's front camera; PullCube uses ManiSkill's base camera. The native evaluator determines task success.}
\label{fig:execution}
\end{figure}

\subsection{Complementary Strengths and Shared Failures}
\begin{figure}[htbp]
\centering
\includegraphics[width=\linewidth]{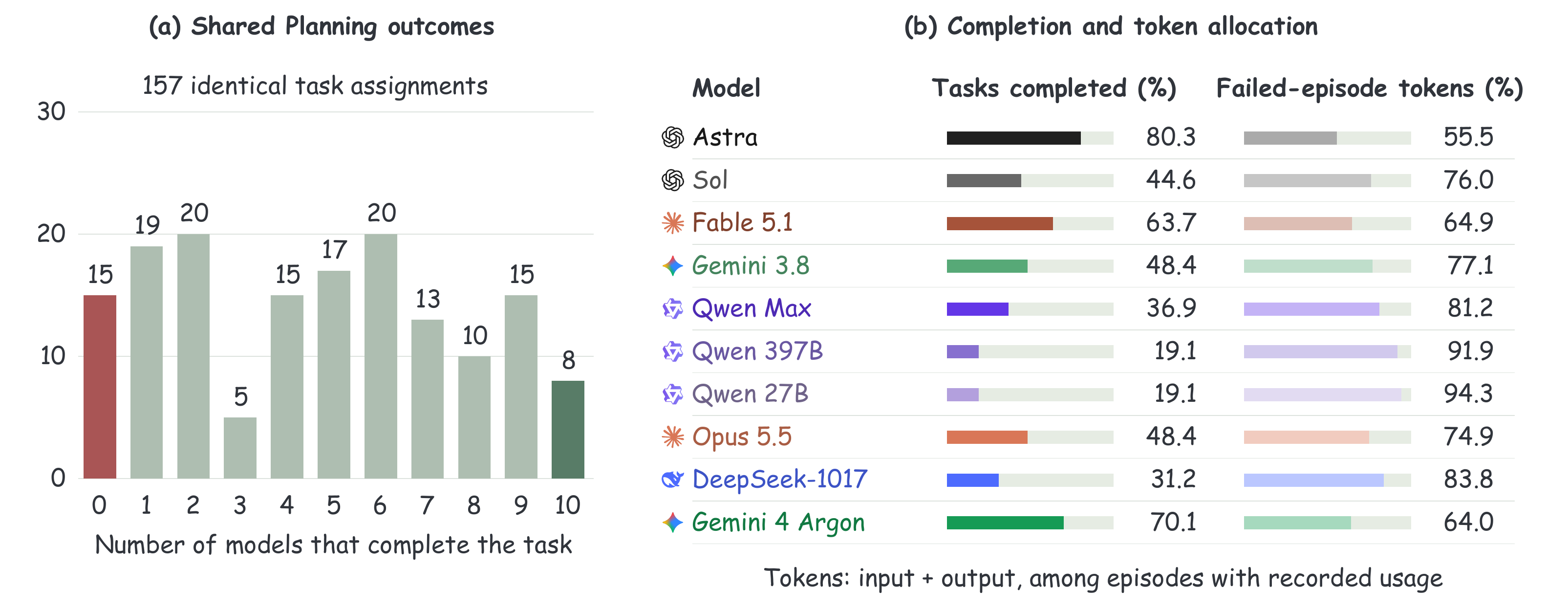}
\caption{\textbf{Planning success overlap and token use in unsuccessful episodes.} (a) Number of tasks completed by exactly zero through ten models, over the same 157 assignments. (b) Completed-task percentage and the share of recorded tokens spent in unsuccessful episodes. Tokens sum recorded input and output usage; Table~\ref{tab:execution-resources} gives usage coverage. All model names use the identifiers in Appendix~\ref{app:settings}.}
\label{fig:execution-profile}
\end{figure}

Astra completes 126 tasks. Taking the union of all ten models' successes covers 142 tasks, adding 16 beyond Astra; 15 tasks remain unsuccessful for every model. Only eight tasks are completed by all ten (Figure~\ref{fig:execution-profile}a). The common failure set contains two ALFRED, two ALFWorld, nine HumanCLAW, and two ScienceWorld tasks. The HumanCLAW subset includes all eight sitting tasks and one television task. This overlap distinguishes tasks on which models complement one another from those on which model selection alone does not resolve the observed failure. These unions measure retrospective coverage across independently executed model episodes.

\noindent\textbf{Concentration of complementary coverage.}
Enumerating all 45 model pairs gives a maximum union of 137 tasks for Astra and Gemini~4~Argon. This covers 11 of the 16 tasks solved by other models but not Astra. Adding Fable raises the union to 140; all ten models reach 142. Of those 142 tasks, 39 have only one or two successful models. The marginal gain falls sharply after the first complementary model.

\noindent\textbf{Sensitivity to task composition.}
Removing the 73 ScienceWorld and 21 VirtualHome assignments leaves 63 tasks. Astra, Gemini~4~Argon, and Fable complete 42, 38, and 21, respectively; every model's rate is lower on this subset than on all 157 tasks. For Manipulation, we additionally compute the unweighted mean of the thirteen source success rates in Table~\ref{tab:w4-protocol-success}. Opus~5.5 changes from 87/182 (47.8\%) to 44.5\%, while Gemini~4 changes from 80/182 (44.0\%) to 47.4\%. Both have complete source coverage. RoboTwin supplies 50/182 assignments and favors Opus, 23 versus 14 successes; equal source weighting reduces its contribution. This diagnostic changes only aggregation weights, retaining the recorded per-source protocols and outcomes.

\subsection{Completion, Latency, and Token Allocation}
\begin{table}[!htbp]\centering\ArenaTableFont
\caption{\textbf{Shared-task planning and placement records.} Rows within each block use the same task identifier. Rounds are recorded model responses; time includes model and tool execution. ALFWorld allows 8 rounds, 120 steps, and 20 minutes. LIBERO-PRO allows 24 rounds, 256 primitive calls, and 640k tokens. Recorded source-interface variants are retained. The complete ten-model drawer example appears in Table~\ref{tab:matched-drawer}.}\label{tab:shared-execution-cases}
\setlength{\tabcolsep}{4pt}
\begin{tabular*}{\linewidth}{@{\extracolsep{\fill}}lrrl@{}}\toprule
Model & Rounds & Time (s) & Recorded outcome\\\midrule
\multicolumn{4}{@{}l}{\textbf{ALFWorld: put a cleaned bar of soap on a counter}}\\*
GPT-6 Sol & 6 & 433.8 & Success \\
Claude Fable 5.1 & 7 & 197.5 & Success \\
\midrule
\multicolumn{4}{@{}l}{\textbf{LIBERO-PRO: put the tomato sauce in the basket}}\\*
Claude Fable 5.1 & 16 & 1,553.5 & Success \\
Gemini~3.8 Flash & 15 & 429.0 & Success \\
Qwen 3.5 397B-A17B & 24 & 934.6 & Budget exhausted \\
Claude Opus~5.5 & 20 & 2,376.8 & Success \\
\bottomrule\end{tabular*}\end{table}
Unsuccessful episodes consume 55.5--94.3\% of each model's recorded Planning tokens (Figure~\ref{fig:execution-profile}b). Astra completes 80.3\% of its assignments, yet its 31 unsuccessful episodes account for 55.5\% of recorded tokens. The mean unsuccessful episode uses 5.1 times as many tokens as the mean successful episode (591,162 versus 116,435 tokens). These attempts dominate aggregate token use despite being a minority of its tasks; the source-conditioned decomposition in Table~\ref{tab:failure-cost-decomposition} locates this concentration in HumanCLAW. The median episode duration ranges from 57.1\,s for \mbox{DeepSeek-1017} to 300.0\,s for \mbox{Qwen-Max} (Table~\ref{tab:execution-resources}). Short duration has to be read with completion: it can describe an efficient solution or an early unsuccessful stop.

\begin{table}[htbp]
\centering
\caption{\textbf{Planning resource use and outcomes after execution errors over 157 tasks per model.} Time is recorded elapsed time in seconds; IQR spans the 25th--75th percentiles. Qwen HumanCLAW durations include worker setup, retries, and teardown. Failed tokens is the percentage of recorded tokens spent in unsuccessful episodes; usage gives the number of episodes with token records, out of 157. The final column counts episodes with native task success after an execution error, divided by episodes containing a code-execution error; provider interruptions are excluded from this column.}
\label{tab:execution-resources}
\ArenaTableFont\setlength{\tabcolsep}{4.2pt}
\begin{tabular}{lrrrrrr}
\toprule
Model & Completed & Median (s) & IQR (s) & Failed tokens (\%) & Usage & Then complete \\
\midrule
Astra & 126/157 & 231.7 & 153.2--361.8 & 55.5 & 157 & 1/1 \\
Sol & 70/157 & 280.7 & 199.9--427.1 & 76.0 & 157 & 1/6 \\
Fable & 100/157 & 109.5 & 62.0--201.4 & 64.9 & 157 & 6/25 \\
Gemini~3.8 & 76/157 & 61.8 & 28.6--175.9 & 77.1 & 157 & 4/24 \\
\mbox{Qwen-Max} & 58/157 & 300.0 & 216.8--506.7 & 81.2 & 145 & 10/38 \\
\mbox{Qwen-397B} & 30/157 & 106.5 & 75.8--166.7 & 91.9 & 157 & 2/44 \\
\mbox{Qwen-27B} & 30/157 & 94.5 & 60.8--146.8 & 94.3 & 157 & 3/31 \\
Opus~5.5 & 76/157 & 135.6 & 105.3--211.0 & 74.9 & 157 & 48/105 \\
\mbox{DeepSeek-1017} & 49/157 & 57.1 & 36.1--78.5 & 83.8 & 157 & 9/50 \\
Gemini~4~Argon & 110/157 & 159.5 & 51.5--612.1 & 64.0 & 157 & 16/28 \\
\bottomrule
\end{tabular}
\end{table}

On the same HumanCLAW chair-navigation task, Gemini~4~Argon completes in 17 rounds and 437.3\,s, Astra in 18 rounds and 470.7\,s, and \mbox{DeepSeek-1017} in 65 rounds but 405.7\,s. Fewer decisions need not imply lower elapsed time. In the selected BEHAVIOR-1K toolbox task, none of the ten models completes the goal. Gemini~4~Argon uses 603,393 tokens before the next round is blocked by the 640,000-token ceiling, after 37.5 minutes of a 180-minute limit. The stopping resource and the unmet goal conditions explain why the episode ends.

\subsection{Task Completion After Execution Errors}
\begin{figure}[htbp]
\centering
\includegraphics[width=\linewidth]{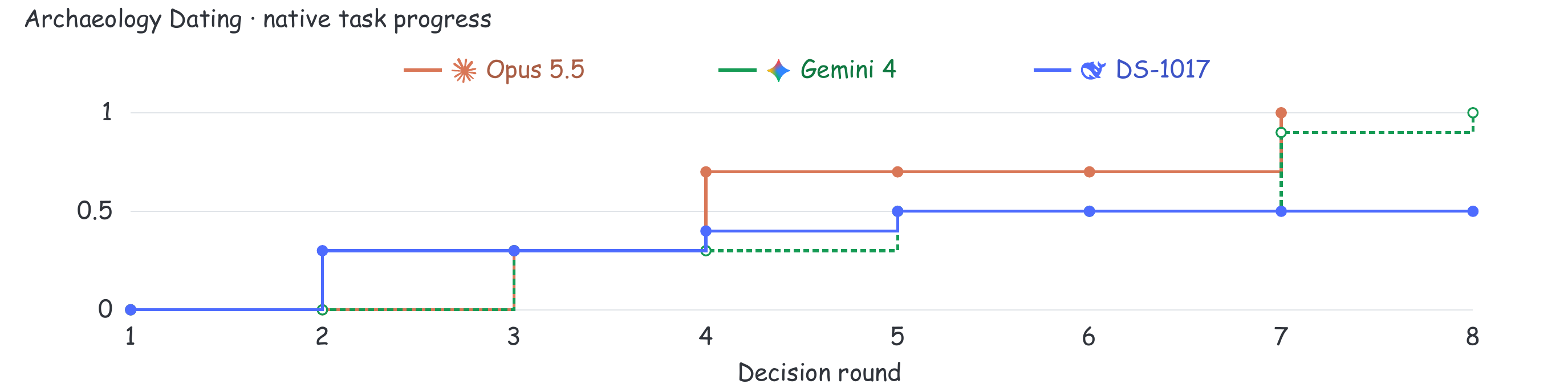}
\caption{\textbf{Task-score histories after code errors.} Archaeology Dating: Opus~5.5 and Gemini~4~Argon recover from early code errors and reach native score 1.0; \mbox{DeepSeek-1017} stalls at score 0.5. Points are recorded decision rounds, with steps showing the latest native score.}
\label{fig:execution-progress}
\end{figure}

Gemini~4~Argon completes 16 of 28 Planning episodes containing an execution error; Opus~5.5 completes 48 of 105. These episodes achieve the task goal despite earlier execution errors. Error records include malformed code and rejected API use in the evaluated interface. The Archaeology Dating task shows the actions preceding eventual success (Figure~\ref{fig:execution-progress}). Opus~5.5's first two rounds fail before any environment action. It subsequently collects the tools, excavates the sites, reads the artifact ages, and places the flag at the oldest site. The native score reaches 1.0 in round seven. Gemini~4~Argon encounters an error in round two, resumes progress in round three, and completes in round eight. \mbox{DeepSeek-1017} reaches 0.5 by round five, but its last three rounds end in code errors without further environment steps.

\subsection{Resumed Code Execution and Full-Sequence Success}
CALVIN's sequence records distinguish error-free code execution from full-sequence success (Figure~\ref{fig:astra-manipulation}). Among 65 episodes containing execution errors, 62 later run code without an execution error, but only three subsequently complete all five ordered goals. Table~\ref{tab:calvin-recovery} gives the per-model counts. An error-free code round does not by itself establish a completed subgoal. Full-sequence success requires all five ordered goals to pass the native verifier.

ALFRED separates activity from goal satisfaction: for cleaning a soap bar and placing it in the bathtub, Gemini~4~Argon satisfies all three conditions in six rounds; \mbox{DeepSeek-1017} ends at one of three after eight. On RoboCasa's left-drawer task, Gemini~4~Argon finishes in 13 rounds despite an error in round ten, whereas Sol exhausts 24 rounds without the required state. On charger insertion, Gemini~4~Argon stops at the token limit after 23 native steps, still 15.9\,cm from the goal. These cases distinguish the number of actions attempted from satisfaction of the final task conditions.

\subsection{Task Alignment and Measurement}
Cross-model comparisons match native task IDs. Post-error success requires native task success after a code-execution error; provider failures, budget guards, and timeouts have separate termination labels. They still count as unsuccessful when they end an episode. Duration and token accounting follow Appendix~\ref{app:scoring}.

The explorer links model decisions to code cells, feedback, and native outcomes by timestamp. Its 189 detailed episodes contain 2,935 rounds and 17,287 recorded tool calls; 271 additional records summarize rounds. The complete collection contains 6,856 rounds and 17,424 recorded tool calls. Curves link recorded native scores or satisfied goal conditions to decisions, source reports, and settings.

\noindent\textbf{Reading the broader case collection.}
Static panels show first-run predictions without YOLOE in Spatial cases. Contact labels test mask membership; box IoU compares rectangles, and polygon IoU compares rasterized contours with reference masks (Appendix~\ref{app:polygon}).

Execution panels retain their source interfaces and observation schedules (Appendix~\ref{app:w4-comparison}). Planning panels preserve saved-observation order; DiscoveryWorld snapshots illustrate text/JSON interaction. Manipulation frames use encoded-video time.

\input{sections/appendix_trace_diagnostics}

%% file: sections/appendix_trace_diagnostics.tex
\subsection{Task-Conditioned Costs and Trace Diagnostics}
\label{app:trace-diagnostics}
Population counts establish how failures are distributed; matched traces identify the actions and feedback that explain representative outcomes. Planning and CALVIN comparisons use shared task identities, while PlaceSphere additionally matches the initial sphere position and execution limits.

\begin{table}[!htbp]
\centering\fontsize{10}{12}\selectfont
\caption{\textbf{Failure cost after separating task sources.} All-failure share is unsuccessful tokens divided by all recorded Planning tokens. Human share is unsuccessful HumanCLAW tokens divided by all unsuccessful Planning tokens. Outside Human is the failed-token share within the other 144 assignments. Usage shows recorded episodes out of 157 overall and thirteen for HumanCLAW. All models have thirteen HumanCLAW usage records; token accounting follows Appendix~\ref{app:scoring}. Qwen-Max has 132/144 outside-Human records; every other model has 144/144.}
\label{tab:failure-cost-decomposition}
\setlength{\tabcolsep}{4pt}
\begin{tabular*}{\linewidth}{@{\extracolsep{\fill}}lrrrl@{}}
\toprule
Model & All failures (\%) & Human share (\%) & Outside Human (\%) & Usage: all / Human \\\midrule
Astra & 55.5 & 81.7 & 19.4 & 157/157; 13/13 \\
Sol & 76.0 & 58.0 & 58.7 & 157/157; 13/13 \\
Fable & 64.9 & 40.0 & 55.3 & 157/157; 13/13 \\
\mbox{Qwen-Max} & 81.2 & 56.7 & 67.8 & 145/157; 13/13 \\
Gemini~3.8 & 77.1 & 53.4 & 62.6 & 157/157; 13/13 \\
\mbox{Qwen-397B} & 91.9 & 56.8 & 85.1 & 157/157; 13/13 \\
\mbox{Qwen-27B} & 94.3 & 63.3 & 85.8 & 157/157; 13/13 \\
Opus~5.5 & 74.9 & 68.8 & 61.0 & 157/157; 13/13 \\
\mbox{DeepSeek-1017} & 83.8 & 57.7 & 73.6 & 157/157; 13/13 \\
Gemini~4~Argon & 64.0 & 72.5 & 33.5 & 157/157; 13/13 \\
\bottomrule\end{tabular*}
\end{table}

\noindent\textbf{The strongest planners have a concentrated costly residual.}
HumanCLAW sitting alone accounts for 63.7\% and 55.9\% of Astra's and Gemini~4's failed tokens. Their HumanCLAW failures average 5.62 and 8.04 times the tokens of successes, versus 0.95 and 1.47 within ALFRED. HumanCLAW also supplies 56.7--63.3\% of the three Qwen models' failed tokens, yet their failed-token shares remain 67.8--85.8\% outside this source. Thus long HumanCLAW episodes amplify expenditure both when most other tasks succeed and when failures remain widespread. These differences combine unresolved task requirements with unequal source allowances: 101 rounds and 2M tokens for HumanCLAW, versus eight rounds and 320k tokens for several other sources.

\noindent\textbf{Search, stopping, and decision allocation are distinct behaviors.}
On HumanCLAW plant case \texttt{104862501\_172226556:17:potted\_plant}, Gemini~4 executes 73 rounds before a budget-guard attempt: one initial inspection followed by 48 turns, 23 forward movements, and one backward movement. The model reports no plant visible in all 72 action descriptions. It records 1,947,369 tokens, of which 99.46\% are prompt tokens. On couch case \texttt{102816036:2:couch}, Gemini~4 issues sit at round~43 and stop at~44, while DeepSeek issues sit at~68, observes at~69, and stops at~70. Both claim a seated state but fail native success. Across seven HumanCLAW traces, including successful chair and plant navigation, prompt tokens comprise 98.23--99.46\% of recorded usage. Token expenditure here mainly counts input processing rather than newly generated response length. A different stopping error appears in \mbox{Qwen-27B}'s plant navigation case \texttt{107734287\_176000160:2:potted\_plant}: its final geometric distance is zero and the geometric proximity flag is true, but no active stop is recorded and the native navigation flag remains false when the budget is exhausted. Reaching the target region and satisfying the navigation termination criterion are separate recorded conditions. On \mbox{Qwen-Max}'s couch case \texttt{104862384\_172226319:19:couch}, twenty of 35 rounds instead end in Python syntax errors, leaving only thirteen environment steps despite 555,225 recorded tokens. The trace separates failed code generation from ineffective physical interaction.

A different bottleneck appears in ScienceWorld \texttt{find-animal-v115-easy}. DeepSeek uses rounds~1, 2, and~5 for inspection without environment steps, searches for the literal word ``animal,'' and reaches the greenhouse only during its final exploration batch. It stops after eight rounds with score~17 and 89/100 environment steps unused. Under the same eight-round allowance, Opus identifies a blue-jay egg, focuses on it (score~67), carries it to the workshop (83), and corrects an invalid put command using the returned move action (100). It uses fourteen environment steps and 25,233 tokens, versus DeepSeek's eleven steps and 63,349 tokens. Neither episode contains a code exception. The contrast concerns target selection and decision allocation.

\noindent\textbf{Shared failure does not imply a shared stopping mechanism.}
Among the fifteen shared failures, ALFWorld clean-spoon case \texttt{T20190911\_131350\_027076} leaves Gemini~4 at score~1/3 from step sixteen to the 120-step limit; DeepSeek also ends at 1/3 after 120 steps. By contrast, Gemini~4's unknown-conductivity case \texttt{v126} progresses to score~83 before a terminal score of $-100$, with no execution error; DeepSeek ends at score~10. These histories distinguish a persistent plateau from terminal failure after partial progress.

\begin{table}[!htbp]
\centering\fontsize{10}{12}\selectfont
\caption{\textbf{Separating interface failures from task failure in CALVIN.} Counts cover the 100 episodes under their recorded budgets. Syntax is the subset of execution-error rounds containing a syntax error; transport failures, timeouts, and budget guards are excluded from execution errors. Rounds are model decision records, not physical action submissions.}
\label{tab:calvin-error-types}
\setlength{\tabcolsep}{5pt}
\begin{tabular*}{\linewidth}{@{\extracolsep{\fill}}lrrrr@{}}
\toprule
Model & Rounds & Error rounds & Syntax & Completed / 10 \\\midrule
Astra & 188 & 3 & 1 & 3 \\
Sol & 193 & 3 & 0 & 0 \\
Fable & 160 & 36 & 34 & 0 \\
\mbox{Qwen-Max} & 200 & 103 & 95 & 0 \\
Gemini~3.8 & 200 & 14 & 0 & 0 \\
\mbox{Qwen-397B} & 99 & 20 & 0 & 0 \\
\mbox{Qwen-27B} & 192 & 22 & 2 & 0 \\
Opus~5.5 & 378 & 27 & 20 & 1 \\
\mbox{DeepSeek-1017} & 388 & 84 & 63 & 1 \\
Gemini~4~Argon & 266 & 7 & 3 & 4 \\
\bottomrule\end{tabular*}
\end{table}

Thirty CALVIN episodes have both an observed native prefix and no execution error. Of these, six finish, thirteen stop after one to four goals, and eleven achieve none. Five additional no-recorded-error \mbox{Qwen-397B} episodes lack a native prefix and are excluded from this breakdown. 

\Needspace{8\baselineskip}
Code admissibility is thus a substantial bottleneck for some agents, while others fail despite valid execution. These groups cross the open-/closed-weight distinction: \mbox{Qwen-Max} and Fable incur many syntax errors, whereas some open-weight episodes execute code without obtaining the requested object state.

\begin{table}[!htbp]
\centering\fontsize{11}{13}\selectfont
\caption{\textbf{Matched traces localize the state change that separates outcomes.} ALFRED uses trial \texttt{T20190908\_064808\_086925}. CALVIN uses sequence~0526, with the same 40-round, 1.28M-token, 120-minute limits for both models and per-round RGB. PlaceSphere-v1 uses the same measured initial sphere centroid and 32-round, 640k-token, 60-minute limits. ``Rounds'' are model decisions; CALVIN action counts are calls to \texttt{submit\_calvin\_action}.}
\label{tab:trace-state-changes}
\setlength{\tabcolsep}{4pt}
\renewcommand{\arraystretch}{1.12}
\begin{tabularx}{\linewidth}{@{}>{\raggedright\arraybackslash}p{.17\linewidth} >{\raggedright\arraybackslash}X >{\raggedright\arraybackslash}p{.18\linewidth}@{}}
\toprule
Model & Recorded action and feedback & Terminal result \\\midrule
\multicolumn{3}{@{}l}{\textbf{ALFRED: wash soap, then put it in the bathtub}}\\*
DeepSeek & Clean soap; valid placement into outer bathtub leaves the native score at 1/3. & 1/3; unsuccessful \\
Gemini~3.8 & Inspect the parent receptacle; pick up soap and place it in the basin. Two further actions change 1/3 to 3/3. & 3/3; successful \\
\mbox{Qwen-Max} & Inspect the parent receptacle, repeat washing, and change the receiving region to the basin. & 3/3; successful \\\midrule
\multicolumn{3}{@{}l}{\textbf{CALVIN: retrieve a block before stacking and switching the light}}\\*
Opus~5.5 & Retrieval at round~11; block height rises with the gripper at round~12; stacking at round~13. & 5/5; 18 rounds; 338 actions \\
DeepSeek & Retrieval remains current during rounds~17--27; block height is near 2\,cm at rounds~20 and~25; stacking starts at~28. & 5/5; 37 rounds; 700 actions \\\midrule
\multicolumn{3}{@{}l}{\textbf{PlaceSphere: establish a grasp, transport, release, and stabilize}}\\*
Opus~5.5 & Grasp confirmed at round~11; release at~12; on-bin and static checks both true at~13. & Success; 51 native steps \\
DeepSeek & Hover error 2.8\,mm at round~7; grasp false at~8 and~11--13; object shifts away from its initial location. & Failure; 100 native steps \\
\bottomrule\end{tabularx}
\end{table}

\noindent\textbf{Reading the physical state rather than the intended action.}
Both CALVIN traces succeed, illustrating different recovery costs. Opus records block--gripper co-motion before advancing; DeepSeek repeatedly returns to retrieval. In PlaceSphere, the 2.8\,mm error is relative to the requested hover target, not a verified grasp configuration. The native Opus step count is 51, distinct from its model-maintained counter of 53.

\begin{table}[!t]
\centering\fontsize{10}{12}\selectfont
\caption{\textbf{Point and box acceptance identify different Affordance errors.} Each row counts recorded case--run predictions on the same 70 cases: five runs for nine models and three for Fable. A point passes if it lies in the reference functional mask; a box passes if its rectangle IoU with the reference box is at least 0.5. This is the native box criterion, distinct from the box--mask IoU in Table~\ref{tab:profile}. Missing metric rows remain separate.}
\label{tab:affordance-joint}
\setlength{\tabcolsep}{5pt}
\renewcommand{\arraystretch}{1.12}
\begin{tabular*}{\linewidth}{@{\extracolsep{\fill}}lrrrrrr@{}}
\toprule
Model & Both & Point only & Box only & Neither & Missing & Total \\\midrule
Astra & 163 & 48 & 0 & 139 & 0 & 350 \\
Sol & 150 & 32 & 3 & 165 & 0 & 350 \\
Fable & 93 & 27 & 4 & 84 & 2 & 210 \\
\mbox{Qwen-Max} & 152 & 33 & 8 & 157 & 0 & 350 \\
Gemini~3.8 & 112 & 44 & 4 & 175 & 15 & 350 \\
\mbox{Qwen-397B} & 18 & 96 & 1 & 234 & 1 & 350 \\
\mbox{Qwen-27B} & 0 & 22 & 1 & 326 & 1 & 350 \\
Opus~5.5 & 178 & 36 & 2 & 134 & 0 & 350 \\
\mbox{DeepSeek-1017} & 96 & 82 & 5 & 165 & 2 & 350 \\
Gemini~4~Argon & 181 & 28 & 16 & 125 & 0 & 350 \\
\bottomrule\end{tabular*}
\end{table}

\noindent\textbf{Coverage, action-point selection, and target identity.}
All sixteen Gemini~4 box-only predictions place the point inside the predicted rectangle but outside the functional mask. For UMD scissors case \texttt{scissors\_02\_00000097\_grasp}, its first-run rectangle IoU is 0.784, yet point $(337,242)$ is outside the grasp mask; the same box-only pattern occurs in four of five runs. This is an action-point error within a localized object, distinct from the wrong-object example in Figure~\ref{fig:latest-affordance}.

\Needspace{28\baselineskip}
\noindent\textbf{Image boundaries and physical boundaries.}
TraceSpatial jointly checks each path: Astra and Gemini~4 hit both image endpoints in 35/56 and 45/56 paths, but 21 and sixteen of these fail 3D initiation or arrival. Collision alone invalidates one Astra path and ten Gemini~4 paths. These joint checks separate endpoint placement from obstacle avoidance.

\noindent\textbf{Submission failure within a difficult spatial group.}
Gemini~4's twenty runs on four hard video-direction cases comprise six correct answers, four incorrect answers, and ten token-guard stops without an answer. The submitted subset is 6/10, but the all-assignment score remains 30\%; conditioning on submission diagnoses a different outcome and does not replace the benchmark score.

\noindent\textbf{Calibrating a metric and detecting motion are separable.}
GeoProbe depth uses twelve cases over five runs. In case~0008, Gemini~4's first-run AbsRel falls from 0.394 to 0.027 after per-case oracle scale alignment. Oracle alignment uses reference depth to isolate scale bias. Each motion regime retains ten moving interventions and two stationary controls. In intervention~0000, Gemini~4 reports approximately 2\,cm with a fixed camera but no motion in all five large-camera runs. Across regimes, \mbox{Qwen-27B} misses all 150 moving observations and correctly labels all thirty stationary controls: unchanged MAE reflects a detection floor.

\noindent\textbf{Review can repair coordinates while reversing correct relations.}
The paired \mbox{Qwen-27B} Spatial records show one wrong-to-correct and two correct-to-wrong changes under self-review; crop review produces two and four, respectively. Both change ``back'' to ``left'' on 3DSRBench case \texttt{WMXV424F}, and ``very different'' to ``similar'' for the TV/fireplace orientations in \texttt{Z9MFBJMX}. Self-review also repairs all six initial out-of-bounds boxes among nine Affordance cases. For RAGNet \texttt{taskonomy\_german\_point\_805\_view\_14\_domain\_rgb}, a rectangle entirely outside the $512\times512$ image becomes an in-image box with IoU~0.447, still below the 0.5 criterion. These paired records explain why representation repair, semantic correction, and accepted task outcomes must be measured separately.